\documentclass[conference]{IEEEtran}
\usepackage{times}
\usepackage{graphicx}
\usepackage[dvipsnames]{xcolor}
\usepackage{amsfonts}

\usepackage[numbers]{natbib}
\usepackage{multicol}
\usepackage[bookmarks=true]{hyperref}
\usepackage{listings}
\usepackage{soul}
\usepackage{subfig}
\usepackage{tabularx}
\usepackage{amsmath}
\renewcommand{\arraystretch}{1.075}
\def\HideTodo{0} %
\newcommand{\todo}[1]{\if\HideTodo0{\color{red} [todo: #1]}\fi}

\definecolor{mygreen}{rgb}{0,0.6,0}
\begin{document}

\title{Robotic Table Tennis: A Case Study\\into a High Speed Learning System}

\author{\authorblockN{David B.\ D'Ambrosio\authorrefmark{1}, Jonathan Abelian\authorrefmark{2}, Saminda Abeyruwan\authorrefmark{1}, Michael Ahn\authorrefmark{1}, Alex Bewley\authorrefmark{1}, \\ Justin Boyd\authorrefmark{2}, Krzysztof Choromanski\authorrefmark{1}, Omar Cortes\authorrefmark{2}, Erwin Coumans\authorrefmark{1}, Tianli Ding\authorrefmark{1}, Wenbo Gao\authorrefmark{1}, \\ Laura Graesser\authorrefmark{1}, Atil Iscen\authorrefmark{1}, Navdeep Jaitly\authorrefmark{1}, Deepali Jain\authorrefmark{1}, Juhana Kangaspunta\authorrefmark{1}, Satoshi Kataoka\authorrefmark{1}, \\ Gus Kouretas\authorrefmark{3}, Yuheng Kuang\authorrefmark{1}, Nevena Lazic\authorrefmark{1}, Corey Lynch\authorrefmark{1}, Reza Mahjourian\authorrefmark{1}, Sherry Q.\ Moore\authorrefmark{1}, \\ Thinh Nguyen\authorrefmark{2}, Ken Oslund\authorrefmark{1}, Barney J Reed\authorrefmark{4}, Krista Reymann\authorrefmark{1}, Pannag R.\ Sanketi\authorrefmark{1}, Anish Shankar\authorrefmark{1}, \\ Pierre Sermanet\authorrefmark{1}, Vikas Sindhwani\authorrefmark{1}, Avi Singh\authorrefmark{1}, Vincent Vanhoucke\authorrefmark{1}, Grace Vesom\authorrefmark{1}, and Peng Xu\authorrefmark{1} \\ Authors beyond the first are listed alphabetically, with full author contributions in the Appendix.}
\authorblockA{\authorrefmark{1}Google DeepMind.}
\authorblockA{\authorrefmark{2}Work done at Google DeepMind via FS Studio}
\authorblockA{\authorrefmark{3}Work done at Google DeepMind via Relentless Adrenaline}
\authorblockA{\authorrefmark{4}Work done at Google DeepMind via Stickman Skills Center LLC}}

\twocolumn[{%
\renewcommand\twocolumn[1][]{#1}%
\maketitle
\begin{center}
    \centering
    \captionsetup{type=figure}
    \includegraphics[width=\textwidth]{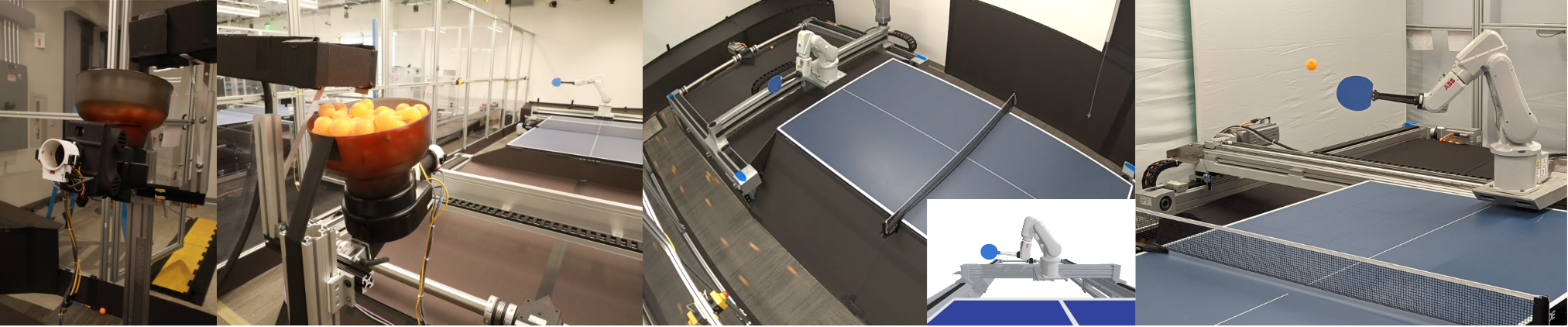}
    \captionof{figure}{The physical robotic table tennis system. Images from left to right show (I) ball thrower, (II) entire system (thrower, arm, gantry), (III) automatic ball refill, (inlay) simulator, and (IV) robot mid-swing.}
    \label{fig:main}
\end{center}%
}]

\begin{abstract}
We present a deep-dive into a real-world robotic learning system that, in previous work, was shown to be capable of hundreds of table tennis rallies with a human and has the ability to precisely return the ball to desired targets. This system puts together a highly optimized perception subsystem, a high-speed low-latency robot controller, a simulation paradigm that can prevent damage in the real world and also train policies for zero-shot transfer, and automated real world environment resets that enable autonomous training and evaluation on physical robots. We complement a complete system description, including numerous design decisions that are typically not widely disseminated, with a collection of studies that clarify the importance of mitigating various sources of latency, accounting for training and deployment distribution shifts, robustness of the perception system, sensitivity to policy hyper-parameters, and choice of action space. A video demonstrating the components of the system and details of experimental results can be found at \url{https://youtu.be/uFcnWjB42I0}.\footnote{Corresponding emails: \{bewley, ddambro, lauragraesser, psanketi\}@google.com.} 
\end{abstract}

\IEEEpeerreviewmaketitle

\section{Introduction}

There are some tasks that are infeasible for a robot to perform unless it moves and reacts quickly.  Industrial robots can execute pre-programmed motions at blindingly fast speeds, but planning, adapting, and learning while executing a task at high speed can push a robotic system to its limits and introduce complex safety and coordination challenges that may not show up in less demanding environments.  Yet many vital tasks, particularly those that involve interacting with humans in real time, necessitate such an \textit{high-speed robotic system}.

The goal of this paper is to describe such a system and the process behind its creation. Building any robotic system is a complex and multifaceted challenge, but nuanced design decisions are not often widely disseminated.  Our hope is that this paper can help researchers who are starting out in high-speed robotic learning and serve as a discussion point for those already active in the area.  %

We focus on a robotic table tennis system that has shown promise in playing with humans (340 hit cooperative rallies)  \cite{abeyruwan2022sim2real} and targeted ball returns (competitive with amateur humans) \cite{ding2022learning}.
This platform provides an excellent case study in system design because it includes multiple trade-offs and desiderata --- e.g. perception latency v.s. accuracy, ease of use v.s. performance, high speed, human interactivity, support for multiple learning methods --- and is able to produce strong real world performance.
This paper discusses the design decisions that went into the creation of the system and empirically validates many of them through analyses of key components.

\begin{figure*}[!t]
    \centering
    \includegraphics[width=0.45\textwidth]{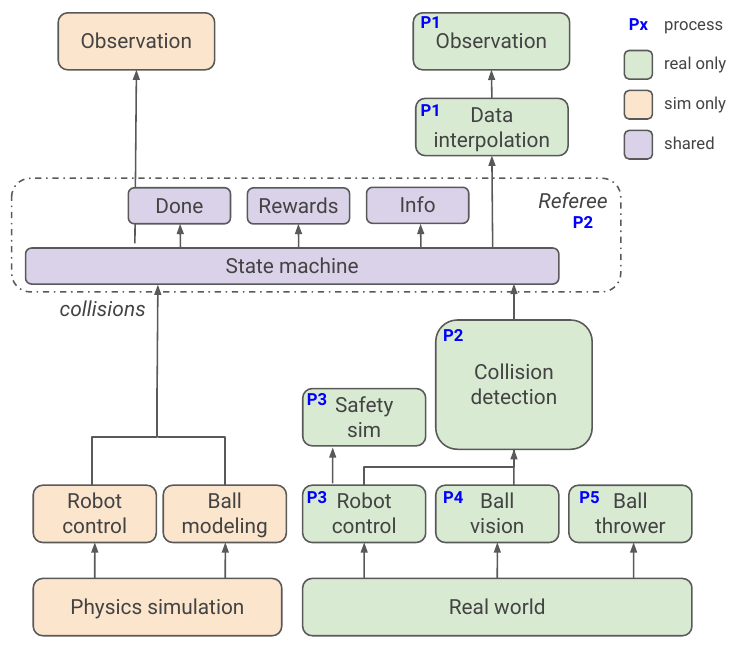}
    \includegraphics[width=0.45\textwidth]{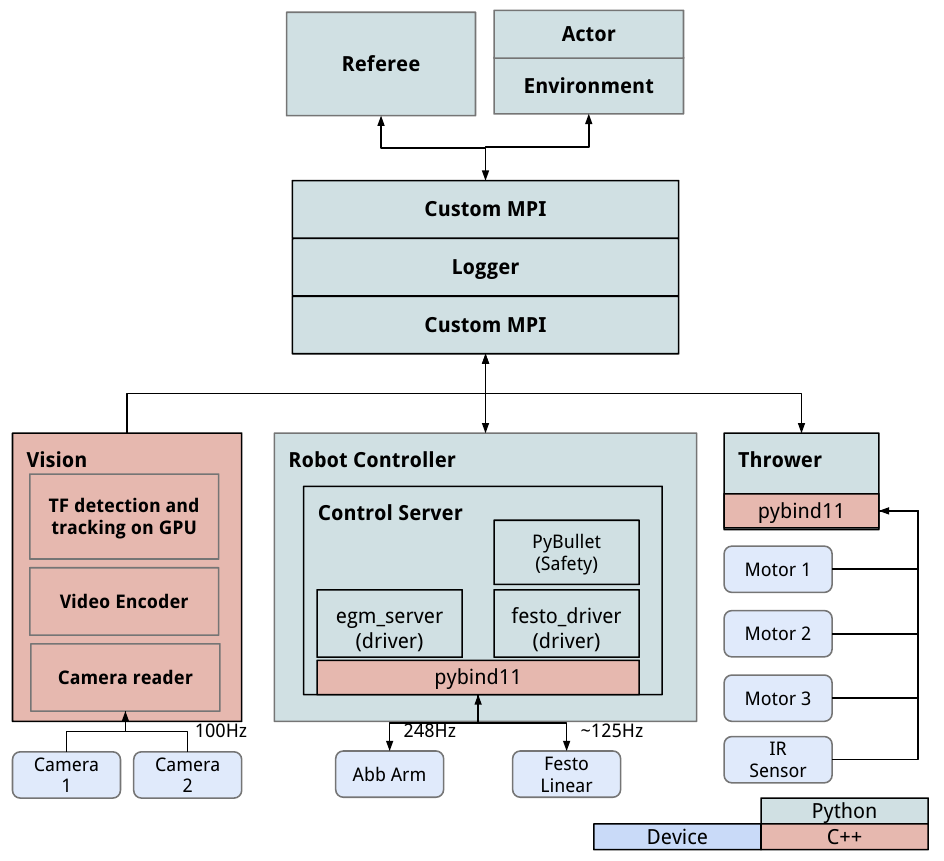}
  \caption{Overview of the components for running simulated and real environments. The diagram on the left shows how the various software components fit to form the environment: in simulation, everything runs in a single process, but the real environment splits the work among several. The diagram on the right shows the components of the real hardware system.  A custom MPI manages communication between the parts and logging of all data.}
    \label{fig:architecture}
\end{figure*}

This work explores all aspects of the system, how they relate to and inform one another, and highlights several important contributions including: (1) a highly optimized perception subsystem capable of running at 125Hz, (2) an example of high-speed, low latency control with industrial robots, (3) a simulation paradigm that can prevent damage in the real world while performing agile tasks and also train policies for zero-shot transfer using a variety of learning approaches, (4) a common interface for simulation and real world deployment, (5) an automatic physical environment reset system for table tennis that enables training and evaluation for long periods without human intervention, and (6) a research-friendly modular design that allows customization and component swapping. A summary of widely applicable lessons can be found in \autoref{sec:takeaways} and a video of the system in operation and experimental results can be found at \url{https://youtu.be/uFcnWjB42I0}.

\section{Table Tennis System}
\label{sec:the_system}

Table tennis is easy to pick up for humans, but poses interesting challenges for a robotic system.  Amateurs hit the ball at up to 9m/s, with professionals tripling that. Thus, the robot must be able to move, sense, and react quickly just to make contact, let alone replicate the precise hits needed for high-level play.

The components of this system are numerous with many interactions (\autoref{fig:architecture}). Therefore, a major design focus was on modularity to enable testing and swapping. At a high level, the hardware components (cameras + vision stack, robot, ball thrower) are controlled through C++ and communicate state to the environment through a custom message passing system called Fluxworks. The various components not only send policy-related information this way (e.g. where the the ball is, the position of the robot) but also synchronize the state of the system (e.g. the robot has faulted or a new episode has started). Note that this process is simplified in simulation where all state information is centralized. Information from the components determines the state of the game (in the Referee) and input to the policy. The policy then produces actions which feed into the low-level controllers while the game state drives the system as a whole (e.g. the episode is over). All logging (Appendix \ref{app:logging}), including videos, is handled with Fluxworks which utilizes highly optimized Protobuffer communication.

The rest of this section describes the components in the system and their dependencies and interactions.

\subsection{Physical Robots}
\label{sec:physical_robot}

The player in this system consists of two industrial robots that work together: an ABB 6DOF arm and a Festo 2DOF linear actuator, creating an 8DOF system (\autoref{fig:main}). The two robots complement each other: the gantry is able to cover large distances quickly, maneuvering the arm into an appropriate position where it can make fine adjustments and hit the ball in a controlled manner with the arm.  The choice of industrial robots was deliberate, to focus on the machine learning challenges of the problem and for high reliability.  However one major limitation of working with off-the-shelf industrial systems is that they may contain proprietary, ``closed-box" software that must be contended with.  For example, the ABB arm runs an additional safety layer that instantly stops the robot when it \textit{thinks} something bad will happen. It took careful effort to work within these constraints because the robot was operating near its limits.  See Appendix \ref{app:control} for details.

For the ABB arms, either an ABB IRB 120T or ABB IRB 1100-4/0.58 are used, the latter being a faster version with a different joint structure. Both are capable of fast (joints rotate up to 420 or 600 degrees/s), repeatable (to within 0.01mm) motions and allow a high control frequency. The arm's end effector is an 18.8cm 3D-printed extension attached to a standard table tennis paddle that has had its handle removed (\autoref{fig:main} right).  While the ABB arms are not perfect analogs to human arms, they can impart significant force and spin on the ball.  

Taking inspiration from professional table tennis where play can extend well to the side of and away from the table, the Festo gantries range in size from $2 \times 2$m to $4 \times 2$m, despite the table tennis table being 1.525m wide. This extra range gives the robot more options for returning the ball. The gantries can move up to 2 m/s in in both axes.  Most other robotic table tennis systems (discussed in \autoref{sec:robot_table_tennis}) opt for a fixed-position arm but the inclusion of a gantry means the robot is able to reach more of the table space and has more freedom to adopt general policies.  The downside is that the gantry complicates the system by adding two degrees of freedom leading to an overdetermined system whilst also imparting additional lateral forces on the robot arm that must be accounted for.

\subsection{Communication, Safety, and Control}
\label{sec:control}

The ABB robot accepts position and velocity target commands and provides joint feedback at 248Hz via the Externally Guided Motion (EGM) \cite{egmmanual} interface. The Festo gantry is controlled through a Modbus \cite{swales1999open} interface at approximately 125Hz. See Appendix \ref{app:control} for full communication details.

Safety is a critical component of controlling robots. While the robot should be hitting the ball, collision with anything else in the environment should be avoided. To solve this problem, commands are filtered through a safety simulator before being sent to the robot (a simplified version of \autoref{sec:simulator}). The simulator converts a velocity action generated by the control policy to a position and velocity command required by EGM at each timestep.  Collisions in the simulator generate a repulsive force that pushes the robot away, resulting in a valid, safe command for the real robot. Objects in the safety simulator are dilated for an adequate safety margin and additional obstacles are added to block off the ``danger zones'' robot should avoid.

Low-level robot control can be extremely time-sensitive and is typically implemented in a lower-level language like C++ for performance. Python on the other hand is very useful for high-level machine learning implementations and rapid iteration but is not well suited to high speed robot control due to the Global Interpreter Lock (GIL) which severely hampers concurrency. This limitation can be mitigated through multiple Python processes, but is still not optimal for speed. Therefore this system adopts a hybrid approach where latency sensitive processes like control and perception are implemented in C++ while others are partitioned into several Python binaries (\autoref{fig:architecture}). Having these components in Python allows researchers to iterate rapidly and not worry as much about low-level details. This separation also allows components to be easily swapped or tested.

\subsection{Simulator}
\label{sec:simulator}

The table tennis environment is simulated to facilitate sim-to-real training and prototyping for real robot training. PyBullet \cite{coumans2021} is the physics engine and the environment interface conforms to the Gym API \cite{brockman2016openai}.

\autoref{fig:architecture} (left) gives an overview of the environment structure in simulation and compares it with the real world environment (see \autoref{sec:real_env}). There are five conceptual components; (1) the physics simulation and ball dynamics model which together model the dynamics of the robot and ball, (2) the \texttt{StateMachine} which uses ball contact information from the physics simulation and tracks the semantic state of the game (e.g. the ball just bounced on the opponent's side of the table, the player hit the ball), (3) the \texttt{RewardManager} which loads a configurable set of rewards and outputs the reward per step, (4) the \texttt{DoneManager} which loads a configurable set of done conditions (e.g. ball leaves play area, robot collision with non-ball object) and outputs if the episode is done per step, and (5) the \texttt{Observation} class which configurably formats the environment observation per step.

The main advantage of this design is that it isolates components so they are easy to build and iterate on. For example, the \texttt{StateMachine} makes it easy to extend the environment to more complex tasks. New tasks are defined by implementing a new state machine in a config file. The \texttt{StateMachine} also makes it easier to determine the episode termination condition and some rewards (e.g. for hitting the ball). Note that whilst related, it is not the same as the transition function of the MDP; the \texttt{StateMachine} is less granular and changes at a lower frequency. Another example is the \texttt{RewardManager}. It is common practice in robot learning when training using the reinforcement learning paradigm to experiment frequently with the reward function. To facilitate this, reward components and their weights are specified in a config file taken in by the \texttt{RewardManager}, which calculates and sums each component. This makes it straightforward to change rewards and easy to define new components.

\begin{table}[!b]
\centering
\scriptsize
\begin{tabular}{|| l | c c || }
\hline
\multicolumn{1}{|| l |}{}  & \multicolumn{2}{c ||}{Latencies (ms)} \\
Component & $\mu$ & $\sigma$  \\
\hline \hline
Ball observation & 40 & 8.2  \\
ABB observation & 29 & 8.2  \\
Festo observation & 33 & 9.0  \\
ABB action & 71 & 5.7  \\
Festo action & 64.5 & 11.5  \\
\hline
\end{tabular}
\caption{Latency distribution values.}
\label{table:latency-distribution}
\end{table}

\subsubsection{Latency modeling} Latency is a major source of the sim-to-real gap in robotics \citep{Tan18}.
To mitigate this issue, and inspired by \citet{Tan18}, latency is modelled in the simulation as follows. During inference, the history of observations and corresponding timestamps are stored and linearly interpolated to produce an observation with a desired latency.
In contrast to \cite{Tan18} which uses a single latency range sampled uniformly for the whole observation, the latency of five main components --- Ball observation (i.e. latency of the ball perception system), ABB observation, Festo observation, ABB action, Festo action --- are modeled as a Gaussian distribution and a distinct distribution is used for each component. The mean and standard deviation per component were measured empirically on the physical system through instrumentation that logs timestamps throughout the software stack (see \autoref{table:latency-distribution}). In simulation, at the beginning of each episode a latency value is sampled per component and the observation components are interpolated to those latency values per step. Similarly, action latency is implemented by storing the raw actions produced by the policy in a buffer, and linearly interpolating the action sent to the robot to the desired latency.

\subsubsection{Ball distributions, observation noise, and domain randomization} A table tennis player must be able to return balls with many different incoming trajectories and angular velocities. That is, they experience different \textit{ball distributions}. Ball dynamics and distributions are implemented following \cite{abeyruwan2022sim2real}. Each episode, initial ball conditions are sampled from a parameterized distribution which is specified in a config. To account for real world jitter, random noise is added to the ball observation. Domain randomization \citep{Peng18, Chebotar19, Lee19, rubikscube} is also supported for many physical parameters. The paddle and table restitution coefficients are randomized by default.

For more details on the simulator see Appendix \ref{app:simulator}.

\subsection{Perception System}
\label{sec:perception}

Table tennis is a highly dynamic sport (an amateur-speed ball crosses the table in 0.4 seconds), requiring extremely fast reaction times and precise motor control when hitting the ball. Therefore a vision system with the desiderata of low latency and high precision is required. It is also not possible to instrument (e.g. with LEDs) or paint the ball for active tracking as they are very sensitive to variation in weight or texture and so a passive vision system must be employed.

A custom vision pipeline that is fast, accurate and passive is designed to provide 3D balls positions. It consists of three main components 1) 2D ball detection across two stereo cameras, 2) triangulation to recover the 3D ball position and 3) a sequential decision making process which manages trajectory creation, filtering, and termination. The remainder of this section will provide details on the hardware and these components.

\begin{figure}[!t]
    \centering
    \includegraphics[width=\linewidth]{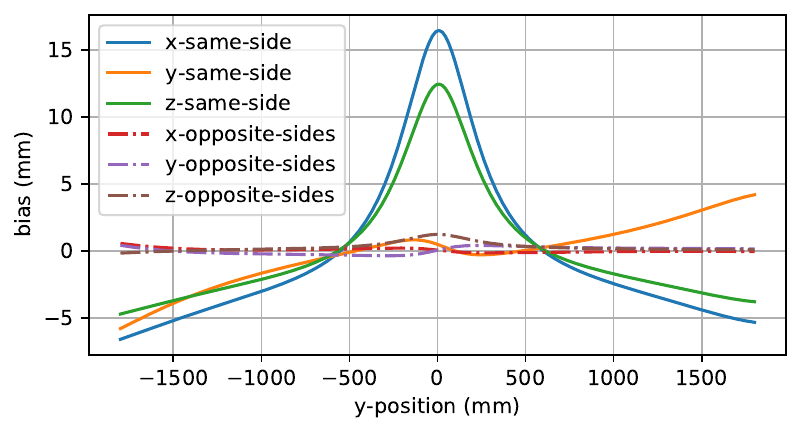}
  \caption{Quantification of triangulation bias over the length of playing area (y-position) at a height of 250mm above the center line. The more orthogonal viewpoints offered by placing cameras on opposite sides of the tables lead to an order of magnitude reduction in triangulation bias.}
    \label{fig:triangulation_bias}
\end{figure}

\subsubsection{Camera Hardware, Synchronization and Setup}

For image capture the system employs a pair of Ximea MQ013CG-ON cameras that have a hardwired synchronization cable and are connected to the host computer via USB3 active optical cables.  Cameras lenses are firmly locked and focused.  Synchronization timestamps are used to match images downstream. Many different cameras were tried, but these had high frame rates (the cameras can run at 125FPS at a resolution of 1280x1024) and an extremely low latency of 838$\mu$s.
Other cameras were capable of higher FPS, at the cost of more latency which is not acceptable in this high-speed domain. To achieve the desired performance the camera uses a global shutter with a short (4ms) exposure time and only returns the raw, unprocessed Bayer pattern.

The ball is small and moves fast, so capturing it accurately is a challenge.  Ideally the cameras would be as close to the action as possible, but in a dual camera setup, each needs to view the entire play area.  Additionally, putting sensitively calibrated cameras in the path of fast moving balls is not ideal.  Instead, the cameras are mounted roughly 2m above the play area on each side of the table and are equipped with Fujinon FE185C086HA-1 ``fisheye'' lenses that expand the view to the full play area, including the gantries. While capturing more of the environment, the fisheye lens distortion introduces challenges in calibration and additional uncertainty in triangulation.

The direct linear transform (DLT) method \cite{hartley2003multiple} for binocular stereo vision estimates a 3D position from these image locations in the table's coordinate frame.
However, the problem of non-uniform and non-zero mean bias known as triangulation bias \cite{freundlich2015exact} must be considered in optimizing camera placement. %
Two stereo camera configurations are considered, two overhead cameras viewing the scene from: 1) the same side of the table and 2) opposite sides. Simulation is used to quantify triangulation bias across these configurations and decouple triangulation from potential errors in calibration. 
Quantifying this bias for common ball positions (see \autoref{fig:triangulation_bias}) indicates that positioning the cameras on opposite table sides results in a significant reduction in the overall triangulation bias. Furthermore, this configuration also benefits from a larger baseline between the cameras for reducing estimation variance \citep{gallup2008variable}.

\subsubsection{Ball Detection}

\begin{figure}
    \centering
    \includegraphics[width=0.45\linewidth]{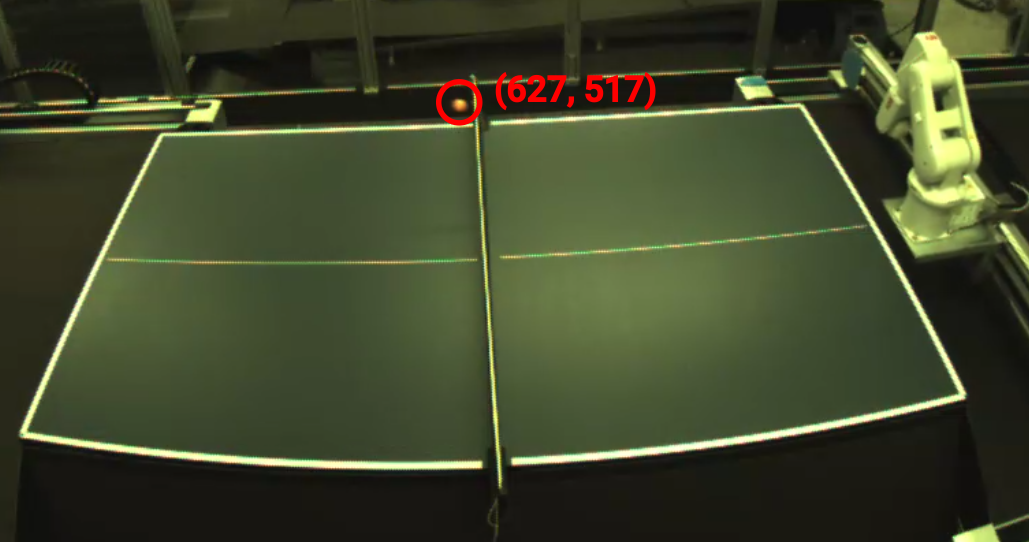}
    \includegraphics[width=0.45\linewidth]{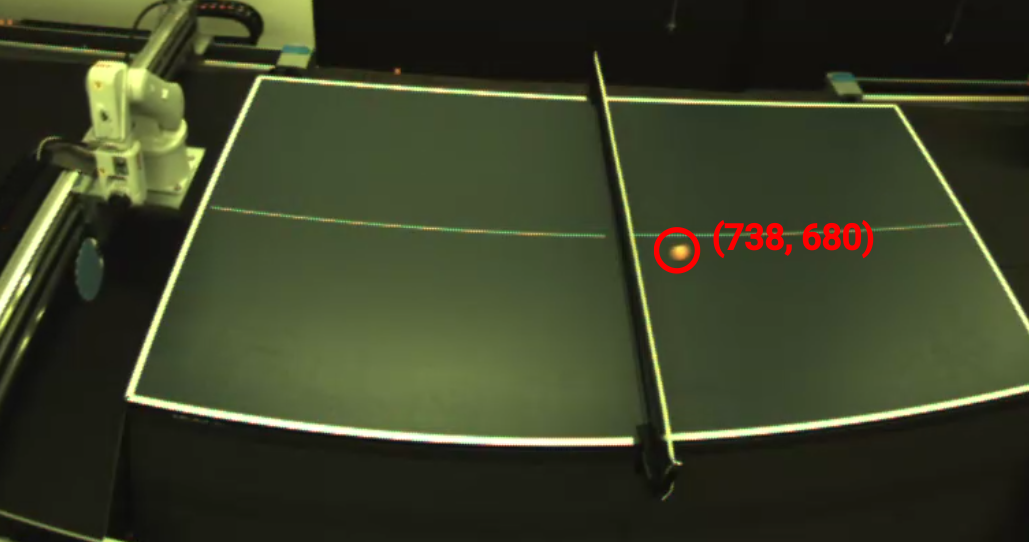}
  \caption{Ball Detection. These synchronized images (cropped to approximately 50\% normal size) show the temporal convolutional network detecting the ball (detected ball center in pixels) independently from cameras on both sides of the table. These detections are triangulated and used for 3D tracking. }
    \label{fig:ball_detection}
\end{figure}

The core of the perception system lies with ball detection. The system uses a temporal convolutional architecture to process each camera's video stream independently and provides information about the ball location and velocity for the downstream triangulation and filtering (see \autoref{fig:ball_detection}). The system uses raw Bayer images and temporal convolutions, which allow it to efficiently process each video stream independently and thus improve the latency and accuracy of ball detection. The output structure takes inspiration from CenterNet \cite{zhou2019objects, zhou2020tracking} by producing per location predictions that include: a ball score indicating corresponding to the likelihood of the ball center at that location, a 2D local offset to accommodate sub-pixel resolution, and a 2D estimate of the ball velocity in pixels.

\paragraph{Direct Processing of Bayer Images}

The detection network takes the raw Bayer pattern image \cite{bayer1976color} as input directly from the high speed camera after cropping to the play area at a resolution of $ 512 \times 1024 $.
By skipping Bayer to RGB conversion, 1ms (or 15\% of the time between images) of conversion induced latency per camera is avoided and data transferred from camera to host to accelerator is reduced by $\frac{2}{3}$, further reducing latency. In contrast to other models utilizing Bayer images \cite{Chandra21Bayer}, no loss in performance was found using the raw format, largely due to special attention given to structure of the $2 \times 2$ Bayer pattern and ensuring the first convolution layer is also set to have a stride of $ 2 \times 2 $. This alignment means that the individual weights of the first layer are only responsible for a single color across all positions of the convolution operation.  The immediate striding also benefits wall-clock time by down-sampling the input to a quarter of the original size. The alignment with the Bayer pattern is also extended to any crop operations during training as discussed later in this section.

\paragraph{Detector Backbone with Buffered Temporal Convolutions}

A custom deep-learning based ball detector is used to learn the right combination of color, shape and motion for identifying the ball in play.
Its architecture falls in the category of a convolutional neural network (CNN) with a compact size of only 27k parameters spread over five spatial convolutional layers and two temporal convolutions to capture motion features. Compared to related architectures such as ConvLSTM \citep{shi2015convolutional}, this fully convolutional approach restricts the temporal influence of the predictions to a finite temporal window allowing for greater interpretability and fault diagnosis. Full details of the architecture are provided in Appendix \ref{app:perception}. %

Temporal convolutional operations are employed to capture motion as a visual cue for detecting the ball in play and the direction of motion. In contrast to the typical implementation that requires a window of frames to be presented at each timestep, the implementation in this system only requires a single frame to be presented to the CNN for each timestep during inference. This change minimises data transfer from the host device to the accelerator running the CNN operations, a critical throughput bottleneck. This temporal layer creates a buffer to store the input feature for the next timestep as in \citet{Khandelwal21RT-TCN}.

\paragraph{Training the Detector Model}

To train the detection model, a dataset of 2.3M small temporal patches were selected to match the receptive field of the architecture ($64 \times 64$ pixels and $n$ frames). The patches are selected from frames with a labeled ball position where a single \emph{positive patch} is defined as being centered on the ball position in the current frame with the temporal dimension filled with the same spatial position but spanning $[t-n+1, t]$. Similarly a \emph{negative patch} can be selected from the same frame at a random location which does not overlap with the positive patch. Examples of positive and negative patches are provided in the Appendix. Special consideration is taken to align the patch selection with the Bayer pattern by rounding the patch location to the nearest even number.
This local patch based training has several benefits; it 1) reduces the training time by 50$\times$\footnote{Two $ 64 \times 64 \times n $ patches are required per frame as opposed to the full $ 512 \times 1024 \times n$ frames. }, 2) helps generalization across different parts of the image as the model is unable to rely on global statistics of ball positions, 3) offers a more fine-grained selection of training data for non-trivial cases e.g. when another ball is still moving in the scene, and similarly 4) allows for hard negative mining \citep{sung1996learning} on sequences where it is known for no ball to exist in play.

For each patch the separate outputs each have a corresponding loss. First, the ball score is optimized using the standard binary cross-entropy loss for both positive and negative patches. For positive patches only, the local offset is optimized using the mean-squared error loss using the relative position between the corresponding pixel coordinate and the ball center in the current frame. The velocity prediction is similarly optimized, instead using the relative position of the ball in next frame to the current frame as the regression target.

\subsubsection{3D Tracking}
To have a consistent representation that is invariant to camera viewpoint, the ball is represented in 3D in the table's coordinate frame. If the maximum score in both images are above a learnt threshold, their current and next image positions using the local offset and velocity predictions are triangulated using DLT \cite{hartley2003multiple}. This corresponds to the 3D position and 3D velocity of the ball in the table frame. Finally these observations are provided to a recursive Kalman filter \citep{kalman1960filter} to refine the estimated ball state before its 3D position is sent to the robot policy.

\subsection{Running on the Real Robot}
\label{sec:real_env}

As an analog to the simulated environment (\autoref{sec:simulator}) there is an equivalent Gym environment for the real hardware. This environment must contend with an additional set of challenges that are either nonexistent or trivial in simulation: 1) continuous raw sensor observation at different frequencies that is subjected to jitter and real world noise, 2) determining the start of an episode, 3) monitoring environment state, 4) environment resets.

\subsubsection{Observation generation}

In the simulator, the state of every object is known and can be queried at fixed intervals. In contrast, the real environment receives sensor readings from different modalities at different frequencies (e.g. the ball, ABB, Festo) that may be inaccurate or arrive irregularly. To generate policy observations, the sensor observations, along with their timestamps are buffered and interpolated or extrapolated to the environment step timestamp. To address noise and jitter a bandpass filter is applied to the observation buffer before interpolation (see Appendix \ref{app:environment}). These observations are afterwards converted according to the policy observation specification. 

\subsubsection{Episode Starts}

Simulators provide a direct function to reset the environment to a start state instantly. In the real world, the robot must be physically moved to a start state with controllers based on standard S-curve trajectory planning at the end of the episode or just after a paddle hit. The latter was shown to be beneficial in \cite{abeyruwan2022sim2real}, so that a human and robot could interact as fast as possible. An episode starts when a valid ball is thrown towards the robot. The real world must rely on vision to detect this event and can be subject to spurious false positives, balls rolling on the table, bad ball throws, etc., which need to be taken into consideration. Therefore an episode is started only if a ball is detected incoming toward a robot from a predefined region of space.

\subsubsection{Referee}

To interface with the GymAPI a process called \textit{Referee} generates the reward, done, and info using the \texttt{StateMachine}, \texttt{RewardManager}, and \texttt{DoneManager} as defined in \autoref{sec:simulator}. It receives raw sensor observations at different frequencies and updates a PyBullet instance. The observations are filtered (see Appendix \ref{app:environment}) and used to update the PyBullet state (only the position). It calculates different ball contact events (see Appendix \ref{app:simulator}), compensates for false positives, and uses simple heuristics and closest point thresholds to determine high confidence ball contact detections to generate the events used by the previously mentioned components.

\subsubsection{Automatic system reset --- continuously introducing balls}
An important aspect of a real world robotic system is environment reset. If each episode requires a lengthy reset process or human intervention, then progress will be slow.  
Human table tennis players also face this problem and so-called ``table tennis robots'' are commercially available to shoot balls continuously and even in a variety of programmed ways.  Almost all of these machines accomplish this task with a hopper of balls that introduces a ball to two or more rotating wheels forcing it out at a desired speed and spin (see \autoref{fig:main} left).  Unfortunately, while many of these devices are ``programmable'', none provide true APIs and instead rely on physical interfaces.  Therefore, an off-the-shelf thrower was customized with a Pololu motor controller and an infrared sensor for detecting throws, allowing it to be controlled over USB.  This setup allows balls to be introduced purely through software control.  

However, the ball thrower is still limited by the hopper capacity.  A system to automate the refill process was designed that exploits the light weight of table tennis balls by blowing air to return them to the hopper.  A ceiling-mounted fan blows down to remove balls stuck on the table, which is surrounded by foamcore to direct the balls into carpeted pathways. At each corner of the path is a blower fan (typically meant for drying out carpet) that directs air across the floor. The balls circulate around the table until they reach a ramp that directs them to a tube that also uses air to transport them back into the hopper. When the thrower detects it hasn't shot a ball for a while, the fans turn on for 40 seconds, refilling the hopper so training or evaluation can continue indefinitely.  See Appendix \ref{app:environment} for a diagram and the video at \url{https://youtu.be/uFcnWjB42I0} for a demonstration.

One demonstration of the utility of this system is through the experiments in this paper. For example, the simulator parameter ablation studies (\autoref{ablations:sim-params}) involved evaluating over 150 policies in 450+ independent evaluations on a physical robot with 22.5k+ balls thrown. All evaluations were conducted remotely and required onsite intervention just once\footnote{Some tape became unstuck and the balls escaped.}.

\subsection{Design of Robot Policies}

 Policies have been trained for this system using a variety of approaches.  This section details the basic structure of these policies and any customization needed for specific methods.

\subsubsection{Policies}
The policy input consists of a history of the past eight robot joint and ball states, and it outputs the desired robot state, typically a velocity for each of the eight joints (joint space policies).
Many robot control frequencies ranging from from 20Hz - 100Hz have been explored, but 100Hz is used for most experiments. Most policies are compact, represented as a three layer, 1D, fully convolutional gated dilated CNN with $\approx$1k parameters introduced in \cite{GaoPPOES2020}. However, it is also possible to deploy larger policies. For example, a 13m parameter policy consisting of two LSTM layers with a fully connected output layer has successfully controlled the robot at 60Hz \cite{ding2022learning}.

\subsubsection{Robot Policies in Task Space}
\label{sec:task_space}
Joint space policies lack the relation between joint movement and the task at hand. A more compact task space --- the pose of the robot end effector --- is especially beneficial in in robotics, showing significant improvements in learning of locomotion and manipulation tasks \citep{Duan2021,martin2019iros,Varin2019iros,Luo2019icra}.

Standard task space control uses the Jacobian Matrix to calculate joint torques or velocities given target pose, target end effector velocities, joint angles and joint velocities. This system employs a reduced (pitch invariant) version with 5 dimensions. Instead of commanding the full pose of the end effector, it commands the position in 3 dimensions and the surface normal of the paddle in 2 dimensions (roll and yaw).
In contrast to the default joint space policies, which use velocity control, task space policies are position controlled, which have the added benefit of easily defining a bounding cube that the paddle should operate in.
The robot state component of the observation space is also represented in task space, making policies independent of a robot's form factor and enabling transfer of learned policies across different robots (see Section \ref{sec:task_space_results}).  

\subsection{Blackbox Gradient Sensing (BGS)}
\label{sec:sim_training}

The design of the system allows for interaction with many different learning approaches, as long as they conform to the given APIs. The system supports training using a variety of methods including BGS \cite{abeyruwan2022sim2real} (evolutionary strategies), PPO \cite{ppo} and SAC \cite{sac-18} (reinforcement learning), and GoalsEye (behavior cloning). The rest of the section describes BGS, since it is used as the training algorithm in all the system studies in this paper (see \autoref{sec:system-studies}).

BGS is an ES algorithm. This class of algorithm maximize a smoothed version of expected episode return, $\mathcal{R}$, given by:
\begin{equation}
    \mathcal{R}_{\sigma}(\theta) = \mathbb{E}_{\mathbf{\delta} \sim \mathcal{N}(0,\mathbf{I}_{d})}[\mathcal{R}(\theta+\sigma \mathbf{\delta})]
\end{equation}
where $\sigma > 0$ controls the precision of the smoothing, and $\delta$ is a random normal perturbation vector with the same dimension as the policy parameters $\theta$.
$\theta$ is perturbed by adding or subtracting $N$ Gaussian perturbations $\delta_{R_i}$ and calculating episode return, $R^{+}_i$ and $R^{-}_i$ for each direction. 
Assuming the perturbations, $\delta_{R_i}$, are rank ordered with $\delta_{R_1}$ being the top performing direction, then the policy update can be expressed as:
\begin{equation}
    \theta^{'} = \theta + \alpha \frac{1}{\sigma^R} \sum^{k}_{i=1} \Bigg[\Big( \Big( \frac{1}{m} \sum^{m}_{j=1} R^{+}_{i,j}\Big) - \Big(\frac{1}{m} \sum^{m}_{j=1} R^{-}_{i,j}\Big)\Big) \delta_{R_i}\Bigg]
\end{equation}
where $\alpha$ is the step size, $\sigma^R$ is the standard deviation of each distinct reward (positive and negative direction), $N$ is the number of directions sampled per parameter update, and $k (< N)$ is the number of top directions (elites). $m$ is the number of repeats per direction to reduce variance for reward estimation. $R^{+}_{i,j}$ is the reward corresponding to the j-th repeat of i-th in the positive direction. $R^{-}_{i,j}$ is the same but in the negative direction.

BGS is an improvement upon a popular ES algorithm ARS~\cite{ars}, with two major changes.

\subsubsection{Reward differential elite-choice.} In ARS, rewards are ranked yielding an ordering of directions based on the absolute rewards of either the positive or negative directions. BGS takes the absolute difference in rewards between the positive and negative directions and rank the differences to yield an ordering over directions. ARS can be interpreted as ranking directions in absolute reward space, whereas BGS ranks directions according to reward curvature:
\begin{align}
&\text{ARS: Sort } \delta_{R_i} \text{ by max}\{R^{+}_{i}, R^{-}_{i}\}. \\
&\text{BGS: Sort } \delta_{R_i} \text{ by } |R^{+}_{i} - R^{-}_{i}|.
\end{align}

\subsubsection{Orthogonal sampling} Orthogonal ensembles of perturbations $\delta_{R_{i}}$ \citep{ICML-2018-ChoromanskiRSTW} relies on constructing perturbations $\delta_{R_{i}}$ in blocks, where each block consists of pairwise orthogonal samples. Those samples are still of Gaussian marginal distributions, matching those of the regular non-orthogonal variant. The feasibility of such a construction comes from the isotropic property of the Gaussian distribution (see: \cite{ICML-2018-ChoromanskiRSTW} for details).

BGS policies are trained in simulation and transferred zero-shot to the physical hardware.
An important note is that the BGS framework can also fine tune policies on hardware through the real Gym API (\autoref{sec:real_env}). Hyperparameters must be adjusted in this case to account for there only being one ``worker'' to gather samples.

\section{System Studies}
\label{sec:system-studies}

This section describes several experiments that explore and evaluate the importance of the various components of the system.

Except where noted, the experiments use a ball return task for training and testing. A ball is launched towards the robot such that it bounces on the robot's side of the table (a standard rule in table tennis). The robot must then hit the ball back over the net so it lands on the opposite side of the table. Although other work has applied this system to more complex tasks (e.g. cooperative human rallies \cite{abeyruwan2022sim2real}), a simpler task isolates the variables we are interested in from complications like variability and repeatability of humans.

For real robot evaluations, making contact with the ball is worth one point and landing on the opposing side is worth another point, for a maximum episode return of 2.0. A single evaluation is the average return over 50 episodes. Simulated training runs typically have additional reward shaping applied that change the maximum episode return to 4.0 (see Appendix \ref{app:simulator}).

\subsection{Effect of Simulation Parameters on Zero-Shot Transfer}
\label{ablations:sim-params}

\begin{figure}[!t]
    \centering
    \includegraphics[width=0.24\textwidth]{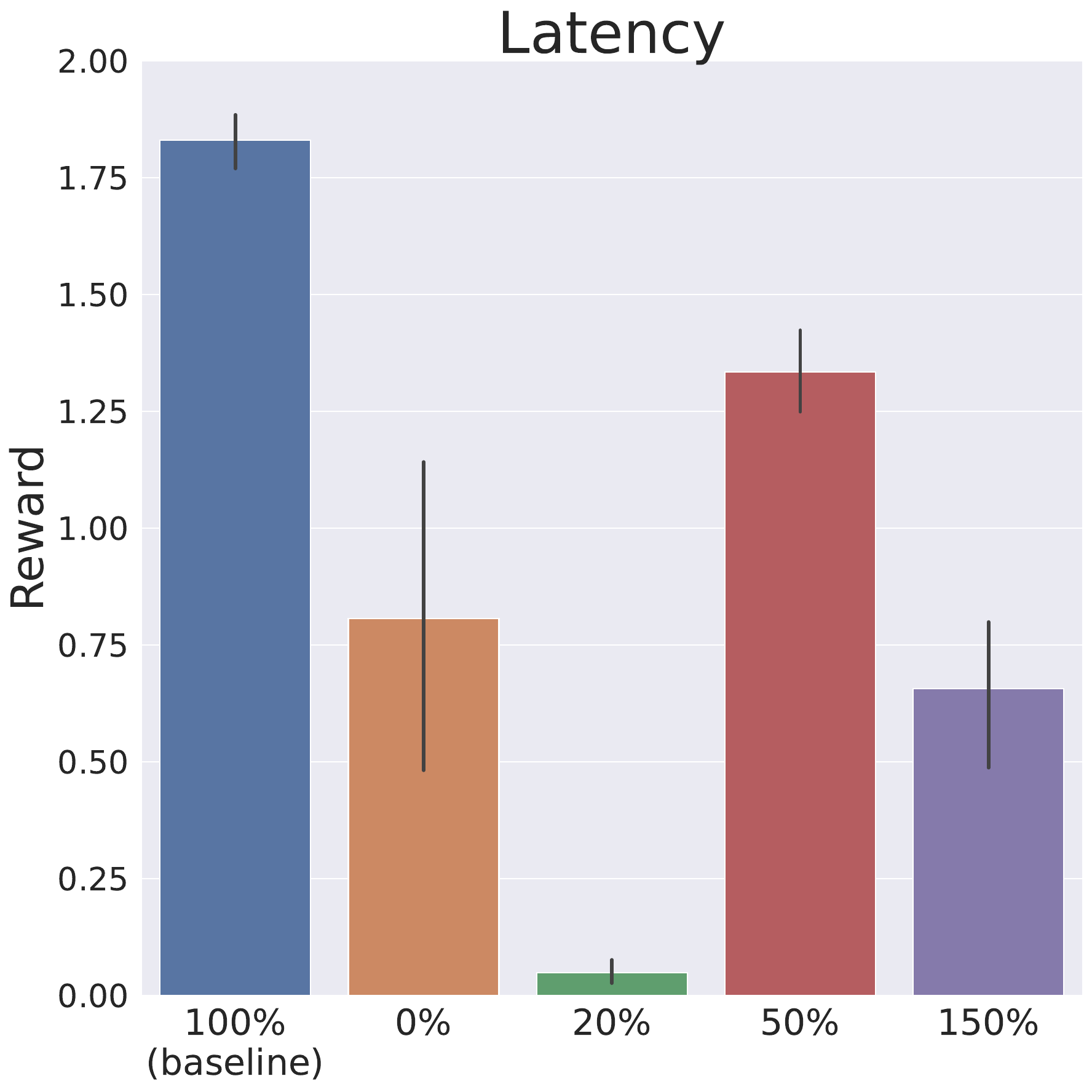}
    \includegraphics[width=0.24\textwidth]{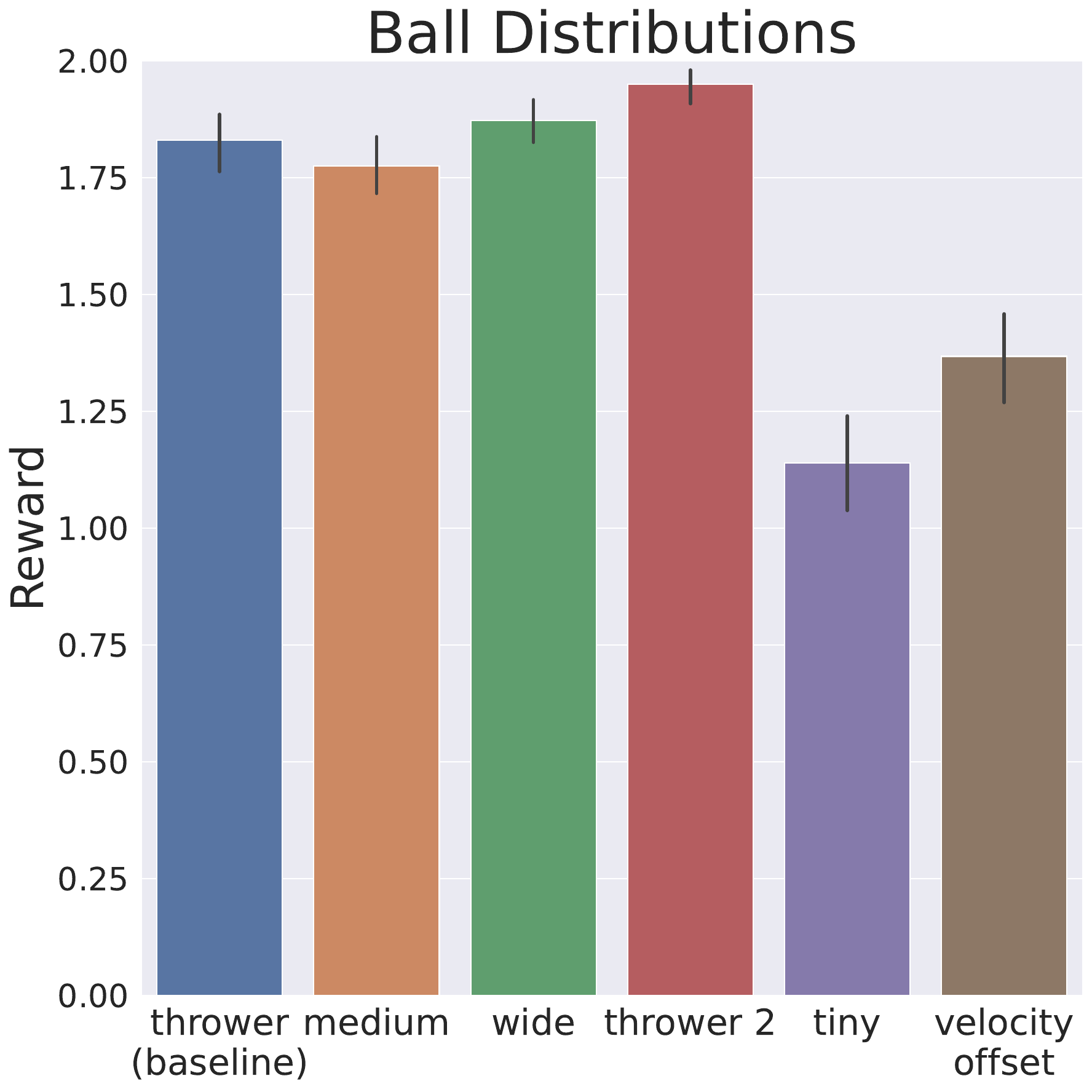}
    \includegraphics[width=0.24\textwidth]{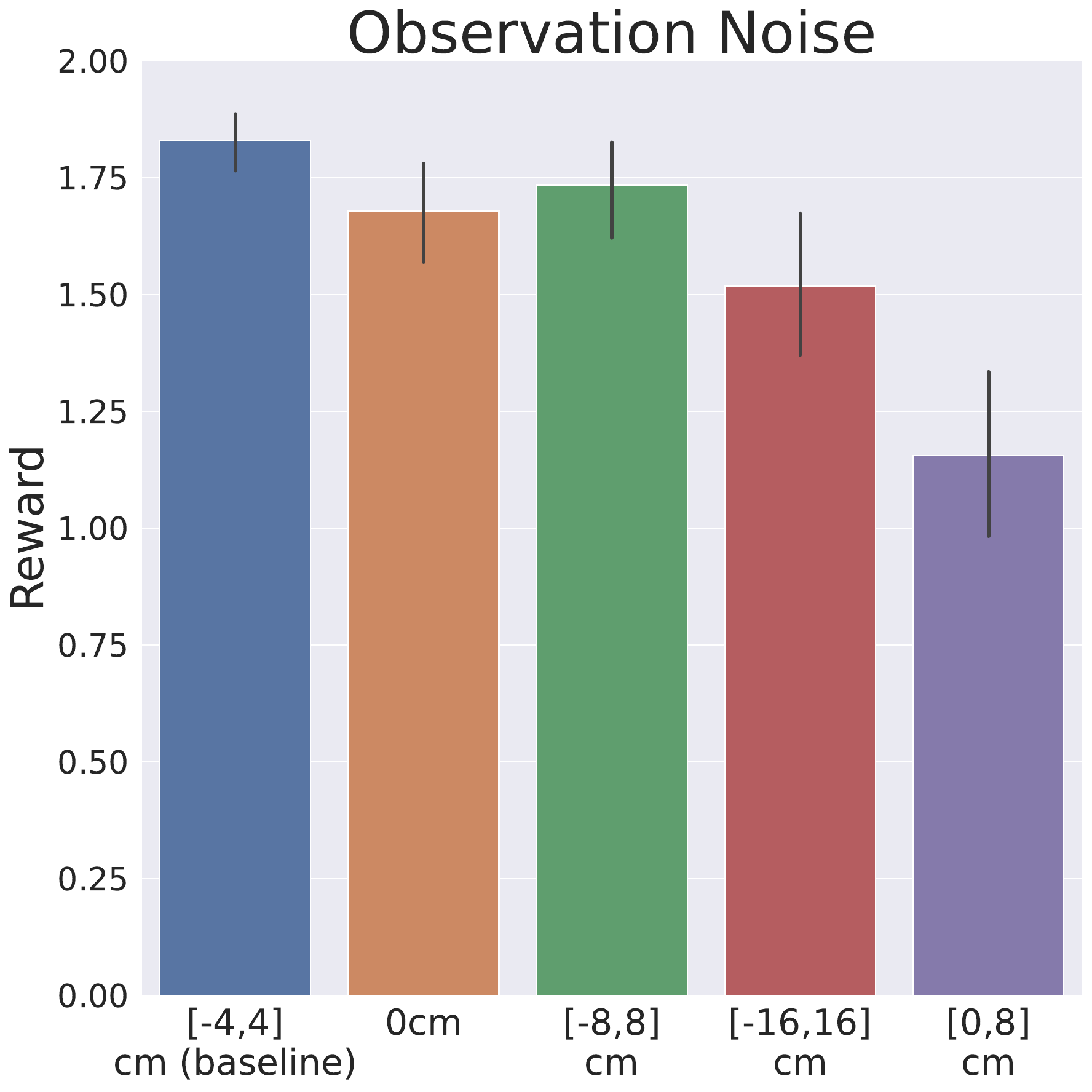}
    \includegraphics[width=0.24\textwidth]{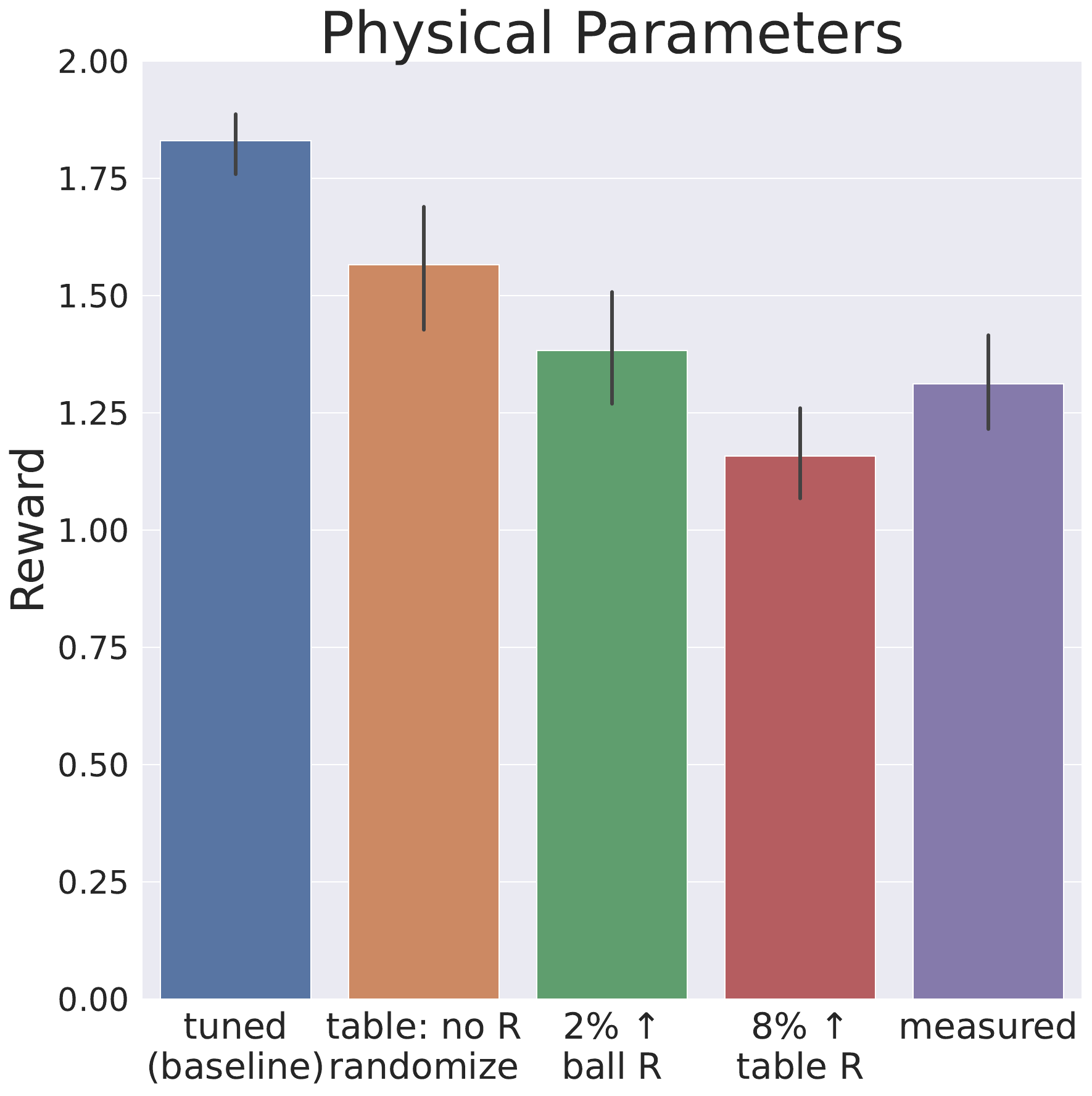}
    \caption{Effect of simulator parameters on zero-shot sim-to-real transfer. Policies are sensitive to latency and physical parameter values, yet surprisingly robust to ball observation noise and changes in the ball distribution. Charts show the mean (with 95\% CIs) zero-shot sim-to-real transfer. 2.0 is a perfect score with a policy returning all balls. R = restitution coefficient.}
    \label{fig:sim-params-ablations}
\end{figure}

Our goal in this section is to assess the sensitivity of policy performance to environment parameters. We focus on the zero-shot sim-to-real performance of trained policies and hope that this analysis (presented in \autoref{fig:sim-params-ablations}) sheds some light on which aspects of similar systems need to be faithfully aligned with the real world and where error can be tolerated. For the effects on training quality see Appendix \ref{app:sim-ablations-details}.

\subsubsection{Evaluation methodology}
For each test in this section, 10 models were trained in simulation using BGS described in \autoref{sec:sim_training} for 10,000 training iterations (equivalent to 60m environment episodes, or roughly 6B environment steps).
In order to assess how different simulated training settings affect transfer independent of how they affect training quality, we only evaluate models that trained well in simulation (i.e., achieved more than 97.5\% of the maximum possible return). The resulting set of policies were evaluated on the real setup for 3 $\times$ 50 episodes.

\subsubsection{Modeling latency is crucial for good performance} The latency study presented in \autoref{fig:sim-params-ablations} (top left) show that policies are sensitive to latency. The baseline model (i.e. the model that uses latency values as measured on hardware) had a significantly higher zero-shot transfer than any of the other latency values tested. The next best model had 50\% of the baseline latency, achieving an average zero-shot transfer of 1.33 compared with 1.83 for the baseline. Zero-shot transfer scores for the other latency levels tested (0\%, 20\% and 150\%) had very poor performance. Interestingly, some policies are lucky and transfer relatively well --- for example one policy with 0\% latency had an average score of 1.54. However, performance is highly inconsistent when simulated latency is different from measured parameters.

\subsubsection{Anchoring ball distributions to the real world matters, but precision is not essential} The ball distribution study shown in \autoref{fig:sim-params-ablations} (top right) indicate that policies are robust to variations in ball distributions provided the real world distribution (thrower) is contained within the training distribution. The medium and wide distributions were derived from the baseline distribution but are 25\% and 100\% larger respectively (see Appendix \ref{app:sim-ablations-details}). The distribution derived from a different ball thrower (thrower 2) is also larger than the baseline thrower distribution but effectively contains it. In contrast, very small training distributions (tiny) or distributions which are disjoint from the baseline distribution in one or more components (velocity offset --- disjoint in y velocity) result in performance degradation.

\subsubsection{Policies are robust to observation noise provided it has zero mean} The observation noise study in \autoref{fig:sim-params-ablations} (bottom left) revealed that policies have a high tolerance for zero-mean observation noise. Doubling the noise to +/- 8cm (4 ball diameters in total) or removing it altogether had a minor impact on performance. However, if noise is biased performance suffers substantially. Adding a 4cm (one ball diameter) bias to the default noise results in a 36\% drop in reward (approximately 80\% drop in return rate).

\begin{figure}[!t]
    \centering
    \includegraphics[width=0.24\textwidth]{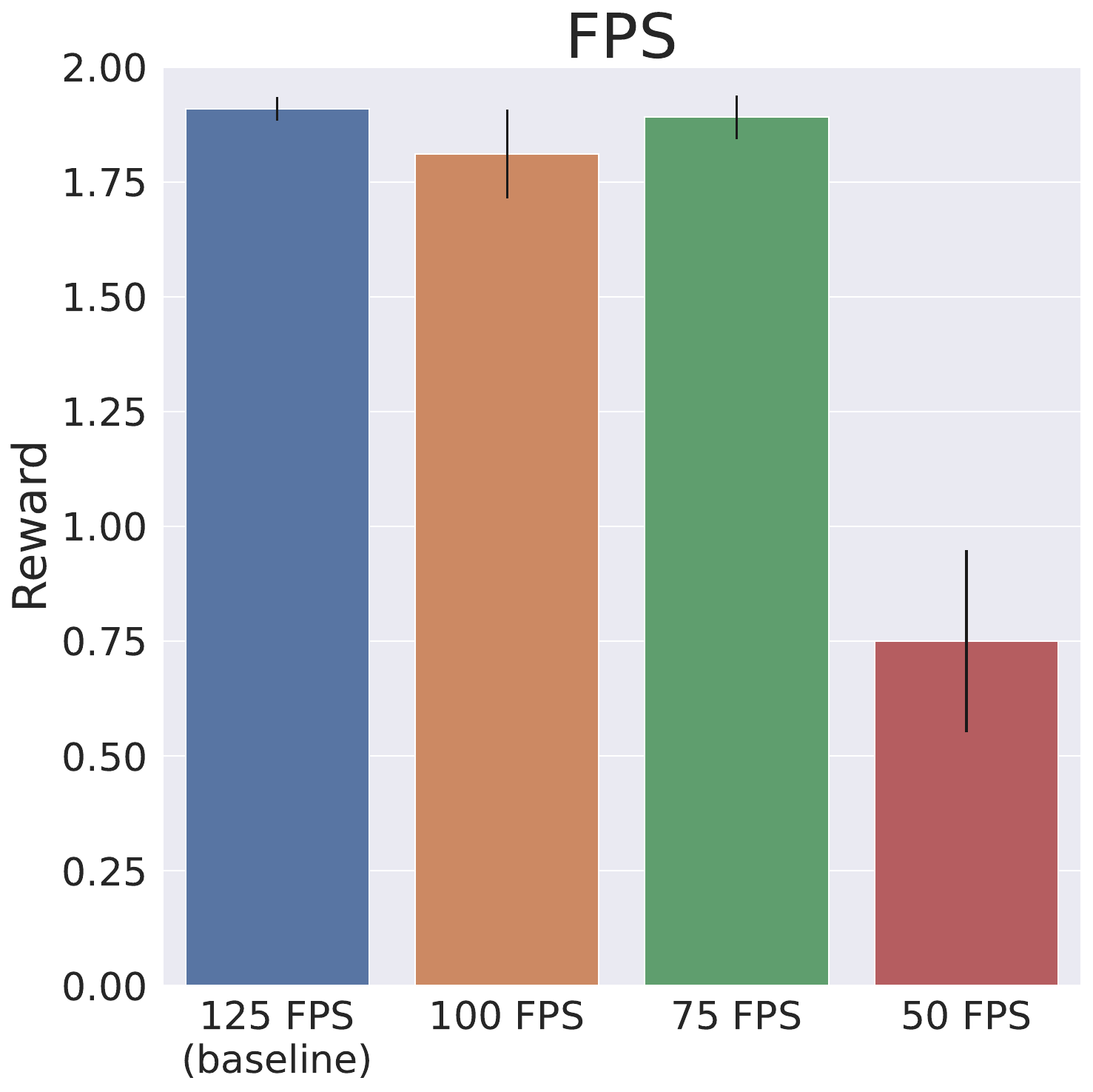}
    \includegraphics[width=0.24\textwidth]{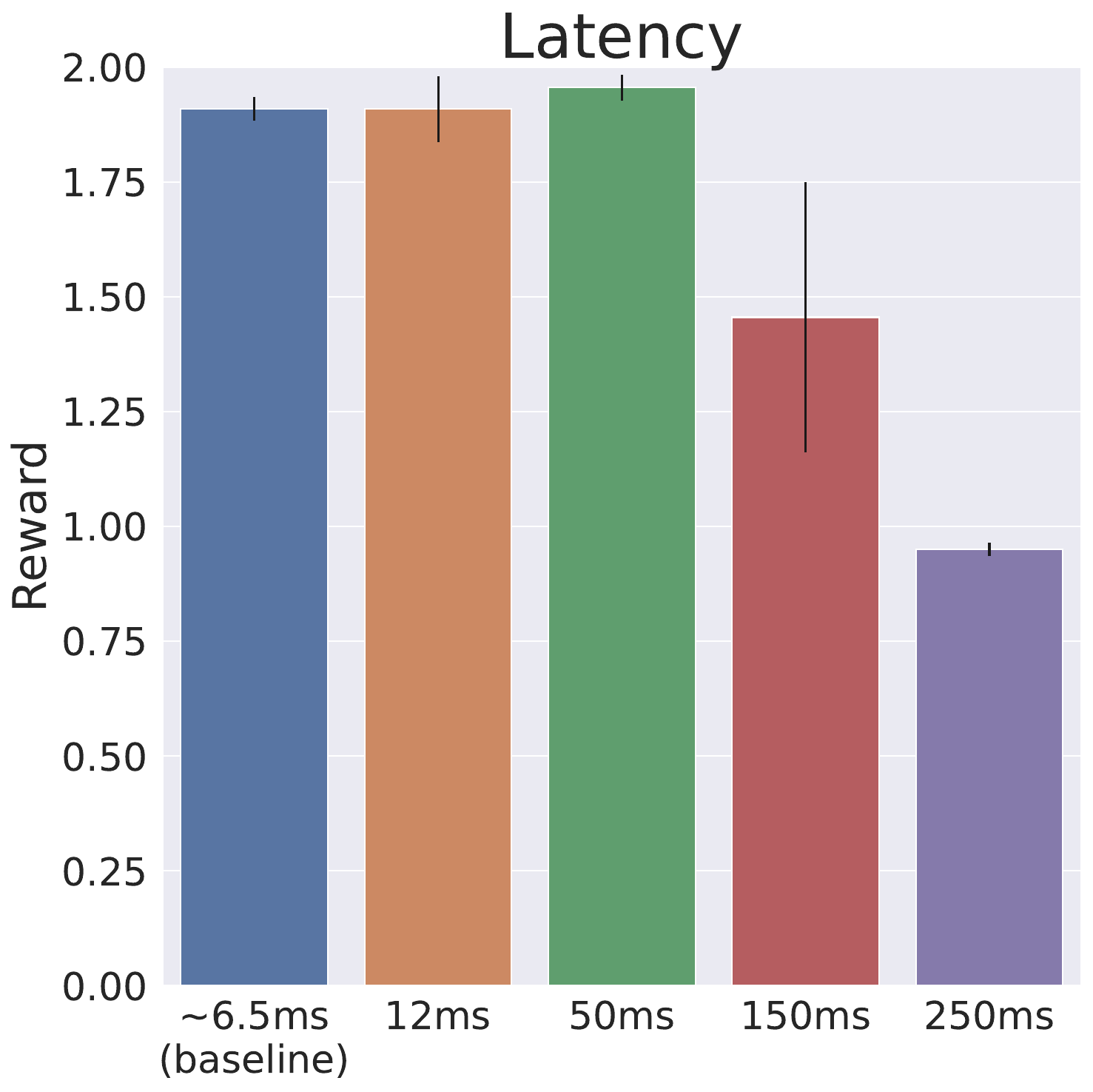}
    \includegraphics[width=0.48\textwidth]{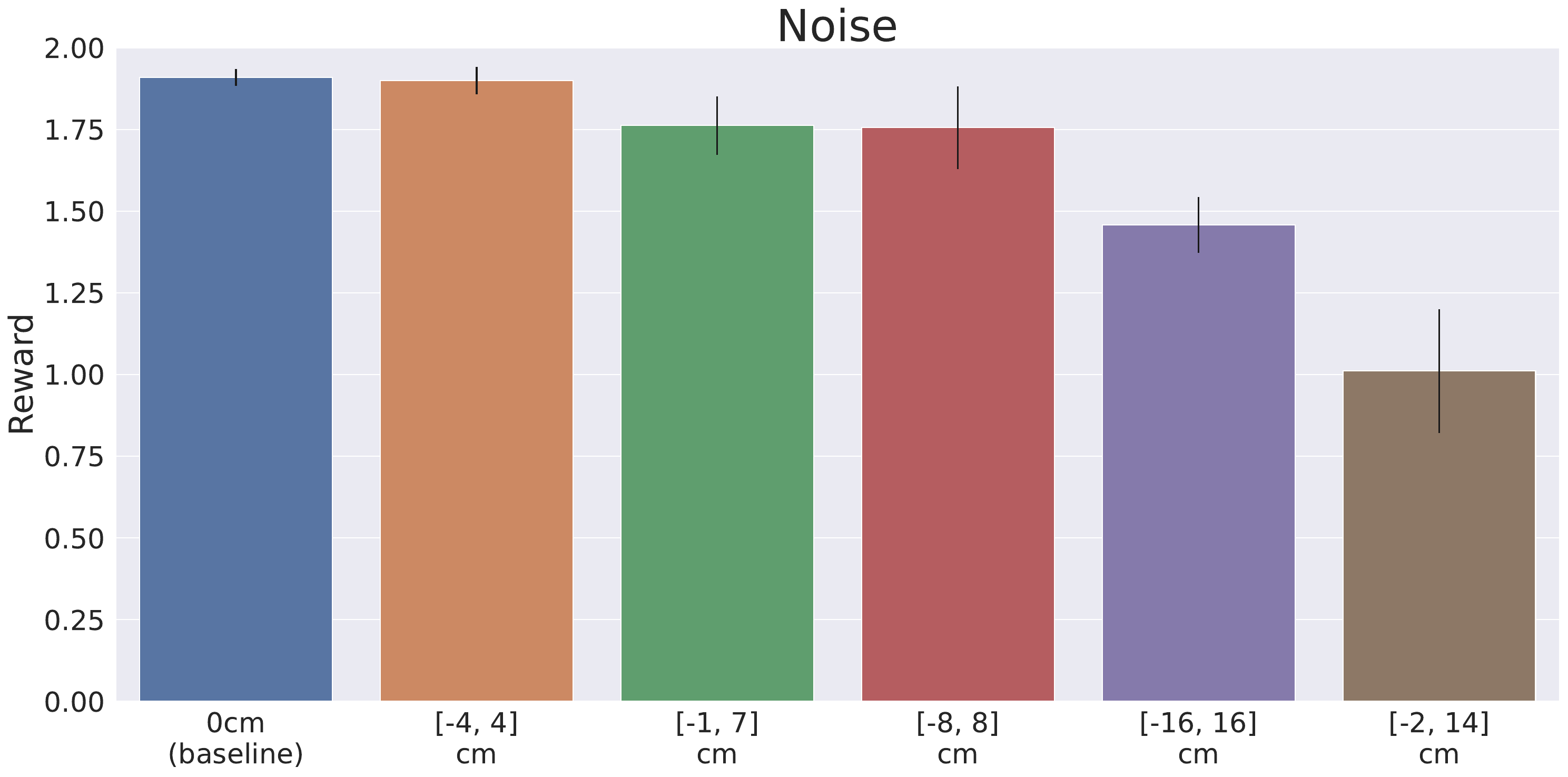}
    \caption{Perception resilience studies.  Reducing FPS and increasing latency have threshold points where performance of the system is stable until they reach a point where the robot can no longer react the to ball in them. Additional noise causes graceful degradation in performance, increased by non-zero mean distributions (common in vision triangulation). }
    \label{fig:vision_sweep}
\end{figure}

\subsubsection{Policies are sensitive to physical parameters, which can have complex interactions with each other} The physical parameter ablations in \autoref{fig:sim-params-ablations} (bottom right) reveal how sensitive policies are to all parameter values tested. Removing randomization from the table restitution coefficient (table: no R randomize) degrades performance by 14\%. Increasing the ball restitution coefficient by just 2\% reduces performance by 25\%, whilst increasing the table restitution coefficient by 8\% reduces performance by 36\%.
 
This study also highlights a current limitation of the system. Setting key parameters in the simulator such as the table and paddle restitution coefficients, or the paddle mass to values estimated following the process described in Appendix \ref{app:simulator} led to worse performance than tuned values (see measured v.s. tuned and also Appendix \ref{app:sim-ablations-details} for all parameter values).
We hypothesize this is because ball spin is not correctly modelled in the simulator and that the tuned values compensate for this for the particular ball distributions used in the real world. One challenge of a complex system with many interacting components is that multiple errors can compensate for each other, making them difficult to notice if performance does not suffer dramatically. It was only through conducting these studies that we became aware of the drop in performance from using measured values. In future work we plan to model spin and investigate if this resolves the performance degradation from using measured values. For further discussion on this topic, see Appendix \ref{app:system_id_revisited}.

\subsection{Perception Resilience Studies}

In this section we explore important factors in the perception system and how they affect end-to-end performance of the entire system. Latency and accuracy are two major factors and typically there is a tradeoff between them. A more accurate model may take longer to process but for fast moving objects (like a table tennis ball) it may be better to have a less accurate result more quickly. Framerate also plays a role. If processing takes longer than frames are arriving, latency will increase over time and eventually require dropping frames to catch up.  

For these experiments we select three high performing models from the baseline simulator parameter studies and test them on the real robot while modulating vision performance in the following ways: (1) reduce the framerate of the cameras , (2) increase latency by queuing observations and sending them to the policy at fixed intervals, and (3) reduce accuracy by injecting zero mean and non-zero mean noise to the ball position (over and above inherent noise in the system).

The results from these experiments can be seen in Figure~\ref{fig:vision_sweep}. For both framerate and latency, the performance stays consistent with the baseline until there is a heavy dropoff at 50 FPS and 150ms respectively, at which point the robot likely no longer has sufficient time to react to the ball and swings too late, almost universally resulting in balls that hit the net instead of going over.  There is a gentle decline in performance as noise increases, but the impact is much greater for non-zero mean noise: going from zero mean ([-4, 4] cm) noise to non-zero mean ([-1, 7] cm) is equivalent to doubling the zero mean noise ([-8, 8] cm).  The interpolation of observations described in \autoref{sec:real_env} likely serves as a buffer against low levels of zero mean noise.  Qualitatively, the robot's behavior was jittery and unstable when moderate noise was introduced.  Overall, the stable performance over moderate framerate and latency declines implies that designing around accuracy would be ideal for this task, although as trajectories become more varied and nuanced higher framerates may be necessary to capture their detailed behavior.

\subsection{ES Training Studies}

\begin{figure}
  \begin{center}
  \includegraphics[width=0.44\textwidth]{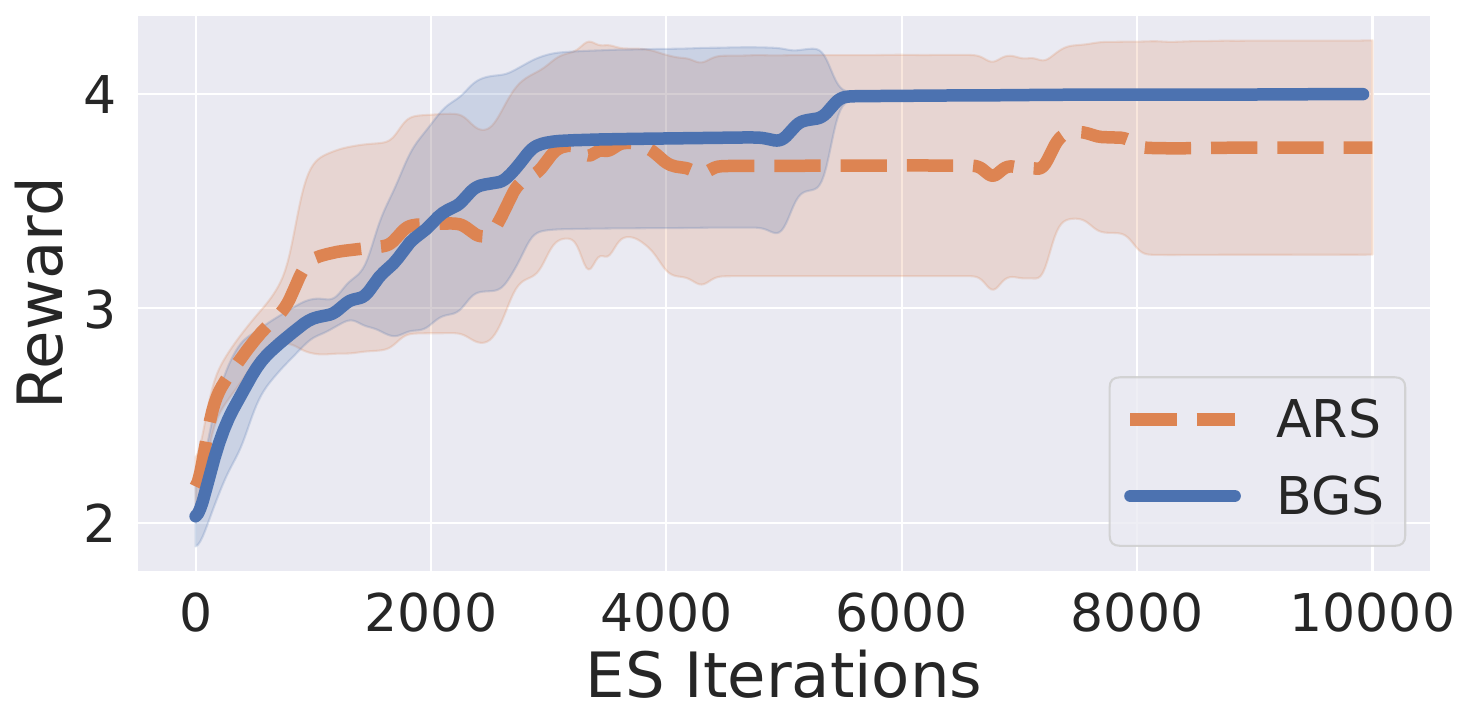}
  \includegraphics[width=0.24\textwidth]{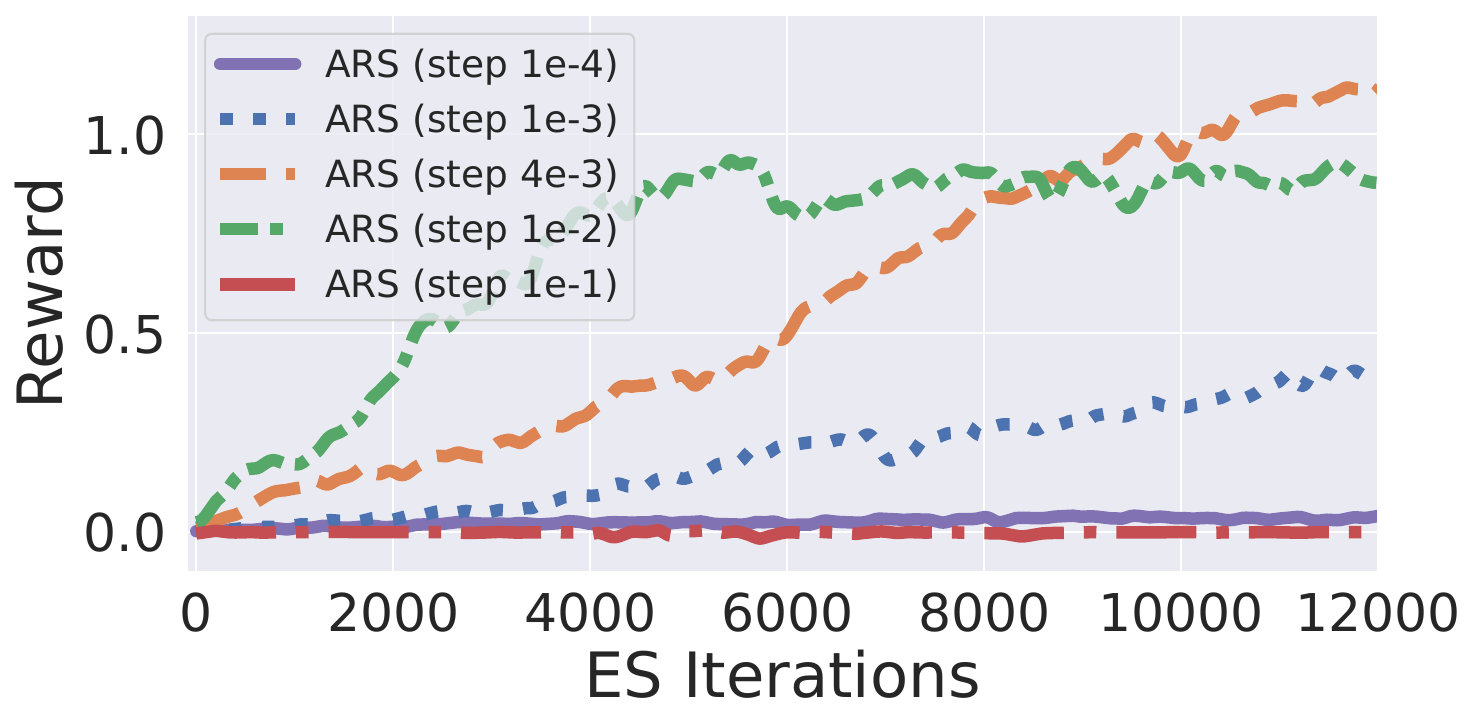}
  \includegraphics[width=0.24\textwidth]{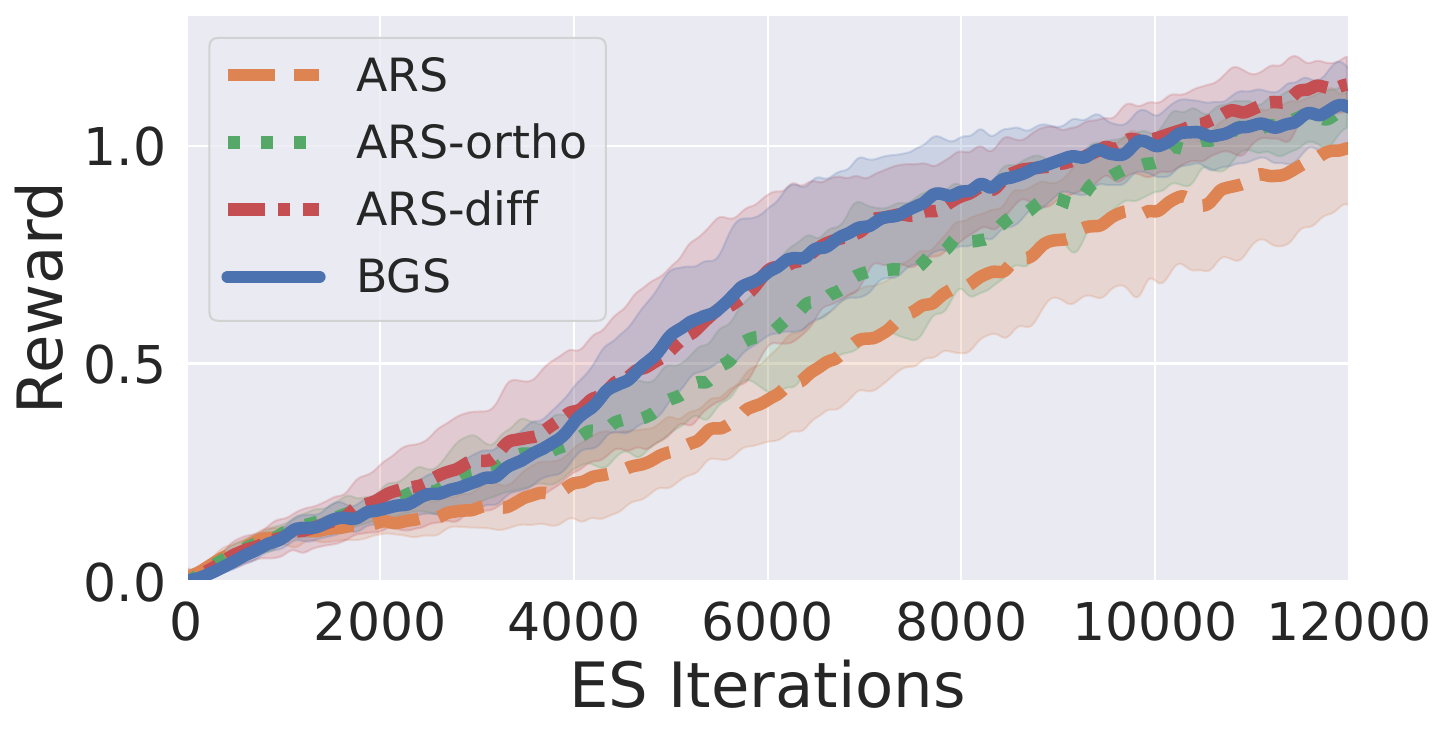}
\end{center}
\caption{BGS ablation studies. (top) BGS and ARS perform comparably on the ball return task with a narrow ball distribution. (bottom) A harder environment, ball targeting with a larger ball distribution. (left) Step-size alpha has a very significant effect on training success. (right) Improvements with reward differential elite-choice technique, orthogonal perturbation sampling and their combination (BGS).}
\label{fig:es_studies}
\end{figure}

BGS has been a consistent and reliable method for learning table tennis tasks on this system in simulation and fine-tuning in the real world. In this section we ablate the main components of BGS and compare it with a closely related method, ARS.

\autoref{fig:es_studies} (top) presents a comparison of BGS and ARS on the default ball return task against a narrow ball distribution. For both methods we set number of perturbations to 200, $\sigma$ to 0.025, and the proportion of perturbations selected as elites to 30\%. We roll out each perturbation for 15 episodes and average the reward to reduce reward variance due to stochasticity in the environment. We also apply the common approach of state normalization~\citep{SHCSS2017, nagabandi}. Under these settings, the methods are comparable.

Next we consider a harder \textit{ball targeting} task where the objective for the policy is to return the ball to a precise (randomized per episode) location on the opponent's side of the table \cite{ding2022learning}. We further increase the difficulty by increasing the range of incoming balls, i.e. using a wider ball distribution, and by decreasing the number of perturbations to 50. Tuning the step size $\alpha$ was crucial for successful policy training with ARS (\autoref{fig:es_studies} bottom left). An un-tuned step-size may lead to extremely slow training or fast training with sub-optimal asymptotic performance.

\autoref{fig:es_studies} (bottom right) shows the enhancements in training made by the BGS techniques independently and collectively compared to baseline ARS. Reward differential elite-choice and orthogonal sampling leads to faster convergence. As a result, BGS is the default ES algorithm for policy training.

\subsection{Acting and Observing in Task Space}
\label{sec:task_space_results}

The previous results use joint space for observations and actions. In this section we explore policies that operate in ``task space" (see \autoref{sec:task_space}). Task space has several benefits: it is compact, interpretable, provides a bounding cube for the end effector as a safety mechanism, and aligns the robot action and the observation spaces with ball observations. In our experiments we show that task spaces policies train faster and, more importantly, can be transferred to different robot morphologies.

\begin{figure}
  \begin{center}
  \includegraphics[width=0.24\textwidth]{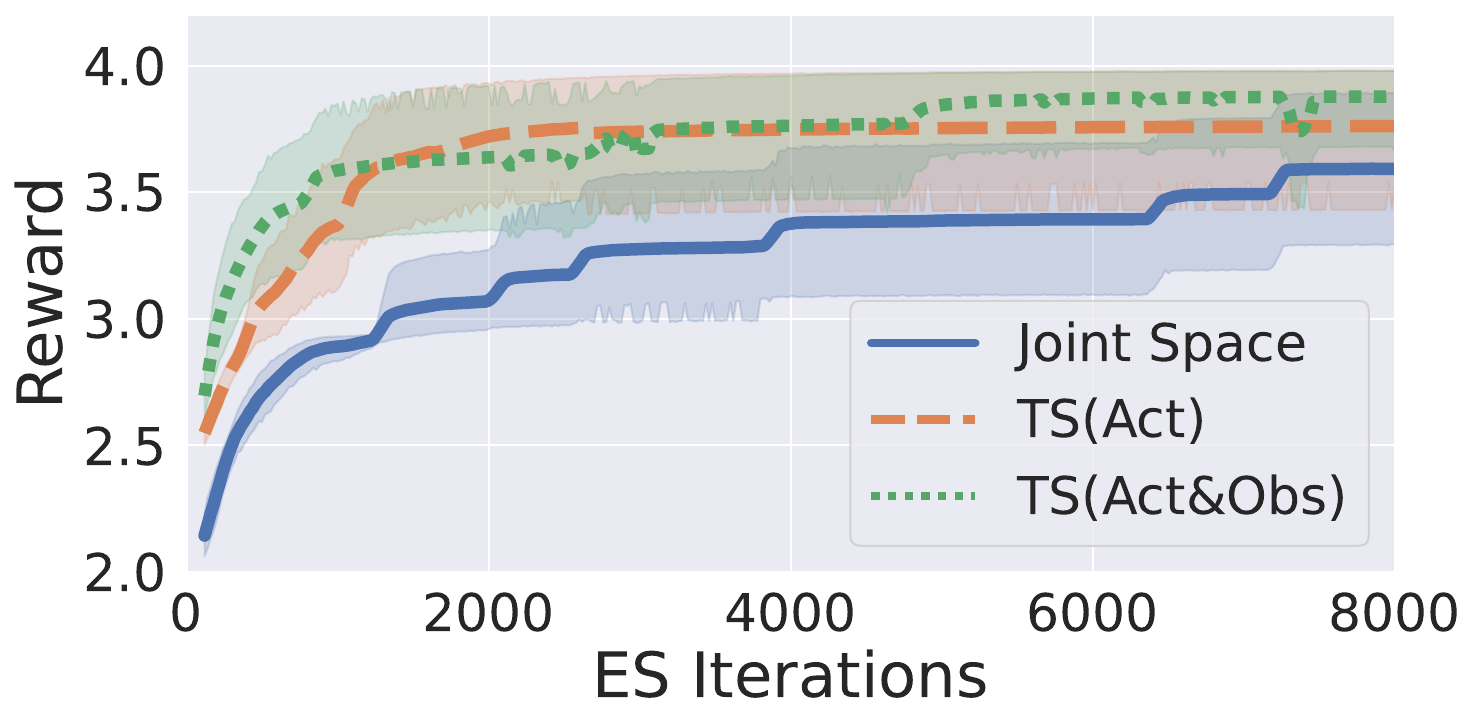}
  \includegraphics[width=0.24\textwidth]{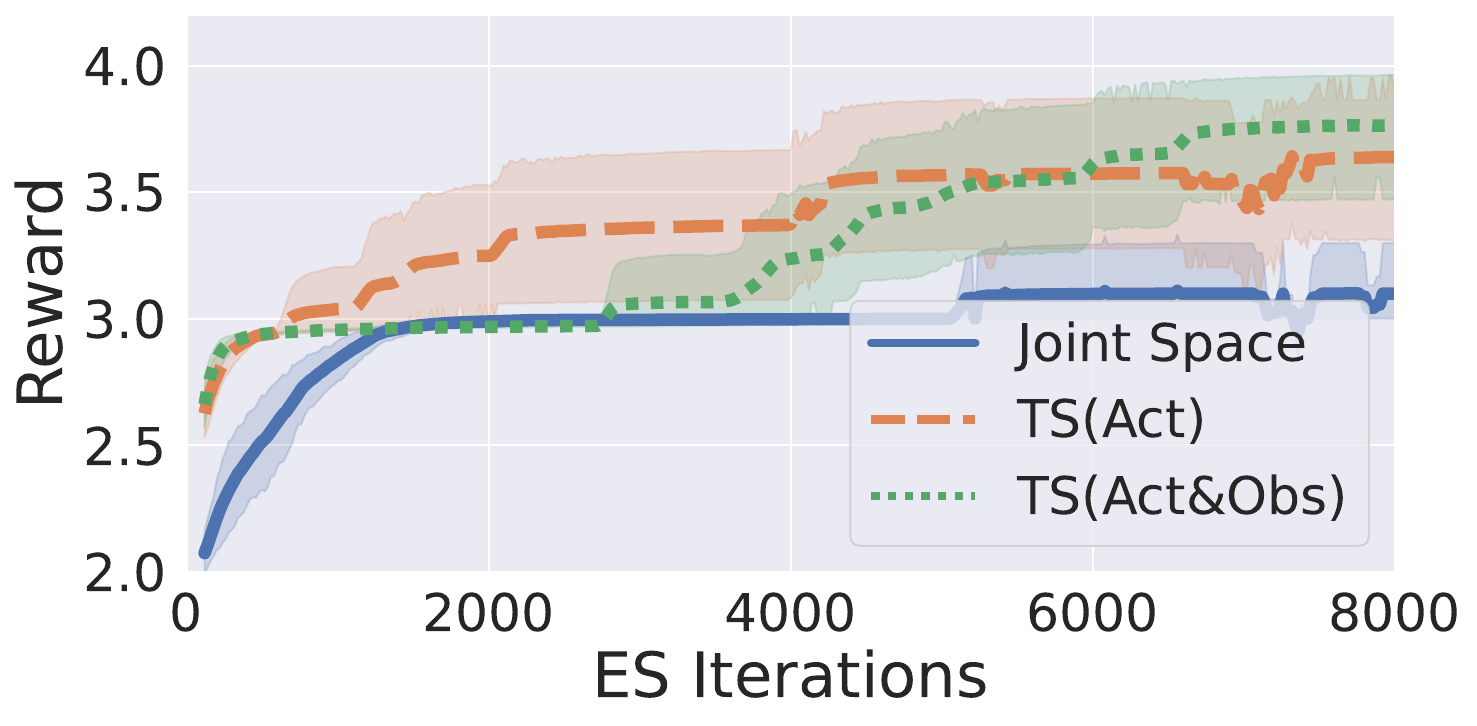}
  \includegraphics[width=0.48\textwidth]{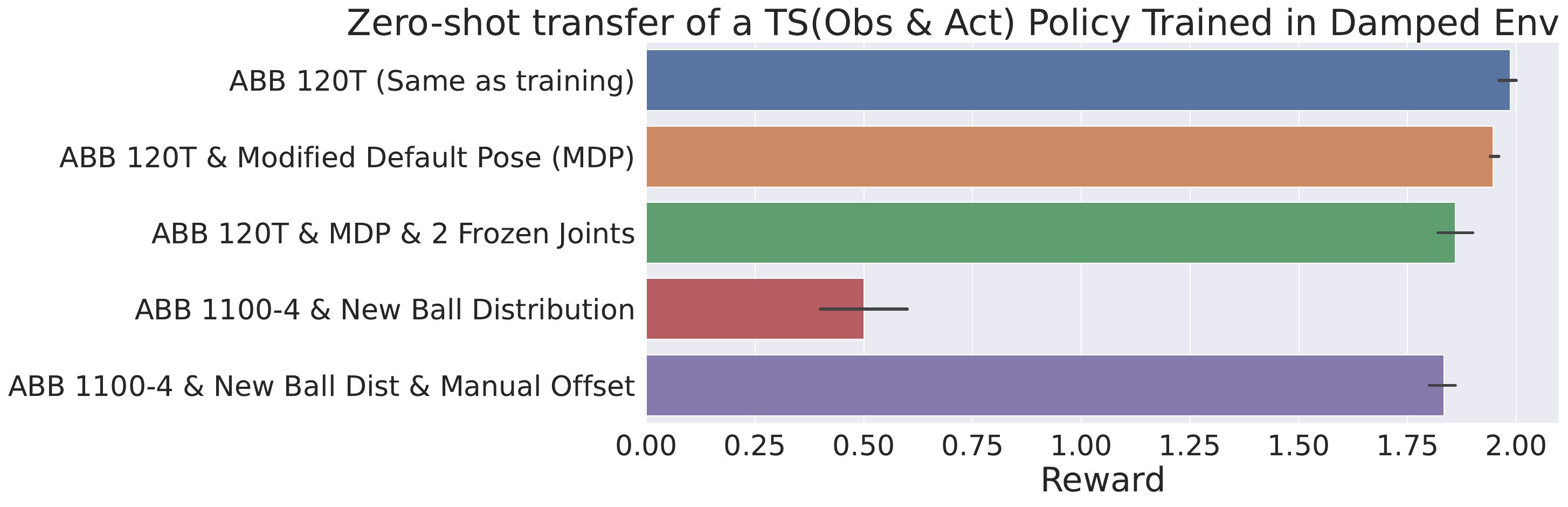}
\end{center}
\caption{Training policies in task space in the baseline environment (top-left) and a harder damped environment (top-right). Training converges faster in task-space for both scenarios. (bottom) A task space policy trained in the damped environment is successfully transferred to different morphologies and a new robot.}
\label{fig:cartesian}
\end{figure}

\autoref{fig:cartesian} (top left) compares training speed between joint space (JS), task space for actions --- TS(Act), and full task space policies (actions and observations) --- TS(Act\&Obs). Both task spaces policies train faster than JS policies. We also assess task space policies on a harder (damped) environment\footnote{Created by lowering the restitution coefficient of the paddle and ball, and increasing the linear damping of the ball.}. Now the robot needs to learn to swing and hit the ball harder. \autoref{fig:cartesian} (top right) shows that task space policies learn to solve the task (albeit not perfectly) while joint space policies gets stuck in a local maxima. For transfer performance of these policies see Appendix \ref{app:task_space}.

One crucial benefit of operating in task space is the robustness to different robots or morphologies. To demonstrate this, we first take the TS(Act\&Obs) model trained in the damped environment and transfer it to the real robot (\autoref{fig:cartesian} bottom). Performance is almost perfect with a score of 1.9. Next we change the initial pose of the robot and freeze two of the arm joints. Policy performance is maintained under a pose change (ABB 120T \& Modified Default Pose (MDP)) and only drops slightly when some joints are also frozen (ABB 120T \& MDP + 2 Frozen Joints). We then evaluate the policy on a robot with a different morphology and ball distribution and see that performance drops substantially. However, a task space policy is easily adaptable to new settings without retraining by adding a residual to actions to shift the paddle position. This is not possible when operating in joint space. Observing the robot showed that it was swinging too low and slightly off-angle and so adding a residual of 7cm above the table and 0.2 radians of roll causes the original policy performance to be nearly recovered (ABB 1100-4 \& New Ball Dist \& Manual Offset).

\subsection{Applying to a New Task: Catching}

While the system described above was designed for table tennis, it is general enough to be applied to other agile tasks. In this section, we apply it to a new task of catching a thrown ball and assess the effect of latency modelling, similar to the latency experiment from \autoref{ablations:sim-params}.

We used a similar system setup with minor modifications: a single horizontal linear rail (instead of two) and a lacrosse head as the end effector. The software stack and agents are similar with small differences: simplified \texttt{RewardManager} and \texttt{DoneManager}, soft body modelling of the net in simulation, trajectory prediction inputs for agents, and handling occlusions when the ball is close to the net. The BGS agents are similarly trained in a simulator before being transferred to the real hardware, where they are fine-tuned. Agents achieve a final catching success rate of $85\sim90\%$. For full details on the task see related work \citep{shankar2022catching}.

This task has a much larger variance in sim-to-real transfer due to difficulty in accurately modelling net \& ball capture dynamics.  As in the table tennis study, agents were trained in simulation with latencies of $100\%$, $0\%$, $20\%$, $50\%$, and $150\%$ of baseline latency.  Experiments with lower latency ($0\%$, $20\%$, and $50\%$) all transferred poorly, between $0\sim10\%$ catch rate.  Curiously, baseline latency and $150\%$ latency performed similarly, with one $150\%$ run achieving the best zero-shot transfer ever: a score equaling policies fine-tuned on the real robot.  This finding contradicts the results in the table tennis task, which prompted further investigation and revealed that the latency for this task was set incorrectly in the configuration file; the real value was much closer to the $150\%$ value.

This revelation dovetails with the $50\%$ latency table tennis results: a close latency can still give decent performance, but accurate values are better.  As such, it may be useful to generally run ablation studies such as these to challenge assumptions about the system and potentially find bugs.

\section{Related Work}

\subsection{Agile Robotic Learning}

The space of agile robotic learning systems is varied. It includes autonomous vehicles such as cars \cite{Pan-RSS-18,alvinn,muller-end-to-end,bojarski2016end,bojarski2017explaining}, legged locomotion \cite{Nguyen-RSS-17, Tan18, haarnoja2018learning, peng2020learning, smith2022legged, smith2022walk, agarwal2022legged}, as well as dynamic throwing \citep{task-level-ball-throwing, Kober-RSS-10, ghadirzadeh2017deep, zeng2020tossingbot}, catching \citep{shankar2022catching}, and hitting --- which is where table tennis fits.

Many of these systems face similar challenges --- environment resets, latency, safety, sim-to-real, perception, and system running speed as exemplified in strict inference and environment step time requirements.

The benefits of automatic resets have been demonstrated in quadrupedal systems \citep{smith2022legged, smith2022walk} and throwing \citep{zeng2020tossingbot}. To our knowledge, this system is the first table tennis learning system with automatic resets, enabling autonomous training and evaluation in the real world for hours without human intervention.

Latency is a well known problem in physical learning systems \citep{Tan18}. The system contributes to this area by extending \citep{Tan18}, modeling multiple latencies in simulation, and by validating its importance through extensive experiments. Orthogonally, the system also includes observation interpolation on the physical system as a useful technique for increasing the robustness of deployed policies to latency variation (e.g. from jitter). We demonstrated empirically the robustness of policies to substantial injections of latency and hypothesize that the observation interpolation plays a crucial role in this.

Safety is another crucial element that becomes very important with fast moving robots. Trajectory planners \cite{kroger2011opening} can avoid static obstacles, neural networks can check for collisions \cite{kew_neural_collision}, safe RL can be used to restrict state spaces \cite{yang2022safe}, or a system can learn from safe demonstrations \cite{Muelling2010LearningTTMOMP, Muelling2012LearningSelectGen, Huang2016JointlyLT}. In contrast, this system runs a parallel simulation during deployment as a safety layer. Doing so is beneficial because the robot policy runs at a high frequency and there are several physical environments and robots and it enables (1) definition of undesirable states and (2) preventing a physical robot from reaching them. To the best of our knowledge this is also a novel component of the system.

Learning controllers from scratch in the real world can be challenging for an agile robot due to sample inefficiency and dangers in policy exploration.  Training first in a simulator and then deploying to the real robot ~\citep{QuadChall2020, rubikscube, Tan18} (i.e. sim-to-real) is an effective way to mitigate both issues, but persistent differences between simulated and real world environments can be difficult to overcome~\citep{S2R-RA-AC, WhySimsFail}.

Perception is crucial in helping robots adapt to changes in the environment \citep{agarwal2022legged, wang2022machine} and interact with relevant objects \citep{zeng2020tossingbot, Kober-RSS-10}. 
When objects need to be tracked at high speed such as in catching or hitting, it is typical to utilize methods such as motion-capture systems~\cite{mori2019high} however for table tennis, the ball needs to adhere to strict standards that prevent instrumentation or altering of the ball properties. Passive vision approaches for detecting the location within a video frame of a bright colored ball from a stationary camera may seem trivial, however, applying image processing techniques \citep{Tebbe2018ATT} such as color thresholding \citep{wong2008}, shape fitting~\citep{hettihewa2021development}, and background subtraction are problematic. When considering the typical video captured from the cameras several factors in the scene render such approaches brittle. For example, the color of the natural light changes through out the day. Even under fixed lighting, the video stream is captured at 125Hz which is above the Nyquist frequency of the electricity powering fluorescent lights, resulting in images that flicker between frames. Additionally, there are typically several leftover balls from previous episodes around the scene which share the same color and shape as the ball in play. These distractors make data association more of a challenge for down stream tracking. Finally, extracting things that move is also a challenge when other basic visual cues are unreliable because there is always a robot and or a human moving in the scene. The perception component of the system in this paper uniquely combined all these visual cues by learning to detect the ball in an end-to-end fashion that is robust to visual ambiguities and provides both precise ball locations and velocity estimates.

Finally, prior work in robot learning varies by how much it focuses on the system compared with the problem being tackled. \cite{Eppner-RSS-16, Jing-RSS-16, Lozano-Perez-RSS-19, smith2022walk, Muelling2010Biomem, Tebbe2018ATT, QuadChall2020} are examples of works which dedicate substantial attention to the system. They provide valuable details and know-how about what mattered for a system to work in practice. This work is spiritually similar.

\subsection{Robotic Table Tennis}
\label{sec:robot_table_tennis}

Robotic table tennis is a challenging, dynamic task~\citep{TableTennisMuscular} that has been a test bed for robotics research since the 1980s~\citep{Billingsley83, Knight1986PingpongplayingRC,Hartley87,Hashimoto1987DevelopmentOP, Muelling2010Biomem}. The current exemplar is the Omron robot~\citep{omron}. Until recently, most methods tackled the problem by identifying a virtual hitting point for the racket~\citep{Miyazaki2002RealizationOT, Miyazaki2006LearningTD, Anderson1988ARP, Muelling2010SimulatingHT, Zhu2018TowardsHL, Huang2015LearningOS, Sun2011BalanceMG, Mahjourian2018HierarchicalPD}. These methods depend on being able to predict the ball state at time $t$ either from a ball dynamics model which may be parameterized~\citep{Miyazaki2002RealizationOT, Miyazaki2006LearningTD, Matsushima2003LearningTT, Matsushima2005ALA} or by learning to predict it~\citep{Muelling2010Biomem, Muelling2010SimulatingHT, Zhu2018TowardsHL}. Various methods can then generate robot joint trajectories given these target states~\citep{Muelling2010Biomem, Miyazaki2002RealizationOT, Miyazaki2006LearningTD, Matsushima2003LearningTT, Matsushima2005ALA, Muelling2010LearningTTMOMP, Muelling2012LearningSelectGen, Huang2016JointlyLT, Ko2018OnlineOT, Tebbe2018ATT, Gao2019MarkerlessRP}. More recently,~\citet{SERL_tebbe} learned to predict the paddle target using reinforcement learning (RL).

Such approaches can be limited by their ability to predict and generate trajectories. An alternative line of research seeks to do away with hitting points and ball prediction models, instead focusing on high frequency control of a robot's joints using either RL~\citep{TableTennisMuscular, Zhu2018TowardsHL, GaoPPOES2020} or learning from demonstrations~\citep{Muelling2012LearningSelectGen, LFSD-GT, Chen2020}. Of these,~\citet{TableTennisMuscular} is the most similar to the system in this paper. Similar to \citet{TableTennisMuscular}, this system trains RL policies to control robot joints at high frequencies given ball and robot states as policy inputs. However~\citet{TableTennisMuscular} uses hybrid sim and real training as well as a robot arm driven by pneumatic artificial muscles (PAMs), whilst this system uses a motor-driven arm. Motor-driven arms are a common choice and used by \cite{LFSD-GT, Tebbe2018ATT, SERL_tebbe, Muelling2010LearningTTMOMP}.

\section{Takeaways and Lessons Learned}
\label{sec:takeaways}
Here we summarize lessons learned from the system that we hope are widely applicable to high-speed learning robotic systems beyond table tennis.

Choosing the right robots is important.  The system started with a scaled down version of the current setup as a proof of concept and then graduated to full-scale, industrial robots (Appendix \ref{app:hardware}).  Industrial robots have many benefits such as low latency and high repeatability, but they can come with ``closed-box" issues that must be worked through (\autoref{sec:control}).

A safety simulator is a dynamic and customizable solution to constraining operations with high frequency control compared to high-level trajectory planners (\autoref{sec:control}).

A configurable, modular, and multi-language (e.g. C++ and Python) system improves research and development velocity by making experimentation and testing easy for the researcher (\autoref{sec:control}).

Latency modeling is critical for real world transfer performance as indicated by our experimental results.  Other environmental factors may have varying effects that change based on the task (\autoref{ablations:sim-params}).  For example, ball spin is not accurately modeled in the ball return task, but can be critical when more nuanced actions are required.

Accurate environmental perception is also a key factor in transfer performance.  In this system's case many factors were non-obvious to non-vision experts: camera placement, special calibration techniques, lens locks, etc. all resulted in better detection (\autoref{sec:perception}).

GPU data buffering, raw Bayer pattern detection, and patch based training substantially increase the performance of high frequency perception (\autoref{sec:perception}).  Rather than using an off-the-shelf perception module, a purpose-built version allows levels of customization that may be required for high-speed tasks.

Interpolating and smoothing inputs (\autoref{sec:real_env}) solves the problem of different devices running at different frequencies.  It also guards against zero-mean noise and system latency variability, but is less effective against other types of noise.

Automatic resets and remote control increase system utilization and research velocity (\autoref{sec:real_env}).  The system originally required a human to manually collect balls and control the thrower.  Now that the system can be run remotely and ``indefinitely'', significantly more data collection and training can occur.

ES algorithms like BGS (\autoref{sec:sim_training}) are a good starting point to explore the capabilities of a system, but they may also be a good option in general.  BGS is still the most successful and reliable method applied in this system.  Despite poor sample efficiency, ES methods are simple to implement, scalable, and robust optimizers that can even fine-tune real world performance.

Humans are highly variable and don't always follow instructions (on purpose or not) and require significant accommodations to address these issues and also to alleviate frustrations (e.g. time to reset) and ensure valuable human time is not wasted.

\subsection{Limitations and Future Work}

A guiding principal of the system has been not to solve everything at once. Starting with a simple task (e.g. hitting the ball) and then scaling up to more complex tasks (e.g. playing with a human) provides a path to progress naturally prioritizes inefficiencies to be addressed.  For example, a long but clean environment reset was sufficient for learning ball return tasks, but needed optimization to be sufficiently responsive to a human. 

The current system struggles with a few key features. More complex play requires understanding the spin of the ball and the system currently has no way to directly read spin and it is not even included in simulation training. While it is possible to determine spin optically (i.e. by tracking the motion of the logo on the ball), it would require significantly higher frame rates and resolutions than what is currently employed. Other approaches more suited to our setup include analyzing the trajectory of the ball (which the robot may be doing implicitly) or including the paddle/thrower pose into the observation; analogous to how many humans detect spin. Additionally learning a model of the opponent if the opponent attempts to be deliberately deceptive, concealing of adding confusion to their hits.

The robot's range of motion is significant thanks to the inclusion of the gantry, but is still limited in a few key ways. Firstly, the safety simulator does not allow the paddle to go below the height of the table, preventing the robot from ``scooping'' low balls. This restriction prevents the robot from catching the arm between the table and gantry, which the safety sim was unable to prevent in testing. The robot is limited in side-to-side motion as well as how far forward over the table it can reach, so there may be balls that it physically cannot return.  Finally, so far the robot has not made significant use of motion away from the table.  We hope that training on more complex ball distributions will require the robot to make full use of the play space as professional humans do.

The sensitivity of policies also increases as the task becomes more complex. For example, slight jitter or latency in inference may be imperceptible for simple ball return tasks, but more complex tasks that require higher precision quickly revealed these gaps requiring performance optimizations. Sim-to-real gaps are also an issue; hitting a ball can be done without taking spin into account, but controlling spin is essential for high-level rallying. Environmental parameters and ball spin both become more important and incorporating domain randomization is a promising path forward to integrating them in a robust manner. Additionally, when human opponents come into play, modeling them directly or indirectly make it possible for the robot to move beyond purely reactive play and to start incorporating strategic planning into the game.

\section{Conclusion} 
\label{sec:conclusion}

In this paper we have explored the components of a successful, real-world robotic table tennis system. We discussed the building blocks, trade-offs, and other design decisions that went into the system and justify them with several case studies. While we do not believe the system in this paper is the perfect solution to building a learning, high-speed robotic system, we hope that this deep-dive can serve as a reference to those who face similar problems and as a discussion point to those who have found alternative approaches.

\section*{Acknowledgments}

We would like to thank Arnab Bose, Laura Downs, and Morgan Worthington for their work on improving the vision calibration system and Barry Benight for their help with video storage and encoding.  We would also like to thank Yi-Hua Edward Yang and Khem Holden for improvements to the ball thrower control stack.  We also are very grateful to Chris Harris and Razvan Surdulescu for their overall guidance and supervision of supporting teams such as logging and visualization.  Additional thanks go to Tomas Jackson for video and photography and Andy Zeng for a thorough review of the inital draft of this paper.  And finally we want to thank Huong Phan who was the lab manager for the early stages of the project and got the project headed in the right direction.

\bibliographystyle{plainnat}
\bibliography{references}

\clearpage

\appendix

\subsection{Author Contributions}
\label{sec:contributions}

\subsubsection{By Type}

Names are listed alphabetically.

\begin{itemize}
\item \textbf{Designed or implemented the vision system:} Michael Ahn, Alex Bewley, David D’Ambrosio, Navdeep Jaitly, Grace Vesom
\item \textbf{Designed or implemented the vision policy training infrastructure:} Alex Bewley, David D’Ambrosio, Navdeep Jaitly, Juhana Kangaspunta
\item \textbf{Designed or implemented the vision policy:} Alex Bewley, David D’Ambrosio, Navdeep Jaitly
\item \textbf{Designed or implemented the robot control stack:} Saminda Abeyruwan,  Michael Ahn, David D’Ambrosio, Laura Graesser, Atil Iscen, Navdeep Jaitly, Satoshi Kataoka, Sherry Moore, Ken Oslund, Pannag Sanketi, Anish Shankar, Peng Xu
\item \textbf{Designed or implemented the real world gym environment:} Saminda Abeyruwan, David D’Ambrosio, Laura Graesser, Satoshi Kataoka, Pannag Sanketi, Anish Shankar
\item \textbf{Designed or implemented the simulator:} Saminda Abeyruwan, Erwin Coumans, David D’Ambrosio, Laura Graesser, Navdeep Jaitly, Nevena Lazic, Reza Mahjourian, Pannag Sanketi, Anish Shankar, Avi Singh
\item \textbf{Designed or implemented the visualization:} Yuheng Kuang, Anish Shankar
\item \textbf{Designed or implemented the nightly monitoring:} Saminda Abeyruwan, Omar Cortes, David D’Ambrosio, Laura Graesser, Pannag Sanketi, Anish Shankar
\item \textbf{Robot operations and mechanical engineering:} Jon Abelian, Justin Boyd, Omar Cortes, Gus Kouretas, Thinh Nguyen, Krista Reymann
\item \textbf{Designed or implemented learning infrastructure and algorithms:} Krzysztof Choromanski, Tianli Ding, Wenbo Gao, Laura Graesser, Deepali Jain, Navdeep Jaitly, Nevena Lazic, Corey Lynch, Avi Singh, Saminda Abeyruwan, Anish Shankar
\item \textbf{Designed or implemented control policy architectures:} Tianli Ding, Laura Graesser, Navdeep Jaitly
\item \textbf{Ran experiments for the paper:} Alex Bewley, David D’Ambrosio, Laura Graesser, Atil Iscen, Deepali Jain, Anish Shankar
\item \textbf{Wrote the paper:} Saminda Abeyruwan, Alex Bewley, David D’Ambrosio, Laura Graesser, Atil Iscen, Deepali Jain, Ken Oslund, Anish Shankar, Avi Singh, Grace Vesom, Peng Xu
\item \textbf{Core team:} Saminda Abeyruwan, Alex Bewley, David D’Ambrosio, Laura Graesser, Navdeep Jaitly, Krista Reymann, Pannag Sanketi, Avi Singh, Anish Shankar, Peng Xu
\item \textbf{Managed or advised on the project:} Navdeep Jaitly, Pannag Sanketi, Pierre Sermanet, Vikas Sindhwani,  Vincent Vanhoucke
\item \textbf{Table tennis coach:} Barney Reed
\end{itemize}

\subsubsection{By Person} Names are listed alphabetically.

\textbf{Jonathan Abelian:} Day to day operations and thrower maintenance.

\textbf{Saminda Abeyruwan:} Worked on multiple components including: real gym environment, optimizing runtime for high frequency control, simulator, system identification, ball distribution protocol. Wrote parts of paper related to ball distribution mapping and real environment design.

\textbf{Michael Ahn:} Built an earlier version of the vision infrastructure; built the low-level ABB/Festo control infrastructure.

\textbf{Alex Bewley:} Led the design and implementation for the vision system. Built components for data infrastructure and model training. Performed noise and bias analysis for different camera configurations. Assisted with experimentation. Collaborated on paper writing and editing.

\textbf{Justin Boyd:} Designed fixtures, tuned and calibrated vision system, robot bring-up and integration.

\textbf{Krzysztof Choromanski:}  Built the ES distributed optimization engine for all ES experiments with Deepali Jain. Co-author of the BGS algorithm with Deepali Jain. Led the research on distributed ES optimization for the project. Ran several ES experiments throughout the project.

\textbf{Omar Cortes:} Assisted with experiments and fine-tuning and maintained the systems' integrity through nightly test monitoring.

\textbf{Erwin Coumans:} Helped set up the simulation environment using PyBullet and provided simulation support.  Also developed early prototypes for exploring the system.

\textbf{David D'Ambrosio:} Worked on the system across multiple parts including: vision, robot control, gym environment, simulation, and monitoring. Coordinated paper writing, wrote and edited many sections and edited the video.  Ran several ablation studies.  Coordinated with operations.

\textbf{Tianli Ding:} Implemented Goal’sEye learning infrastructure, conducted extensive experiments to train goal-targeting policies.

\textbf{Wenbo Gao:} Experimentation with ES methods and developing curriculum learning for multi-modal playstyles.

\textbf{Laura Graesser:} Worked on multiple parts of the system including: simulation, policy learning, real gym environment, robot control. Shaped paper narrative, wrote in many sections, and edited throughout. Designed and ran simulation parameters system studies.

\textbf{Atil Iscen:} Added the task space controller and observation space. Trained and deployed policies in task space. Compiled experimental results and contributed to writing for these sections.

\textbf{Navdeep Jaitly:} Conceived, designed, and led the initial stages of the project, built and sourced multiple prototypes, laid the foundation for the design of major systems like control and vision.  Created initial vision inference pipeline and supervised algorithm development.

\textbf{Deepali Jain:} Built the ES distributed optimization engine used for all policy training experiments with Krzysztof Choromanskii. Co-author of the BGS algorithm with Krzysztof Choromanski. Ran several ES experiments throughout the project. Conducted ablation study comparing ARS and BGS techniques for the paper.

\textbf{Juhana Kangaspunta:} Built an early version of the computer vision system used in the robot and created a labeling pipeline.

\textbf{Satoshi Kataoka:} Developed and maintained the custom MPI system. Initial consultation on cameras and other infrastructure-related components.

\textbf{Gus Kouretas:} Work cell assembly and day to day operation.

\textbf{Yuheng Kuang:} Technical lead for data infrastructure and provided visualization tools.

\textbf{Nevena Lazic:} Built an initial version of the simulator, implemented and ran initial ES experiments.

\textbf{Corey Lynch:} Implemented GoalsEye learning infrastructure and advised on goal-targeting experiments.

\textbf{Reza Mahjourian:} Built an initial version of the simulator, developed early RL agent control, and defined the land ball task.

\textbf{Sherry Q. Moore:} Control, sensing and ball-thrower infrastructure decisions. Designed and implemented the first working version of C++ control stack.

\textbf{Thinh Nguyen:} Thrower mechanical design, linear axis design and maintenance, ball return system design, and day-to-day operations.

\textbf{Ken Oslund:} Designed and implemented the C++ client and controller backend along with the bindings which make them interoperable with Python. Also directly wrote several paragraphs in the final paper.

\textbf{Barney J Reed}: Expert table tennis advisor, coaching engineers, human data collection.

\textbf{Krista Reymann:} Operations project manager, overseeing operations team, sourcing parts and resources, coordinating with vendors and managing repairs.

\textbf{Pannag R Sanketi:} Overall lead on the project. Guided the algorithm and system design, wrote parts of the system, advised on the research direction, managed the team, project scoping and planning. 

\textbf{Anish Shankar:} Core team member working on the system across multiple parts including: performance, hardware, control, tooling, experiments.

\textbf{Pierre Sermanet:} Advised on the GoalsEye research.

\textbf{Vikas Sindhwani:} Initiated and developed ES research agenda for table tennis and catching. Supported and advised on an ongoing basis. Edited the paper.

\textbf{Avi Singh:} Worked on multiple parts of the system, focusing on simulation and learning algorithms. Helped write the paper.

\textbf{Vincent Vanhoucke:} Infrastructure decisions, project scoping and research milestones.

\textbf{Grace Vesom:} Built the camera driver and an early version of the ball detection pipeline. Built camera calibration software and hardened camera hardware stability.

\textbf{Peng Xu:} Worked on early versions of many parts of the system including: vision, robot control, and designing the automatic ball collection.  Wrote part of the paper.

\subsubsection{Contact Authors}

\texttt{\{bewley, ddambro, lauragraesser, psanketi\}}@google.com

\subsection{Hardware Details}
\label{app:hardware}

\subsubsection{Host Machines}

The entire system is run on a single Linux host machine with an AMD Ryzen Threadripper 3960X and multiple NVIDIA RTX A5000 accelerators.  One accelerator handles perception inference and another encodes the images from the cameras with \textit{nvenc} for storage (see Appendix \ref{app:logging}). Policy inference runs on CPU. The standard robot policies are so small that the time to transfer the input data from CPU to accelerator exceeds any savings in running inference on the accelerator.

Previous iterations of this system used a dual Intel Xeon E5-2690s, two Nvidia Titan V accelerators, and a Quadro M2000. The Quadro handled video encoding and the two Titans each handled a single camera stream in an older iteration of the perception system that could not maintain framerates without splitting the load across multiple GPUs.  The current system was a substantial upgrade in terms of performance; by switching machines perception inference latency halved.

\subsubsection{Robots}

\begin{figure*}[!h]
    \centering
    \includegraphics[width=0.30\textwidth]{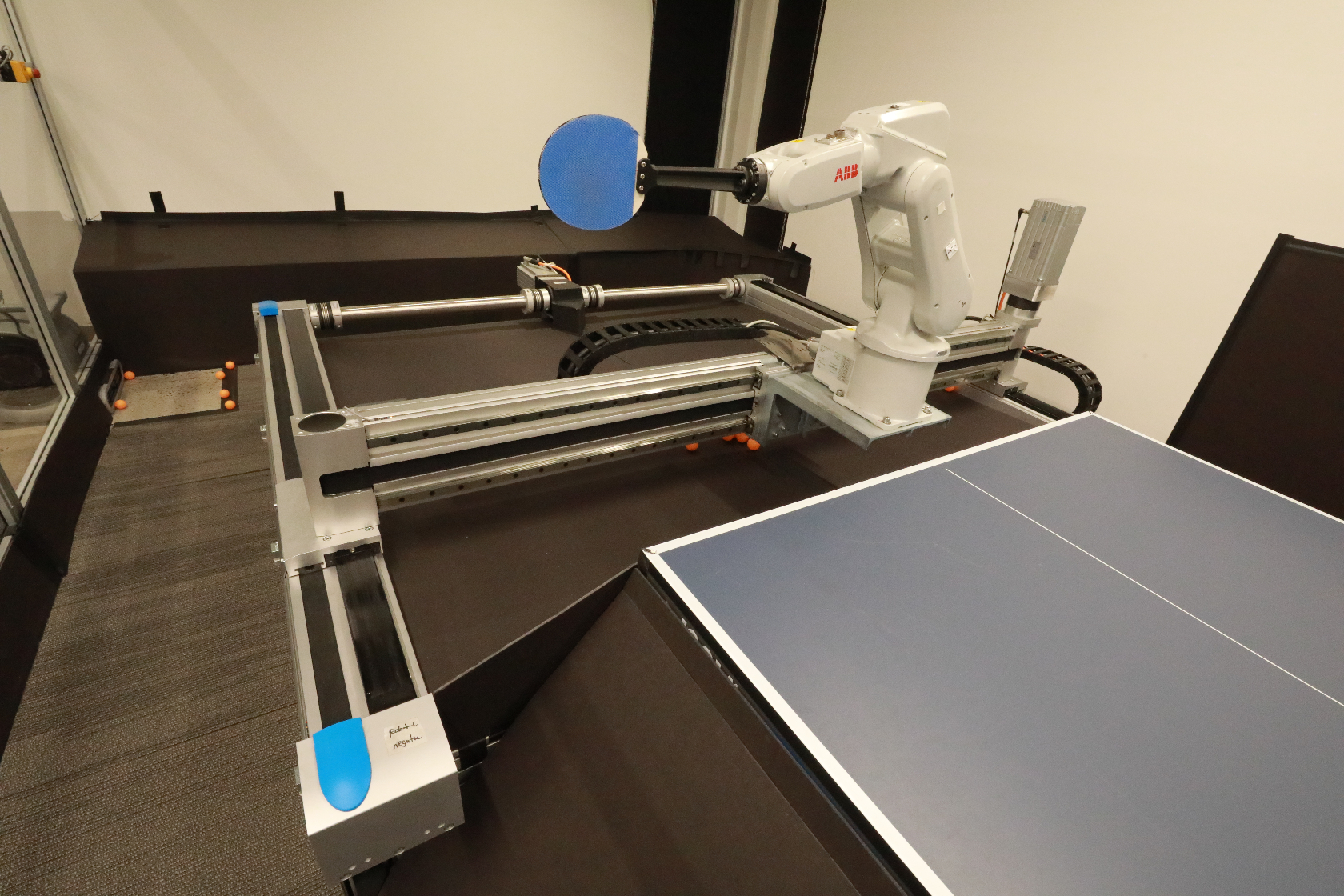}
    \includegraphics[width=0.30\textwidth]{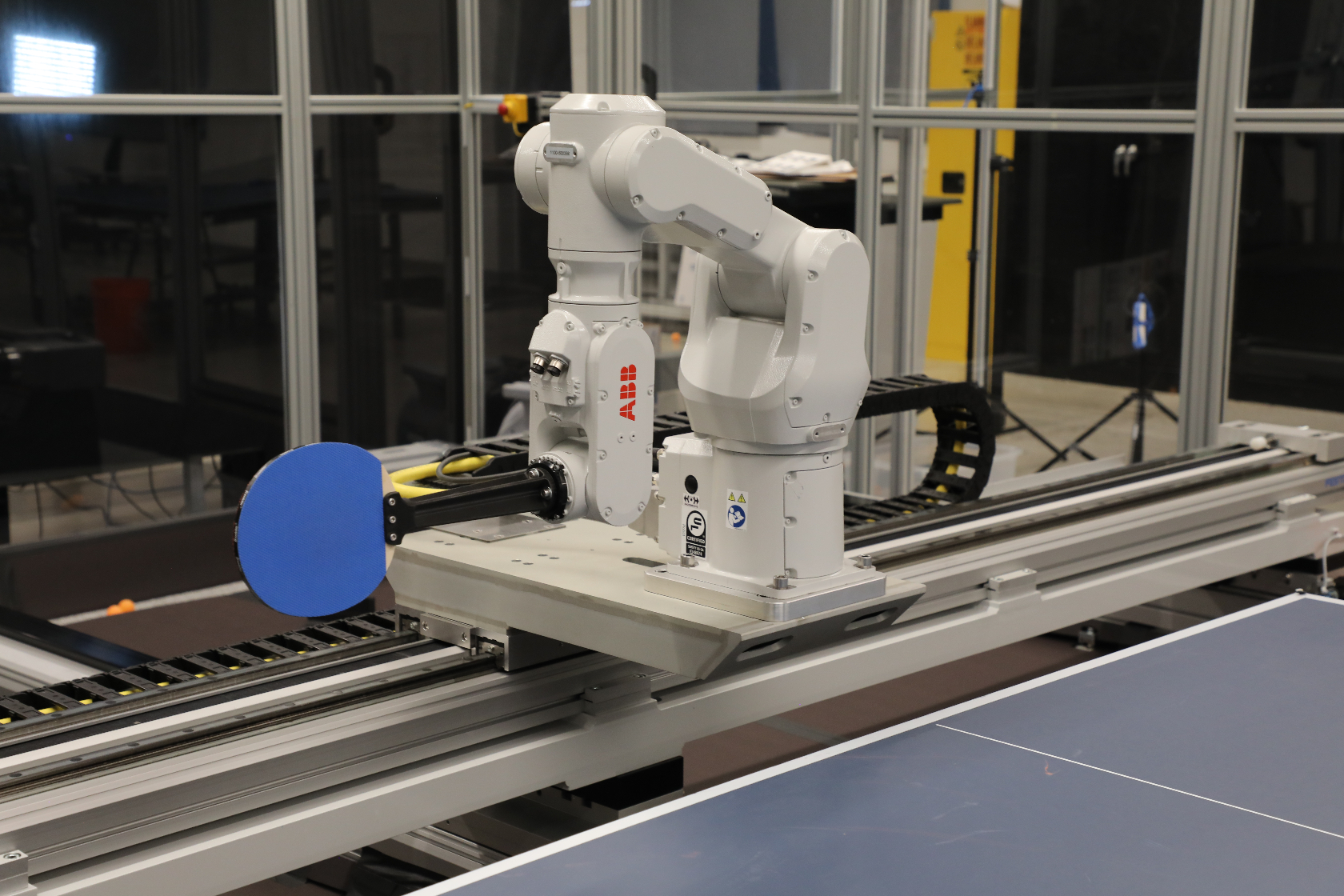}
    \includegraphics[width=0.30\textwidth]{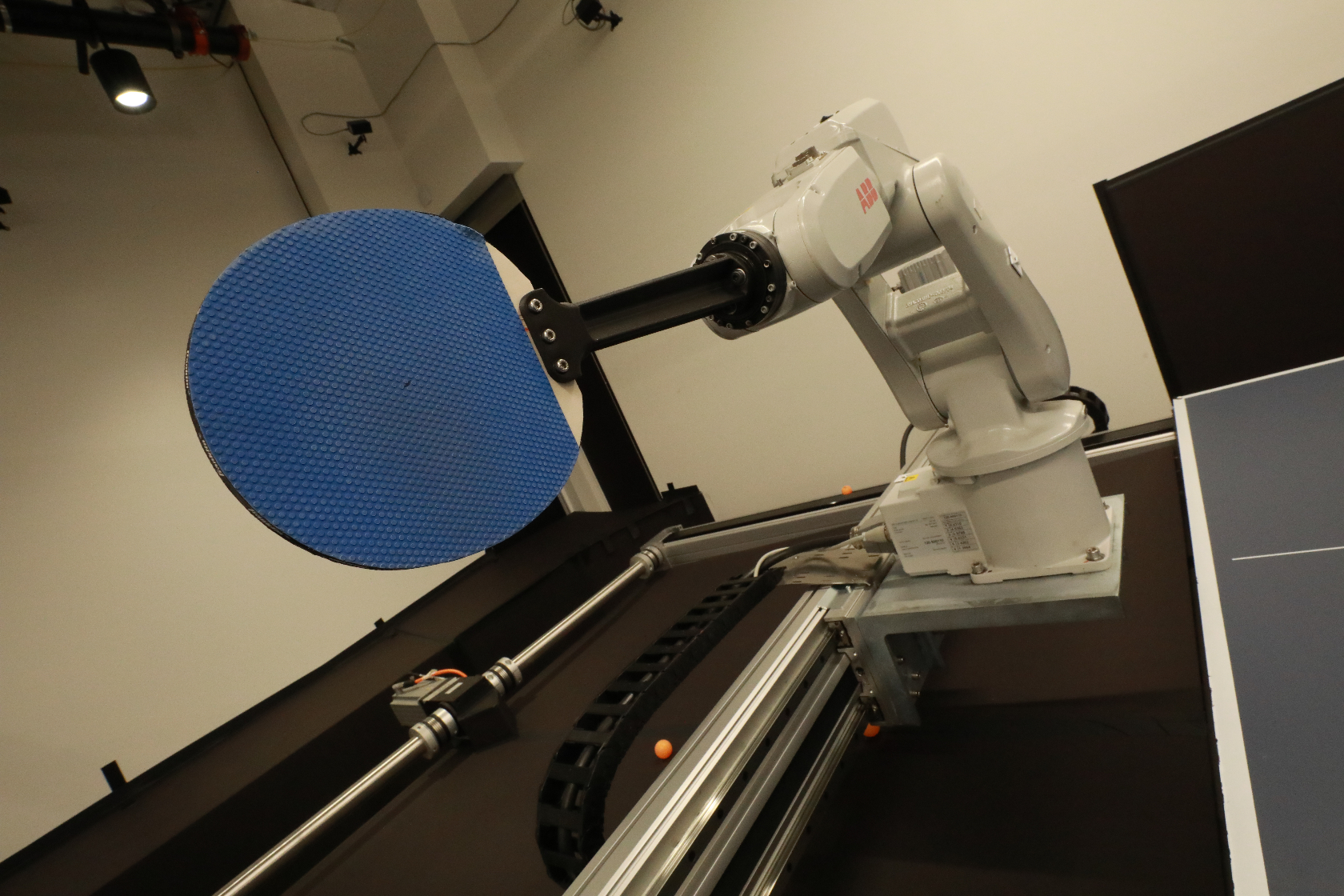}
  \caption{Two ABB arms --- ABB-IRB120T (left), ABB IRB 1100-4/0.58 (center) --- and their end effector (right). Most experiments in the paper use the 120, but the task space experiments in \autoref{sec:task_space_results} were able to transfer to the 1100 with minimal modifications despite the different joint layout.}
    \label{fig:abbs}
\end{figure*}

The configuration of arm and gantry was initially prototyped with a Widow-X arm kit and one-dimensional Zaber linear stage.  The Widow-X was comprised of several hobbyist Dynamixel servos.  The gripper the arm came with was made to hold a table tennis paddle. This prototype was nowhere near as capable as the current set of arms used in the system (see \autoref{fig:abbs}) but it was able to regularly return simple balls, enabling testing of many initial ideas. Ultimately a small-scale system like this was not meant to survive long term frequent use: the repeated hitting motions and impact of the balls would strip the delicate internal gears of the servos, which was a major factor in pursing industrial robots for reliability and robustness.

\subsubsection{Ball Thrower}

Existing consumer table tennis ball throwers offer a high level of customization and capability but require some sort of manual input to operate.  However, the construction of the devices are not easy to replicate.  Therefore all iterations of ball throwers in this system took some form of off-the-shelf device and made it more automated and robust.  Initially, reverse engineering, breadboards, and a programmable microcontroller were used to simulate the manual inputs through a USB interface.  Ultimately, a more robust system was required and a simpler thrower was obtained and almost all the electronic components of were replaced.  Aside from being more repairable and reliable, the customized thrower has higher quality parts including motors with encoders that can provide feedback to the system to alert if there is a failure or reduction in performance. A "throw detection" sensor has also been added in the form of two infrared sensors in the nozzle.  This sensor reports back when a ball has been thrown as well as an approximation of the speed, based on the time between when the two sensors were triggered.

Although the current thrower is more reliable, it is not built to the specifications of the industrial robots in the rest of the system. The two wheels that launch the balls make physical contact with them degrade over time and get dirty, requiring cleaning.
  
\subsection{Control Details}
\label{app:control}

\subsubsection{Additional Communication Protocols}

As discussed in \autoref{sec:control}, ABB arms are controlled through the Externally Guided Motion (EGM) interface \cite{egmmanual}.  However, the robot requires an additional interface Robot Web Services (RWS) provided by ABB to control basic functions.  RWS is a RESTful interface to the robot that allows access to various variables and subroutines on the robot with simple POST and GET requests.  The main usage of RWS is to reset the robot at the beginning of an experiment or when there is a problem, and to start the onboard EGM server.

\subsubsection{Python to C++ Interface}
\label{app:pybind}

The interface between Python and C++ is implemented with Pybind11 \cite{pybind11}. This library provides a convenient and compact syntax for calling C++ functions from Python and passing values in both directions between the languages. However, just wrapping function calls in the low-level driver with Pybind11 is insufficient because those functions would still execute in the Python thread, subjecting them to all the same performance constraints as regular Python threads. Releasing the GIL while executing the C++ function is possible but would not help due to the overhead of switching thread contexts.

Instead, a pure-C++ thread is started to handle low-level control for each of the Festo and ABB robots. These threads do not access any Python data directly, so they do not require the GIL and can execute in parallel to each other and the Python thread. Communication with these threads is done asynchronously via circular C++ buffers. C++ mutexes protect access to these buffers and ensure thread-safety, but they only have to be held briefly for read and write operations, not for an entire control cycle. This low-level controller can be used independently from Python (ie, in a pure-C++ program), but in this system, the circular buffers are accessed from Python via Pybind11-wrapped C++ function calls.

Each loop iteration of the low-level controller checks for a new command sent from Python. If none is available, it executes the remainder of the loop iteration with the previously sent command. Since the mutex protecting communication with the Python thread is only held briefly, this helps isolate the low-level controller from timing variation in the Python thread, thereby increasing robustness. Minimizing latency variation contributed more to improving performance than minimizing the absolute latency because the policies could learn to account for that latency in simulation.

\subsubsection{Robot Lockups and System Identification}
The ABB arms are primarily operated using the EGM interface. A sequence of commands are transmitted at 248Hz which includes a position and speed reference parameter per joint. The arms are sensitive to the commands and can lockup by tripping a hardware safety stop in several situations including:
\begin{enumerate}
    \item Physical collisions are detected
    \item Joints are predicted to exceed ranges
    \item Joints are being commanded to move too fast and have hit internal torque limits.
\end{enumerate}

These safety stops are controlled by an ABB proprietary hardware controller whose predictions are not accessible in advance so as to pro-actively avoid them. Safety lockups freeze the robot. At best they interrupt experiments, and at worst cause joints to physically go out of range requiring manual re-calibration. It was therefore important to implement mitigation in the control stack to prevent sending commands that would cause the robot to lockup. In addition, the actual movement of the arm in response to a position + speed reference command is internally processed by the hardware controller using a low pass filter + speed/position gain parameters. For the above reasons a system identification of the arms was performed to infer such parameters and uses them to both bring parity with simulation and prevent safety stops.

The agents produce velocity actions. In simulation these are directly interpreted by the PyBullet simulator. On the real robot system they are run through the safety simulator stack as described in \ref{sec:control}, providing a position and velocity per joint of the arm to reach. Directly using this result as a position + speed reference to the ABB arms does not faithfully move the arms through the same trajectory as seen in simulation. While the position portion of the command is accurate (following the exact intended), the speed reference parameter is interpreted differently by the hardware controller through gain parameters. The actual speed reference command is modeled as a combination of the velocity + torque, scaled by varying gain factors per component and joint. These gain factors were learned through a process of system identification, by optimizing them with a blackbox parameter tuning framework. The optimization objective was to minimize differences vs the trajectory described by the position portion of the commands. This tuning process replays some trajectories with different gain parameters to find the optimal way to set the speed reference portion. The result was fixed multiplicative gain factors that were primarily driven by the per-joint velocity as obtained from the safety simulator to use as the optimal speed reference command.

The problem of avoiding safety stops is mitigated in a few ways. First the safety simulator predicts collisions and sends the "bounced-back" commands so that they don't collide with environment as described earlier. Secondly to prevent exceeding joint ranges, a predictive system caps the speed reference portions of the command as the robot gets closer to the joint limits. The system predicts motion of the joint from the commanded position + an assumed inertial motion using the speed reference projected 250ms into the future and caps the speed ref portion of the command to prevent the predicted position from exceeding joint limits. This was modelled experimentally to identify the cases in which the hardware controller faults due to exceeding joint ranges, which helped discover this predictive window of 250ms. Lastly, over-torquing is minimized by reducing the safety simulator's max joint force limits.

\subsection{Simulation Details}
\label{app:simulator}

Four desiderata guided the design of the simulator.
\begin{enumerate}
    \item Compatibility with arbitrary policy training infrastructure so as to retain research flexibility, motivating the choice to conform to the Gym API.
    \item Flexibility, especially for components that researchers experiment with often, such as task definition, observation space, action space, termination conditions, and reward function.
    \item Simulation-to-reality gap is low. ``Low" means (1) algorithmic or training improvements demonstrated in simulation carry over to the real system and (2) zero-shot sim-to-real policy transfer is feasible. Perfect transfer is not required --- 80\%+ of simulated performance is targeted.
    \item Easy to apply domain randomization to arbitrary physical components.
\end{enumerate}

The design isolates certain components so they are easy to iterate on. For example, the tasks are encoded as sequences of states. Transitions between states are triggered by ball contact events with other objects. The task of a player returning a ball launched from a ball thrower is represented by the following two sequences. The first is $P1\_LAUNCH$ (the ball is in flight towards the player after being launched from the thrower) $\rightarrow$ $P1\_TABLE$ (the ball has bounced on the player's side of the table) $\rightarrow$ $P1\_PADDLE$ (the player hit the ball) $\rightarrow$ $P2\_TABLE$ (the ball landed on the opponent's side of the table) $\rightarrow$ $DONE\_P1\_WINPOINT$. The second is $P1\_LAUNCH$ $\rightarrow$ $P1\_TABLE$ $\rightarrow$ $P1\_PADDLE$ $\rightarrow$ $P1\_NET$ (the ball just hit the net) $\rightarrow$ $P2\_TABLE$ $\rightarrow$ $\rightarrow$ $DONE\_P1\_WINPOINT$. This accounts for the case where the ball first hits the net after being hit by the player and then bounces over and onto the opponent's side of the table. All other sequences lead to $DONE\_P1\_LOSEPOINT$.
For each task the complete set of \texttt{(state, event) $\rightarrow$ next\_state} triplets is enumerated in a config file. Tasks are changed by initializing the \texttt{StateMachine} with different configs. 

Another example is the reward function. It is common practice in robot learning, especially when training using the reinforcement learning paradigm, for the scalar reward per step to be a weighted combination of many reward terms. These terms are often algorithm and task dependent. Therefore it should be straightforward to change the number of terms, the weight per term, and to implement new ones. Each reward term is specified by name along with its weight in a config. The \texttt{RewardManager} takes in that config and handles loading, calculating, and summing each component. If a user wants to try out a new reward term, they write a reward class conforming to the \texttt{Reward} API (see below), which gets automatically registered by the \texttt{RewardManager}, and add it and its weight to the config. Over 35 different reward components have been tried $\approx$ 20 are in active use.

\subsubsection{Latency}

The latency of key physical system components were empirically measured as follows. Timing details are tracked starting with when the system receives hardware inputs (perception and robot feedback), through various transformations and IPCs (including policy inference), to when actual hardware commands are sent. This tracing gives a drill-down of latency throughout the stack with the ability to get mean and percentile metrics. The other half of the latency is how long the robot hardware takes to physically move after being given a command, which is separately measured by comparing time between a command being issued and receiving a feedback for reaching it. This completes the latency picture, covering the full time taken across the loop. See \autoref{table:latency-study-values} 100\% (baseline) column for the default latency values per component.

\subsubsection{Ball distributions}

Modeling the ball has two components, the dynamics and training distribution. PyBullet models the contact dynamics and the in-flight ball dynamics are modeled as in~\citet{abeyruwan2022sim2real}. Drag is modelled with a fixed coefficient of 0.47 but neither an external Magnus nor wind force is applied to the ball in the simulation. We refer readers to \cite{abeyruwan2022sim2real} Appendix C4 for more details on the in-flight ball model. The initial position and velocity of the balls are derived following~\citet{abeyruwan2022sim2real} and this determines the distribution of balls that are sampled during simulated training. See \autoref{table:ball-study-values} Thrower (baseline) column for the parameters of the default ball distribution used for training the BGS policies in this paper.

\subsubsection{Physical Parameters}

The restitution coefficients of the ball, table, and paddle, and the friction of the paddle are measured using the method from~\cite{DBLP:journals/corr/abs-2109-03100}. The mass of the ball and paddle is also measured. All other components have estimated un-tuned values or use PyBullet defaults. See \autoref{table:physical-study-values} Tuned (baseline) column for the values used.

\subsubsection{Gymnasium API}

The real world and simulated environments were developed according to the Gymnasium standard API for reinforcement learning\footnote{https://github.com/Farama-Foundation/Gymnasium}. Dictionary formats are used for observation and action specifications (for further details on individual components, see Appendix \ref{app:environment}). All extended environment functionalities are implemented as wrappers. In addition, the environments are compatible with agent learning frameworks, for example, TF-Agents \cite{TFAgents}, ACME \cite{acme}, Stable-Baselines3 \cite{stable-baselines3}, and so on. For consistent policy evaluation in simulation and hardware, TF-Agent’s Actor API is employed. All the supported policies (section \ref{sec:sim_training}) are wrapped in PyPolicy and integrated to Actor API. TF-Agents provides transformations to convert the Gymnasium environment to a PyEnvironment.

\subsubsection{Reward API}

The \texttt{Reward} class API is outlined below. Latex code style from \cite{kingma2018glow}.

\begin{lstlisting}[language=Python]
class Reward:

  def __init__(self, *kwargs):
  # Initializes the reward class.
    
  def compute_reward(self, state_machine_data, done, *kwargs):
  # Computes and returns the reward per step along with a dictionary which optionally contains reward specific  information.
    return reward, reward_info
    
  def reset(self, done):
  # Resets any stored state as needed when the episode is done.
\end{lstlisting}
 
\subsection{Perception Details}
\label{app:perception}

\subsubsection{Cameras and Calibration}

Previous iterations of the system included a variety of other camera types and positions. Larger arrays of slower cameras were effective during prototyping for basic ball contact, but struggled on tasks that required more accurate ball positioning. Adding more cameras to the current setup could produce still more accurate position estimations, but there start to be bandwidth limitations on a single machine and it may require remote vision devices (increasing latency and system complexity) or switching away from USB3.

Cameras are calibrated individually for intrinsic parameters first and later calibrated extrinsically to the table coordinate system via sets of AprilTags \cite{apriltag} placed on the table. Both calibrations are done with APIs provided by OpenCV \cite{opencv_library}. Calibration is an important factor in the performance of the system given the small size of the ball it needs to track. While it is relatively easy to get decent camera performance with basic camera knowledge, it required the help of vision experts to suggest hardware solutions like lens spacers and locks as well as calibration tools such as focus targets to get truly stable performance.

\subsubsection{Patch based Training Data}
\begin{figure}
    \centering
    \includegraphics[width=0.24\textwidth]{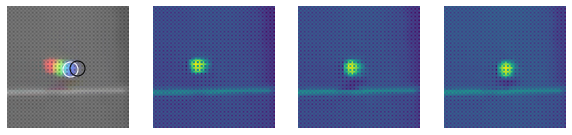}
    \includegraphics[width=0.24\textwidth]{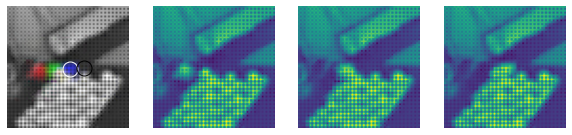}
    \includegraphics[width=0.24\textwidth]{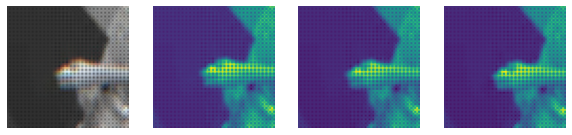}
    \includegraphics[width=0.24\textwidth]{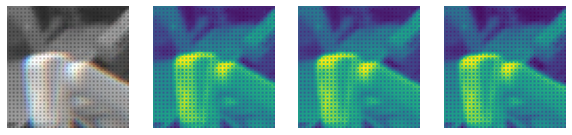}
    \caption{Examples of training patches for the Ball detector which consist of the past three frames. The left-most RGB formatted patch is for visualization purposes to highlight the motion of the ball in play with the current and next labeled position indicated with white and black circles respectively. The three single channel images to the right of the RGB image show the raw Bayer pattern as expected by the detector. Top row shows two sequences of three frames centered on the final ball position (modulo 2 to match Bayer stride). Bottom row shows hard negative examples where the center position contains a bright spot with some motion originating from a person carrying a ball-in-hand or from the robot itself.}
    \label{app:bayer-patches}
\end{figure}

\autoref{app:bayer-patches} shows some examples of the patches extracted from the raw Bayer images used to train the ball detector network. These patches are centered on the ball in the current frame where the two previous frames are included to prime the temporal convolutional layers within our custom architecture.

\subsubsection{Ball Detector Network Structure}

The spatial convolution layers capture shape and color information whilst down-sampling the image size to reduce computation. Operating on the single channel raw images means it is important that the first layer has a $2 \times 2$ stride matching the Bayer pattern, so that the weights of the convolutional kernel are applied to the same colored pixels at all spatial locations. In total, five convolutional layers are applied with the first three including batch normalization \cite{ioffe2015batch} before a ReLU activation \cite{fukushima1969visual, pmlr-v15-glorot11a}. Two of these layers employ a buffered temporal mechanism resulting in an extremely compact network overall with only 27K parameters. Full details of the architecture is shown in Table \ref{table:detector_architecture}. Note that the shape is represented as $(B, H, W, C)$ corresponding to the typical batch size, height, width and channels, however during inference the batch is set to the number of cameras. Also note that in contrast to typical temporal convolutions operating on video data there is no time dimension. Instead the temporal convolutional layers simply concatenate their previous input to the current features along the channel dimensions. Here the next convolutional layer with weights will effectively span two timesteps.

\begin{table*}[h!]
\small
\centering
\bgroup
\def\arraystretch{1}%
\begin{tabular}{|lccccr|}

\hline
Layer Type & Kernel Size & Strides & Dilation Rate & Output Size & \# Trainable Parameters \\

\hline \hline
Input & -- & -- & -- & $(2, 512, 1024, 8)$ & -- \\
2D Spatial Convolution & 4 & 2 & 1 & $(2, 256, 512, 8)$ & 128 \\
Batch Norm & & & & & 16 \\
Buffered Temporal Convolution & -- & -- & -- & $(2, 256, 512, 16)$ & -- \\
2D Spatial Convolution & -- & -- & 2 & $(2, 256, 512, 8)$ & 1152 \\
Batch Norm & & & & & 16 \\
2D Spatial Convolution & 4 & 2 & 1 & $(2, 128, 256, 16)$ & 2048 \\
Batch Norm & & & & & 32 \\
Buffered Temporal Convolution & -- & -- & -- & $(2, 128, 256, 32)$ & -- \\
2D Spatial Convolution & -- & -- & 1 & $(2, 128, 256, 64)$ & 18496 \\
Dropout (drop-rate=0.1) & -- & -- & -- & $(2, 128, 256, 64)$ & -- \\
Prediction Head & 4 & -- & 2 & $(2, 128, 256, 5)$ & 5125 \\

\hline \hline

\multicolumn{1}{l}{Optimizer} & \multicolumn{3}{l}{Adam ($\alpha = 1\mathrm{e}{-4}$, $\beta_1 = 0.9$, $\beta_2 = 0.999$)}  \\
\multicolumn{1}{l}{Learning Rate Schedule} & \multicolumn{5}{l}{Linear ramp-up (5000 steps) then exponential decay.} \\
\multicolumn{1}{l}{Batch size} & \multicolumn{5}{l}{128} \\
\multicolumn{1}{l}{Weight decay}  & \multicolumn{5}{l}{None} \\

\hline

\end{tabular}
\egroup
\caption{Ball Detector, Architecture and Training Details. 
All layers employ ReLU non-linearities.}
\label{table:detector_architecture}
\end{table*}

\subsubsection{Tracking Performance}
To assess tracking performance independently of the downstream processes, the output of the perception pipeline is compared against human annotated ball positions. These annotations capture the ball's image position in each camera view for the entire duration the ball is considered in-play, i.e. has not touched the ground or any object below the table height. Both views are recombined with annotations, triangulated to their 3D position, and stitched over time into 3D trajectories.

For tracking evaluation the 3D trajectories of the ball across 10 annotated sequences are used as target reference positions consisting of 514 annotated trajectories over 93,978 frames. To measure the alignment of predicted trajectories to these annotations, the recently proposed Average Local Tracking Accuracy (ALTA) metric \cite{valmadre2021local} is applied by defining a true positive detection as the predicted 3D position of the ball at time $t$ to correspond to with in 5cm of the annotated position in frame $t$. Since the temporal aspect of the tracking problem is important from both a short history as used by the policy and a longer history for locating hits and bounces by the referee (Section \ref{sec:real_env}) the temporal horizon of ALTA is set to 100 frames with the results reported in Table \ref{tab:perception_results_summary}. These results show the benefit that hard negative mining can bring to a patch-based training method.

\begin{table}
\centering
\scriptsize
\begin{tabular}{@{}||l|ccc||c@{}}
 \hline
 Training Data Source  & ATA &  ATR & ATP \\ \hline \hline
 With HNM & 66.4 \% & 69.0 \% & 64.0 \% \\
 Without HNM & 58.5 \% & 64.0 \% & 53.8 \% \\ \hline
\end{tabular}
\caption{Ball tracking performance comparing different training datasets including hard negative mining (HNM). Average tracking accuracy (ALTA) is the key metric used for tracking quality over the local temporal horizon of 100 frames with a $< 5cm$ true positive criteria. ATR and ATP denote the average tracking recall and precision respectively \cite{valmadre2021local}.}
\label{tab:perception_results_summary}
\end{table}

\subsection{Real World Details}
\label{app:environment}

\subsubsection{Referee}

The primary role of the Referee is to generate ball and robot contacts to drive the \texttt{StateMachine}, \texttt{RewardManager}, and \texttt{DoneManager} as defined in \autoref{sec:sim_training}. The different contact events are as follows; \texttt{TABLE\_ARM} (ball contact with robot side of the table), \texttt{TABLE\_OPP} (ball contact with opponent side of the table), \texttt{PADDLE\_ARM} (ball contact with robot paddle), \texttt{NET} (ball contact with net), \texttt{GROUND} (ball contact with the ground), \texttt{TABLE} (robot contact with table), and \texttt{STAND} (robot contact with stand). The real environment and the referee communicate using a custom MPI (ROS \cite{ros_288} is an alternative), where the Referee initializes a server and the environment uses a client to request reward, done and info at step frequency. The Referee updates its internal state at 100 Hz regardless of the step frequency.

\subsubsection{Real gym environment}

The real environment interfaces with the policy and hardware. In addition to the real world challenges described in \autoref{sec:real_env}, it must also ensure the environment step lasts for the expected duration given the environment Hz. An adaptive throttling function facilitates this. The throttling function is initialized with the first observation. When the next step call completes, the throttler waits for the remaining time of the environment timestep before returning. If the next step call consumes more computational time than the timestep budget, the throttler advances to the next nearest multiple of the timestep. Both the actor and environment run on the same thread, therefore, the timestep also consumes the computational time required by the policy. A recommendation is that if the policy requires more computational time than timestep budget, either reduce the step frequency or use an asynchronous policy framework.

\subsubsection{Subprocesses}

The actor, environment, and referee components are implemented using Python, therefore, process speedup is limited by the GIL. Threading increased process contention and the sim-to-real gap. If a process is known to get throttled by thread contention or a high computational workload, the code should be distributed to a different process.

\subsubsection{Observation Filters}

The system uses the Savitzky-Golay FIR filter \cite{aitken_1961} for observation smoothing in \texttt{TableTennisRealEnv} and \texttt{Referee}. The filter coefficients were calculated once for a window length of 9 and a 1-D convolution is applied for each sensor modality independently. For boundary values, input is extended by replicating the last value in the buffer.

The real environment uses interpolation/extrapolation to generate observations for the given timestep. Noise from the sensor can cause the interpolated/extrapolated values to show a zig-zag pattern. In some cases, where a false positive ball observations occurs, the calculated value does not generate an observation within the expected observation range. Feeding these observations to real policy tends to produce jittery or unsafe actions. Similarly, \texttt{Referee} filters raw observations prior to being used to calculate ball contacts. 

\begin{figure*}
    \centering
    \includegraphics[width=\linewidth]{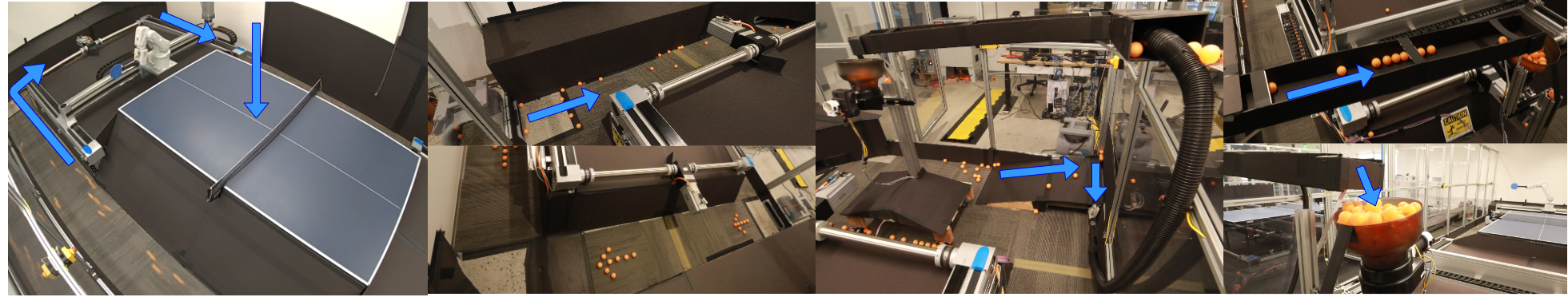}
    \caption{Ball collection system. Images from left to right. (I) Air blows down from a ceiling mounted fan, pushing any balls on the table down to the floor. (II-top) \& (II-bottom) Blower fans at each corner push the balls around the table. (III)  At one corner of the table is a ramp that guides the balls to a tube... (IV-top) ...where air pushes them to a higher ramp... (IV-bottom) ...which returns balls to the thrower's hopper.}
    \label{fig:ball_collection}
\end{figure*}

\subsection{Training Parameters}

\begin{table}[!t]
\centering
\scriptsize
\begin{tabular}{||l c ||}
\hline
Hyper-parameter & Value  \\
\hline \hline
Number of directions ($\delta$), $N$ &  200\\
Number of repeats per direction, $m$ & 15\\
Number of top directions, $k$ &  60\\
Direction standard deviation, $\sigma$ &  0.025\\
Step size, $\alpha$ & 0.00375 \\
Normalize observations &  True \\
Maximum episode steps & 200 \\
Training iterations & 10,000 \\
\hline
\end{tabular}
\caption{Hyper-parameters used for training BGS policies in simulation.}
\label{table:bgs-hyper-params}
\end{table}

\begin{table}[!t]
\centering
\scriptsize
\begin{tabular}{||l c c||}
\hline
Reward & Min. per episode & Max. per episode  \\
\hline \hline
Hit ball & 0 & 1.0 \\
Land ball & 0 & 1.0 \\
Velocity penalty & 0 & 0.4 \\
Acceleration penalty & 0 & 0.3 \\
Jerk penalty & 0 & 0.3 \\
Joint angle & 0 & 1.0 \\
Bad collision & -1 & 0 \\
Base rotate backwards & -1 * timesteps & 0 \\
Paddle height & -1 * timesteps & 0 \\
\hline
Total & variable & 4.0 \\
\hline
\end{tabular}
\caption{Rewards used for training BGS policies in simulation.}
\label{table:bgs-rewards}
\end{table}

\autoref{table:bgs-hyper-params} contains the hyper-parameter settings and \autoref{table:bgs-rewards} details the rewards used for training the BGS policies in this paper. A brief description of each reward is given below.
\begin{itemize}
    \item Hit ball: +1 if the policy makes contact with the ball, 0 otherwise.
    \item Land ball: +1 if the policy successfully returns the ball such that it crossed the net and lands on the opposite side of the table.
    \item Velocity penalty: 1 - \% points (timesteps * number of joints) which violate the per joint velocity limits: [1.0, 2.0, 4.5, 4.5, 7.6, 10.7, 14.5]$m/s$.
    \item Acceleration penalty: 1 - \% points (timesteps * number of joints) which violate the per joint acceleration limits: [0.2, 0.2, 1.0, 1.0, 1.0, 1.5, 2.5, 3.0]$m/s^2$.
    \item Jerk penalty: 1 - \% points (timesteps * number of joints) which violate the per joint jerk limits: [0.92, 0.92, 1.76, 0.9, 0.95, 0.65, 1.5, 1.0]$m/s^3$.
    \item Joint angle penalty: 1 - \% points (timesteps * number of joints) which lie outside the joint limits (minus a small buffer).
    \item Bad collision: -1 per timestep if the robot collides with itself or the table, 0 otherwise. The episode typically ends immediately after a bad collision, hence the minimum reward of -1.
    \item Paddle height penalty: -1 for every timestep the center of the paddle is $<$12.5cm above the table, 0 otherwise.
    \item Base rotate backwards penalty: -1 each timestep the base ABB joint has position $<$-2.0 (rotated far backwards).
\end{itemize}

\subsection{Simulator Parameter Studies: Additional Results \& Details}
\label{app:sim-ablations-details}

This section contains additional details about the simulator parameter studies. First we discuss additional results and then give details of all the parameter values for each study.

\subsubsection{Additional Results}

\begin{figure}
    \centering
    \includegraphics[width=0.24\textwidth]{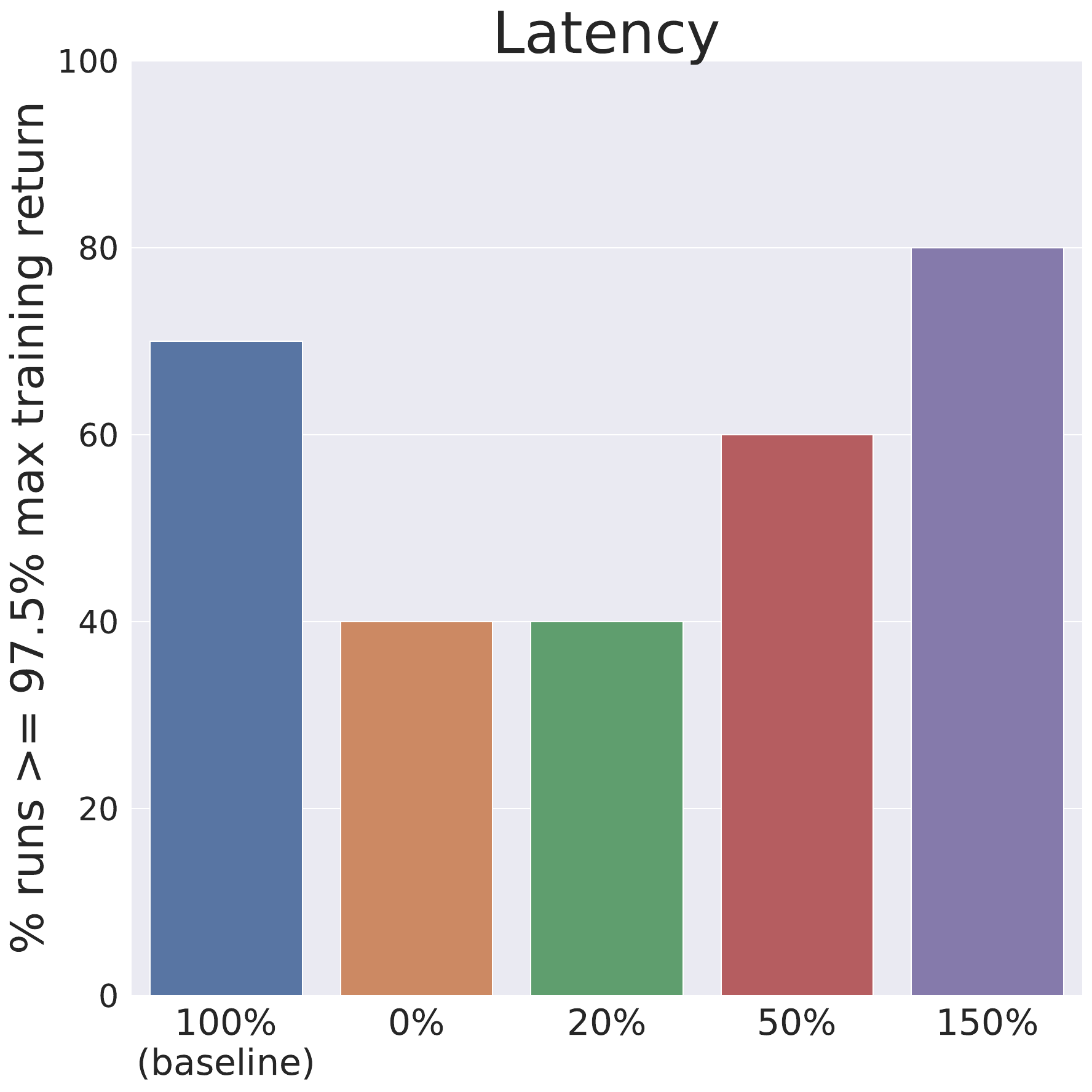}
    \includegraphics[width=0.24\textwidth]{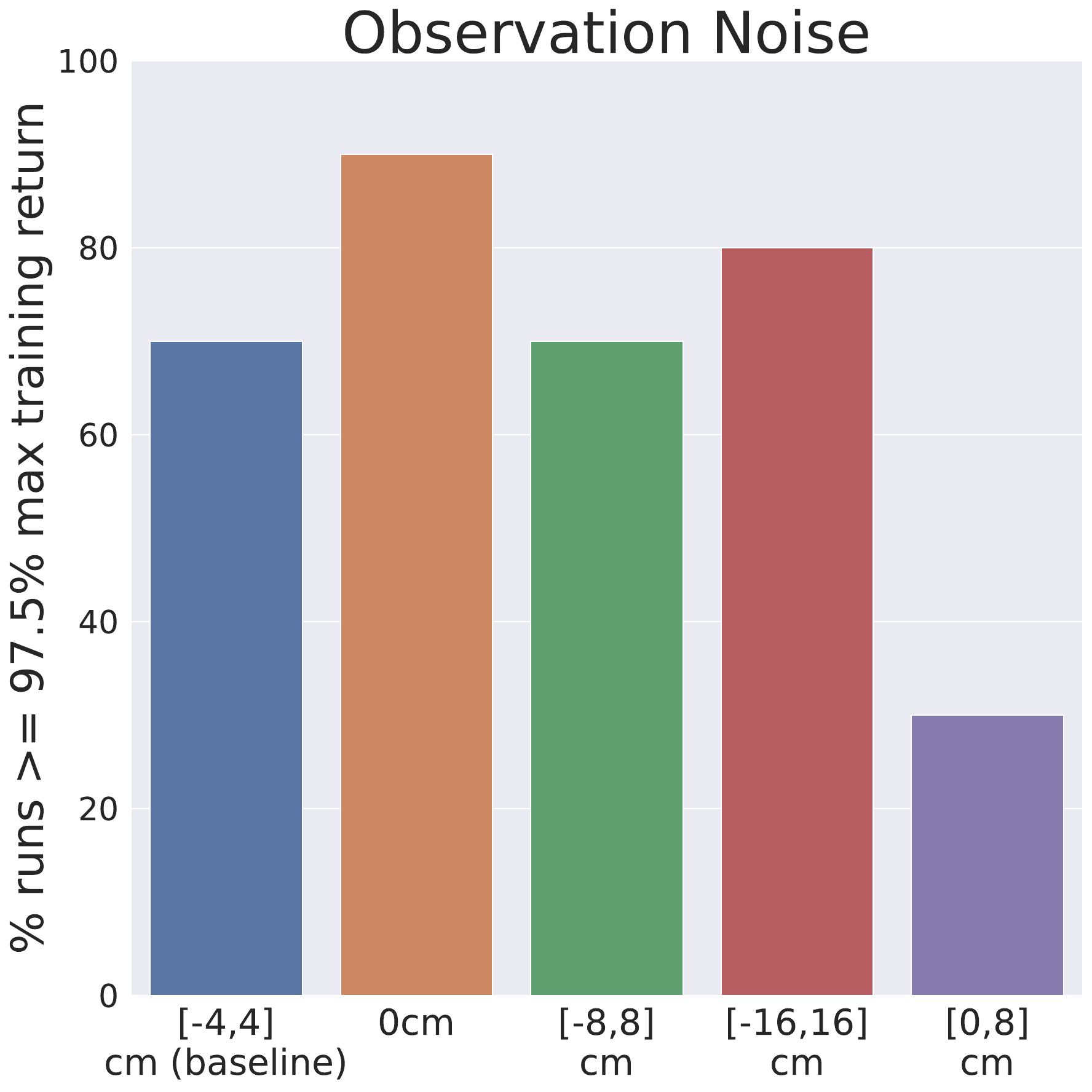}
    \includegraphics[width=0.24\textwidth]{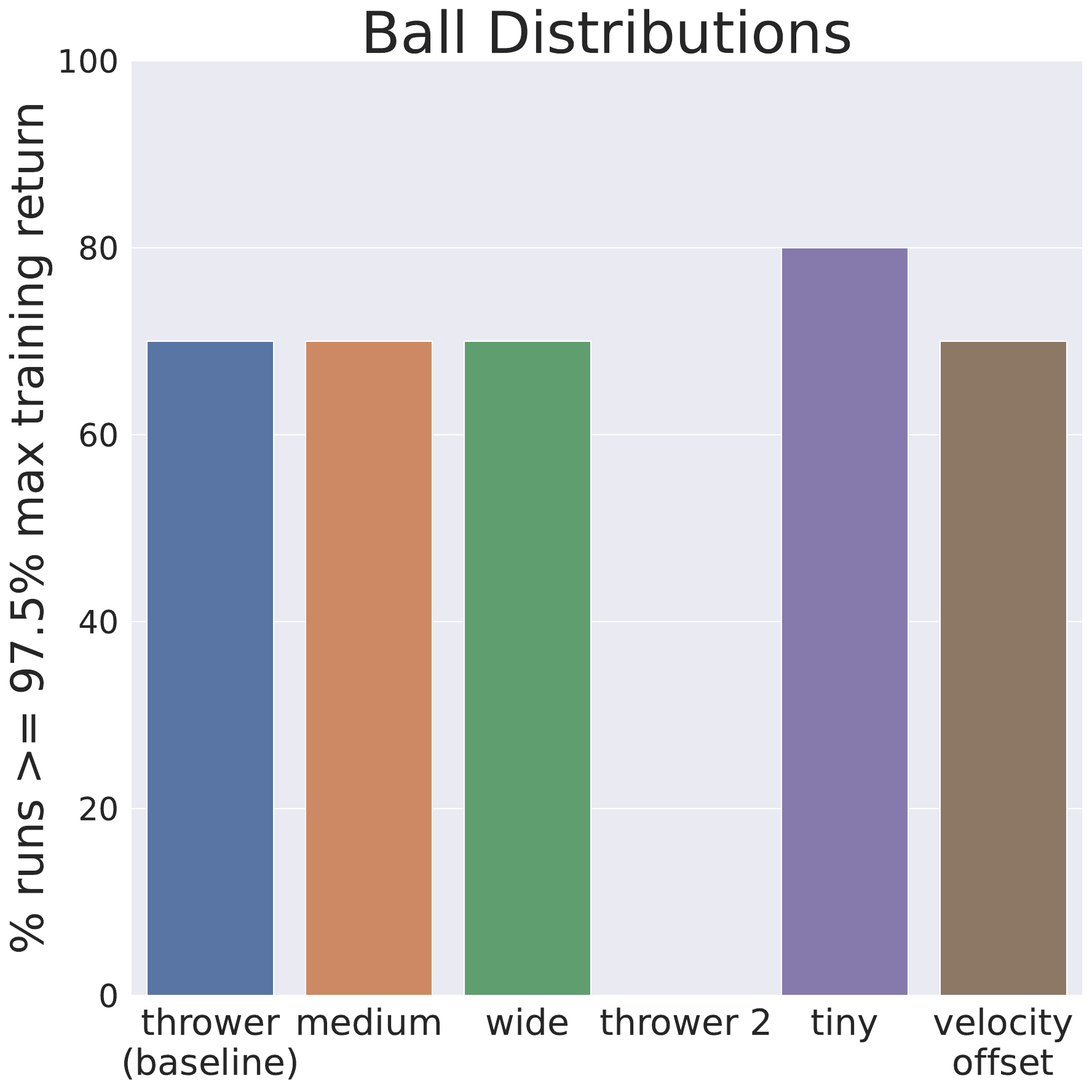}
    \includegraphics[width=0.24\textwidth]{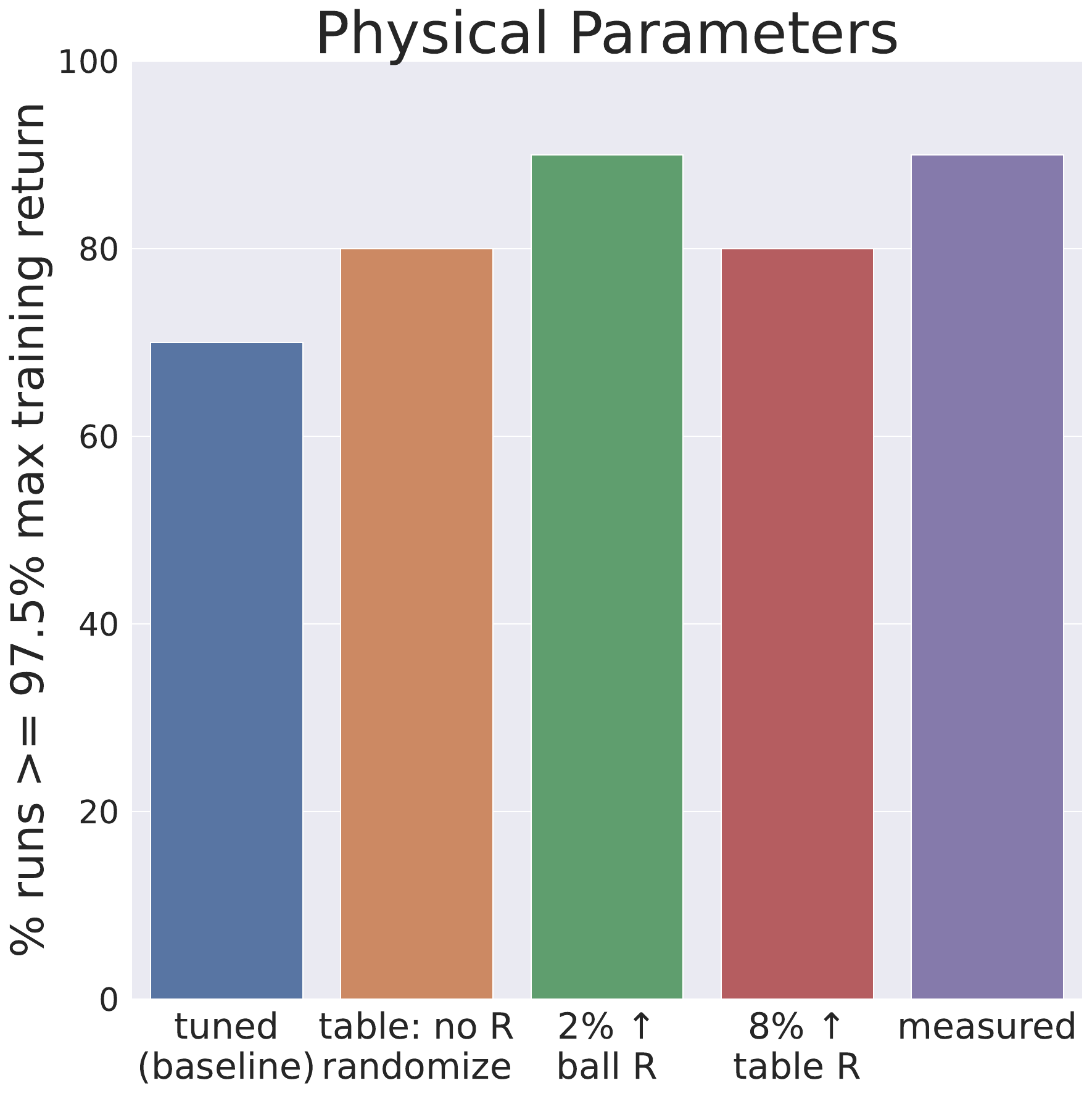}
    \caption{Effect of simulator parameters on training quality defined as \% runs $\ge97.5$ maximum reward. Most settings are similar to baseline performance of 70\%. Notably very low latency (0-20\%), biased observation noise ([0,8]cm), and large ball distributions (thrower 2) make the task harder and reduce the \% of runs that trained well within 10k training steps. R = restitution coefficient.}
    \label{app:sim-ablations-quality}
\end{figure}

We present additional results from the simulator parameter studies. In \autoref{ablations:sim-params} we assessed the effect of varying simulator parameters on the zero-shot sim-to-real transfer performance. Here we discuss the effect on training quality, defined as the percentage of the 10 training runs that achieved $\ge$97.5\% of the maximum score during training. Agents with scores above this threshold effectively solve the return ball task. The results are presented in \autoref{app:sim-ablations-quality}.

Training contains significant randomness from two main sources. First the environment has multiple sources of randomness; primarily from the ball distribution, latency sampling and observation noise, but also from domain randomization of some physical parameters and small perturbations to the robot's initial starting position. The extent of the environment randomness in each of these areas is affected by parameter values. Second, randomness comes from the training algorithm, BGS. During each BGS training step, directions are randomly sampled and a weighted average of a subset of these forms the update to the parameters.

Additionally, there appears to be distinct learning phases for the return ball task, with corresponding local maxima. Two common places for training to get stuck are (1) a policy never learns to make contact with the ball and (2) a policy always makes contact with the ball but never learns to return it to the opposite side.

Consequently, we observe that about 70\% of training runs with the baseline parameters settings solve the problem. Substantially reducing latency to 0-20\% appears to make the task harder. Only 40\% of runs in these settings train well (see \autoref{app:sim-ablations-quality} (top left)). Removing observation noise (see \autoref{app:sim-ablations-quality} (top right)) makes the problem easier, with 90\% runs training well. Increasing zero-mean noise does not affect training quality for the settings tested, however introducing biased noise does appear to make the problem much harder, with only 30\% runs training well. Changing the ball distribution (see \autoref{app:sim-ablations-quality} (bottom left)) does not have a meaningful impact on training quality except in one case. All training runs for the different thrower distribution (thrower 2) failed to reach the threshold in 10k steps. This is likely because the ball distribution is more varied than baseline, medium, or wide distributions (see \autoref{fig:ball_dists_viz}). However no policies got stuck in local maxima. All achieved 93\% of the maximum reward and 50\% achieved 95\%. This is unlike the low latency or biased observation noise settings where 20-60\% runs got stuck in local maxima. Finally, changing the values of different physical parameter settings appears to make the task slightly easier, with all experiments having 80-90\% of runs that trained well compared with the baseline 70\% (see \autoref{app:sim-ablations-quality} (bottom right)).

\begin{figure}
  \begin{center}
  \includegraphics[width=0.5\textwidth]{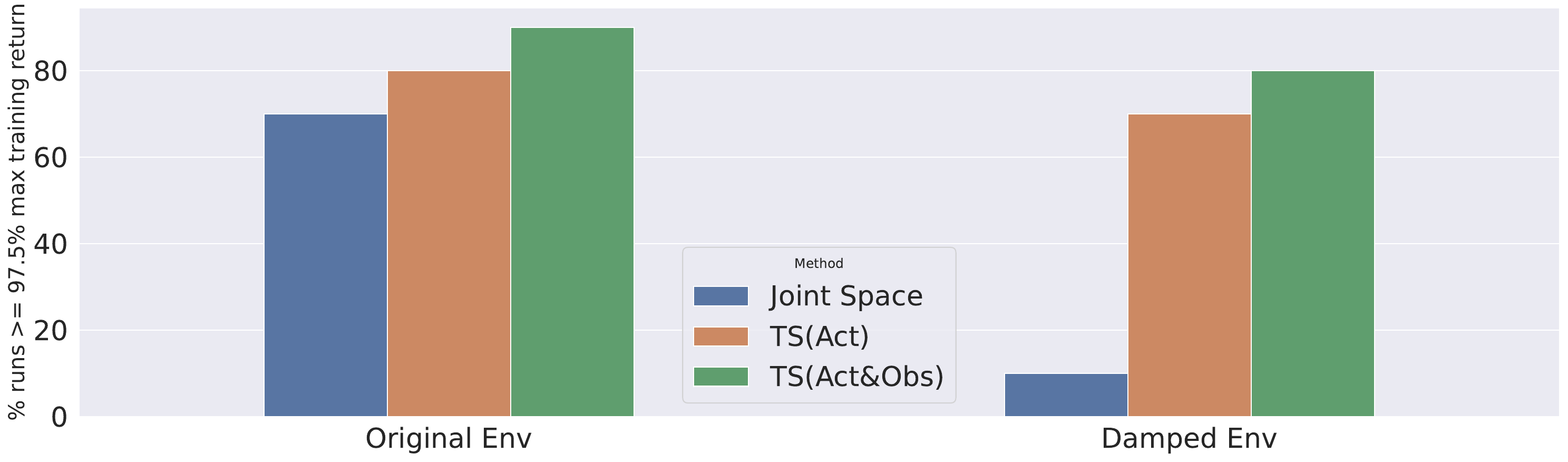}\\
  \includegraphics[width=0.5\textwidth]{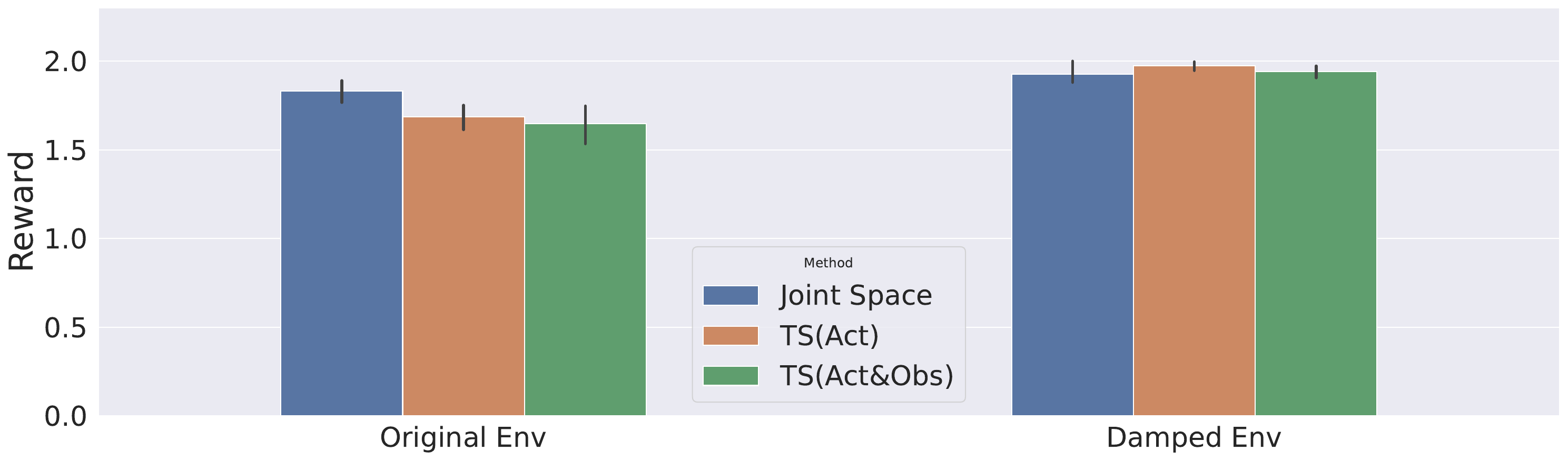}
\end{center}
\caption{(Top) Percent of seeds that solve the task (\% runs $\ge$ 97.5\% max training return) when trained in task space, combined with the damped (harder) environment. (Bottom) Zero shot transfer results of the seeds that succeeded the training.}
\label{fig:task_space_app}
\end{figure}

\subsection{Simulator Parameter Studies: Physical parameter measurements, revisited}
\label{app:system_id_revisited}

\begin{figure}
    \centering
    \includegraphics[width=0.48\textwidth]{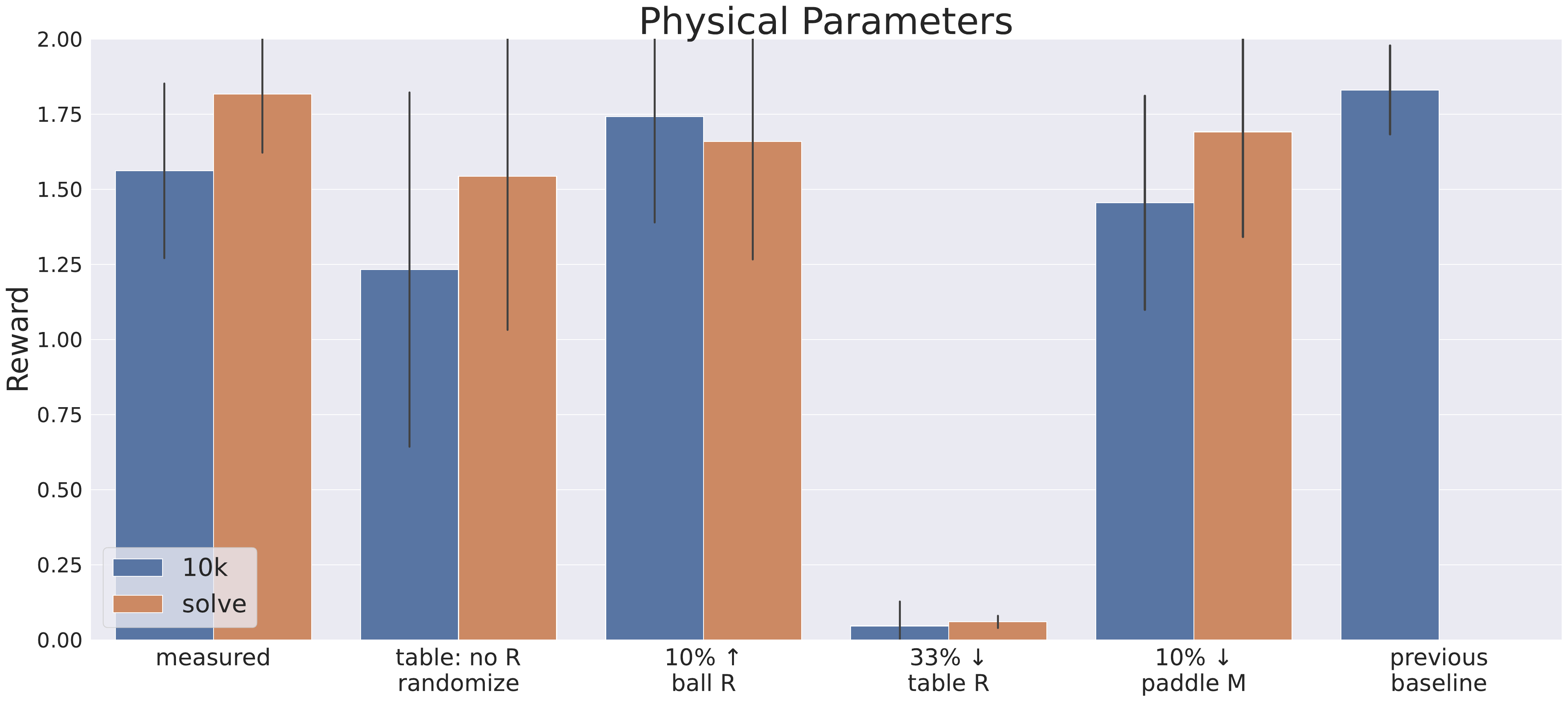}
    \includegraphics[width=0.48\textwidth]{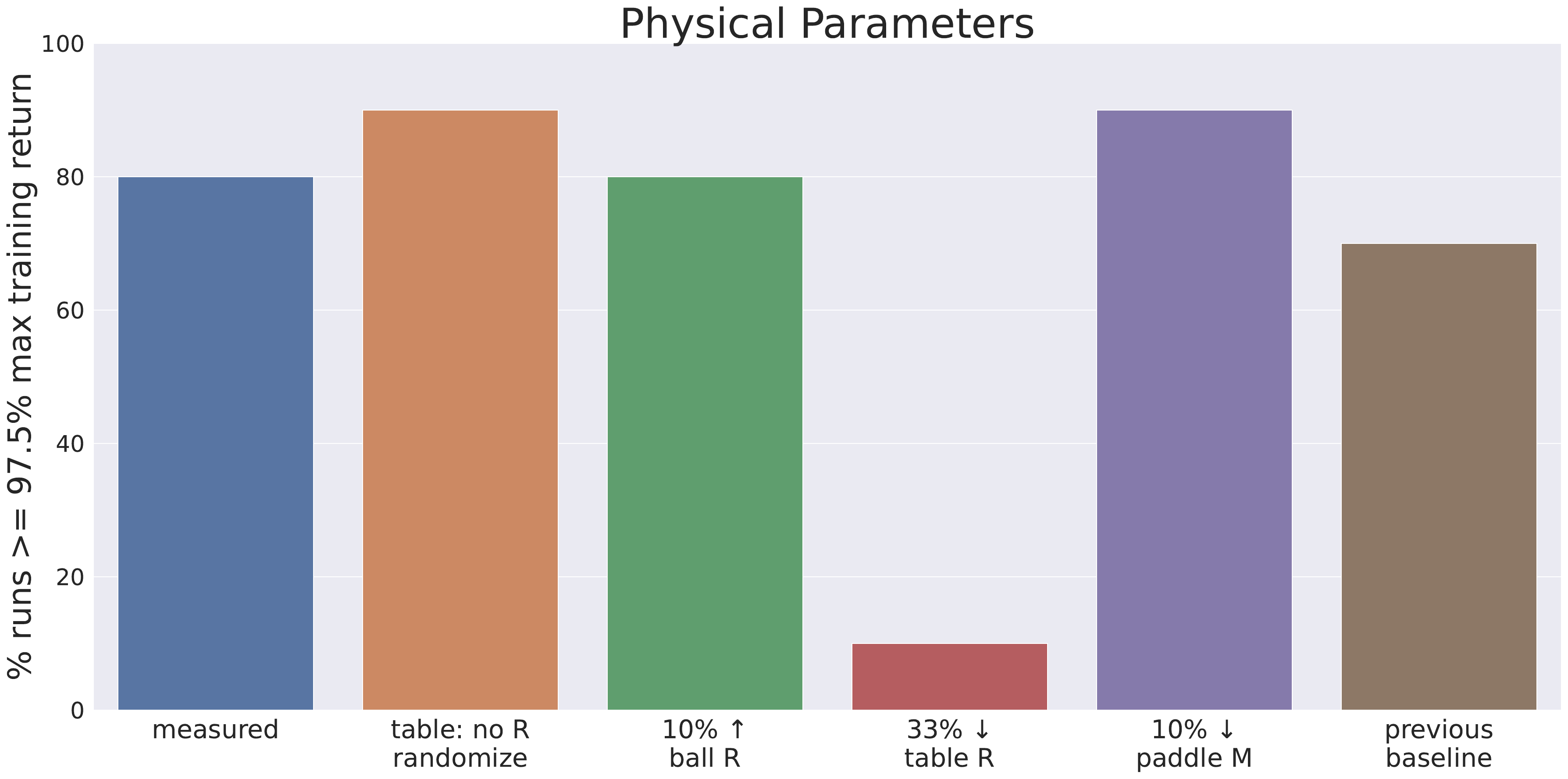}
    \caption{After making a number of improvements to the system, zero-shot sim-to-real transfer scores 1.82 (measured - solve) on average whilst following a principled procedure for measuring physical parameter values used in simulation. This is on par with the previous baseline presented in \autoref{fig:sim-params-ablations} reproduced here as \textit{previous baseline}. \textbf{Top} Mean reward (with 95\% CIs) on zero-shot real world evaluation. Measured physical parameters performed best. 10k = evaluation at 10k steps. solve = evaluation at earliest step $<$ 10k which solved the task ($\ge 97.5$ maximum training return). Policies evaluated at the solve step had slightly higher performance on average. R = restitution coefficient. M = mass. \textbf{Bottom} Percent of seeds that solve the task.}
    \label{app:sim-ablations-post-fix}
\end{figure}

It was unsatisfactory not to follow the process outlined in Appendix \ref{app:simulator} to set physical parameter values in the simulator. In \autoref{ablations:sim-params} we hypothesized this was due to not modeling spin correctly. To investigate this, we modeled spin in the simulator following the method from \cite{abeyruwan2022sim2real}. We extended the simulation ball model to incorporate the magnus force. Then we collected a set of ball trajectories from a ball thrower. For each trajectory we set up an optimization problem to solve for the initial position, linear velocity, and angular velocity of the ball, and used these values to derive ball distributions in simulation.

During the same time period we also changed the paddle on the robots to a Tenergy 05\footnote{https://butterflyonline.com/Templates/RubberSpecifications.pdf}. This paddle has a softer and higher friction surface than the previous paddle and can impart substantially more spin on the ball. We re-measured the physical parameters of the system following the process described in Appendix \ref{app:simulator}. We performed a grid search over the restitution coefficient of the paddle, setting the other parameters to measured values, to find the value that resulted in the best zero-shot sim-to-real transfer. The grid search was necessary because the new paddle surface is soft but is modeled in simulation as a rigid body. Thus we use the restitution coefficient to approximate `softness`. Values are detailed in \autoref{table:physical-study-values} (see column "Re-measured post changes"). Note that the paddle restitution coefficient that led to the best transfer, 0.44, is much lower than the measured value of 0.84. Finally, we observed that simulated training was harder in this setting, likely due to higher finesse required to return the ball with the new paddle. To remedy this we added a small bonus for hitting the ball towards the opponent's side of the table even if the policy did not return it over the net. The reward increases proportionally to how close to ball was to the net.

After making these changes, we re-ran the physical parameter study and present the results in \autoref{app:sim-ablations-post-fix}\footnote{2 / 50 seeds got stuck. One for table no R randomize, and one for -10\% paddle M. These results are reported over 9 instead of 10 seeds. We do not think this materially affects any of the findings.}. We find that zero-shot sim-to-real transfer is on par with the previous baseline (see top chart, measured vs previous baseline). We also observe that evaluating policies at the checkpoint when they first solve the task (i.e. first checkpoint to score $\ge 97.5\%$ maximum training return) perform slightly better on average than evaluating at the end of training, after 10k training steps. However the difference is not statistically significant.

Performance loss by not randomizing the table restitution is similar to our original study, however (fortunately) performance is less sensitive to small changes in parameter values. For example, increasing the ball restitution by 10\% or reducing the paddle restitution by 10\% only led to a small reduction in performance. However large changes may lead to performance collapse as indicated by reducing the table restitution coefficient by 33\%.

With the addition of the extra reward component, most policies solve the task in most settings (see \autoref{app:sim-ablations-post-fix}, bottom). The exception is when the table restitution coefficient was reduced by 33\%, reducing the bounciness of the incoming ball and likely making the task very difficult to learn.

\subsection{Simulator Parameter Studies: Study values}

\begin{figure*}
\centering
\includegraphics[width=0.28\textwidth]{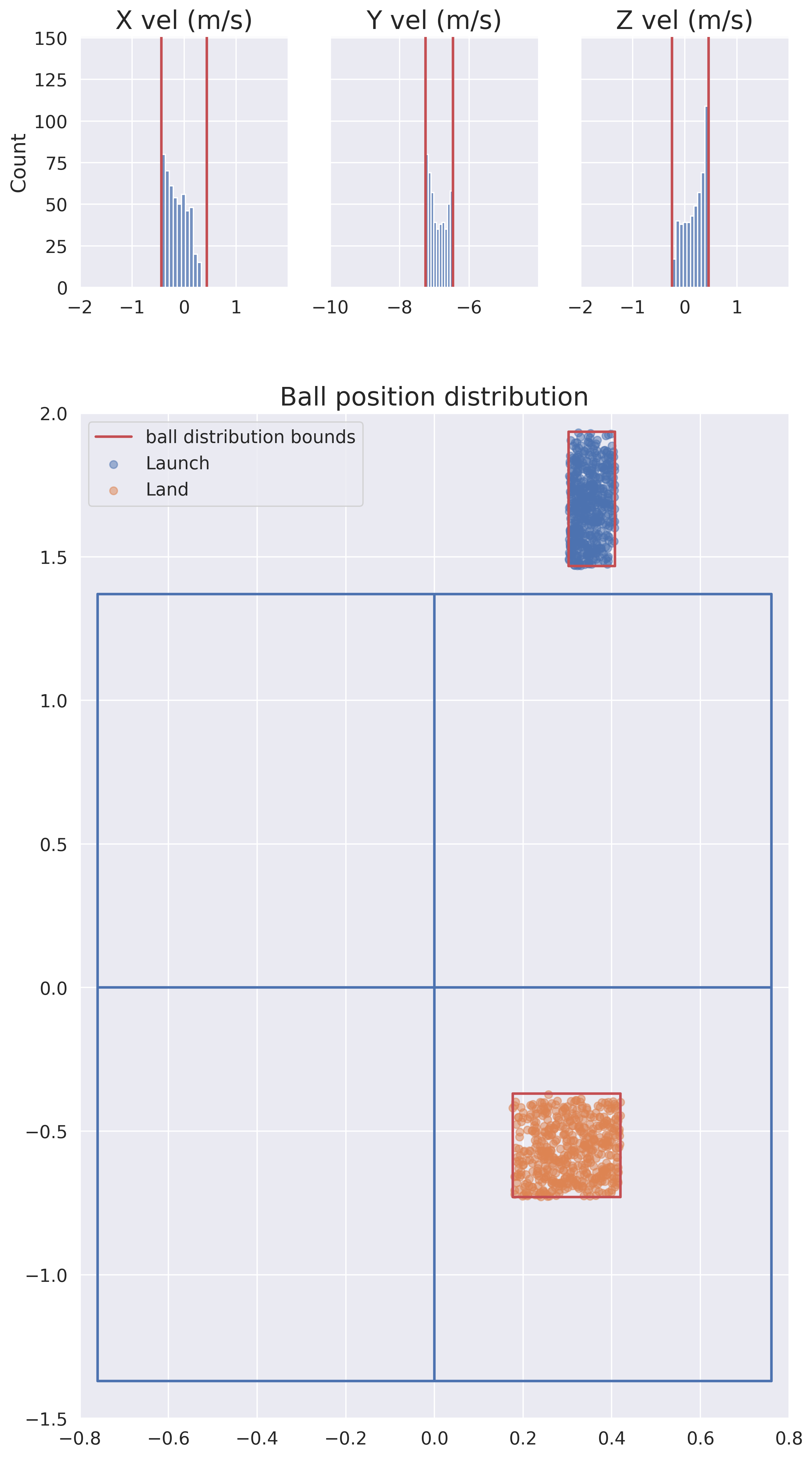}
\includegraphics[width=0.28\textwidth]{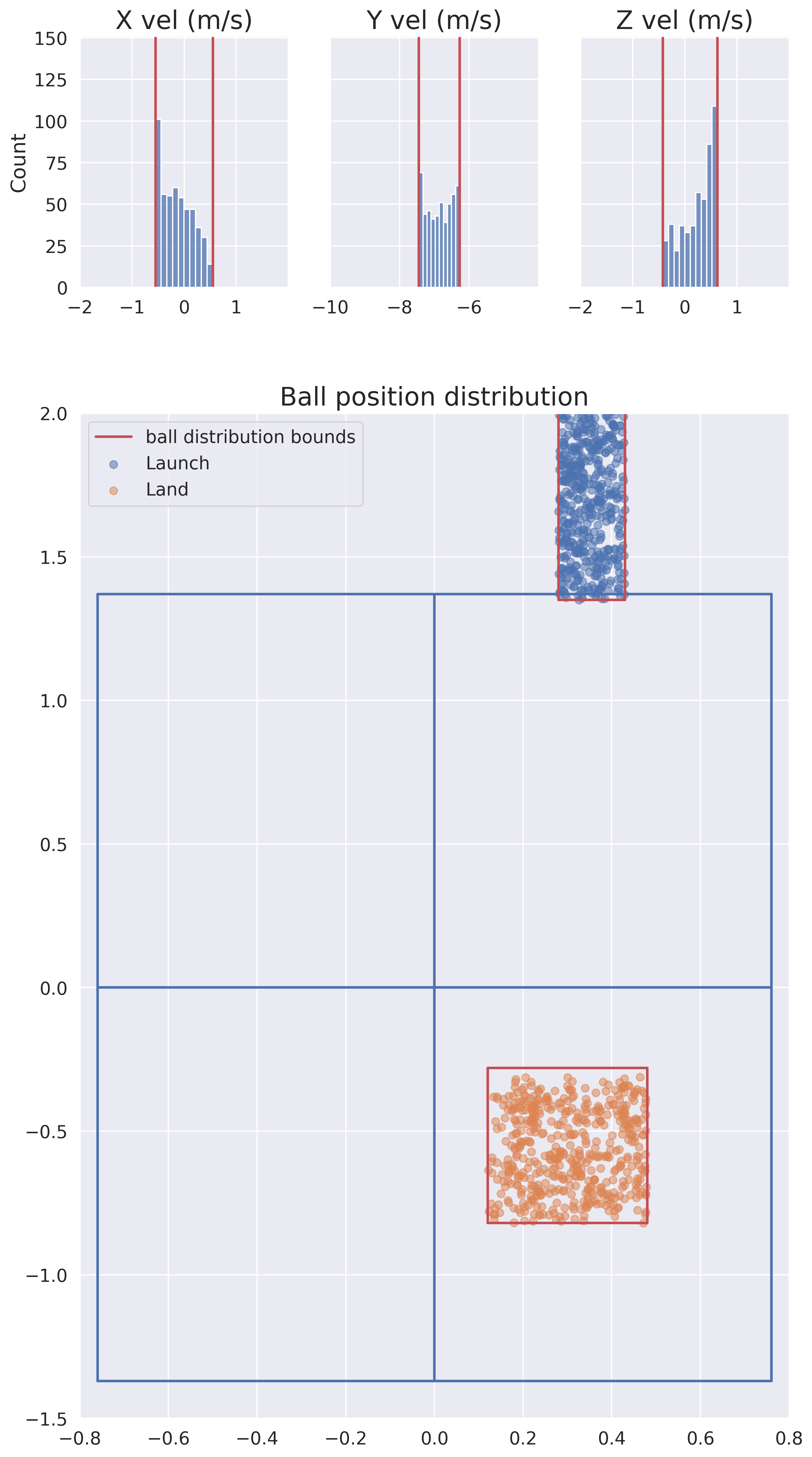}
\includegraphics[width=0.28\textwidth]{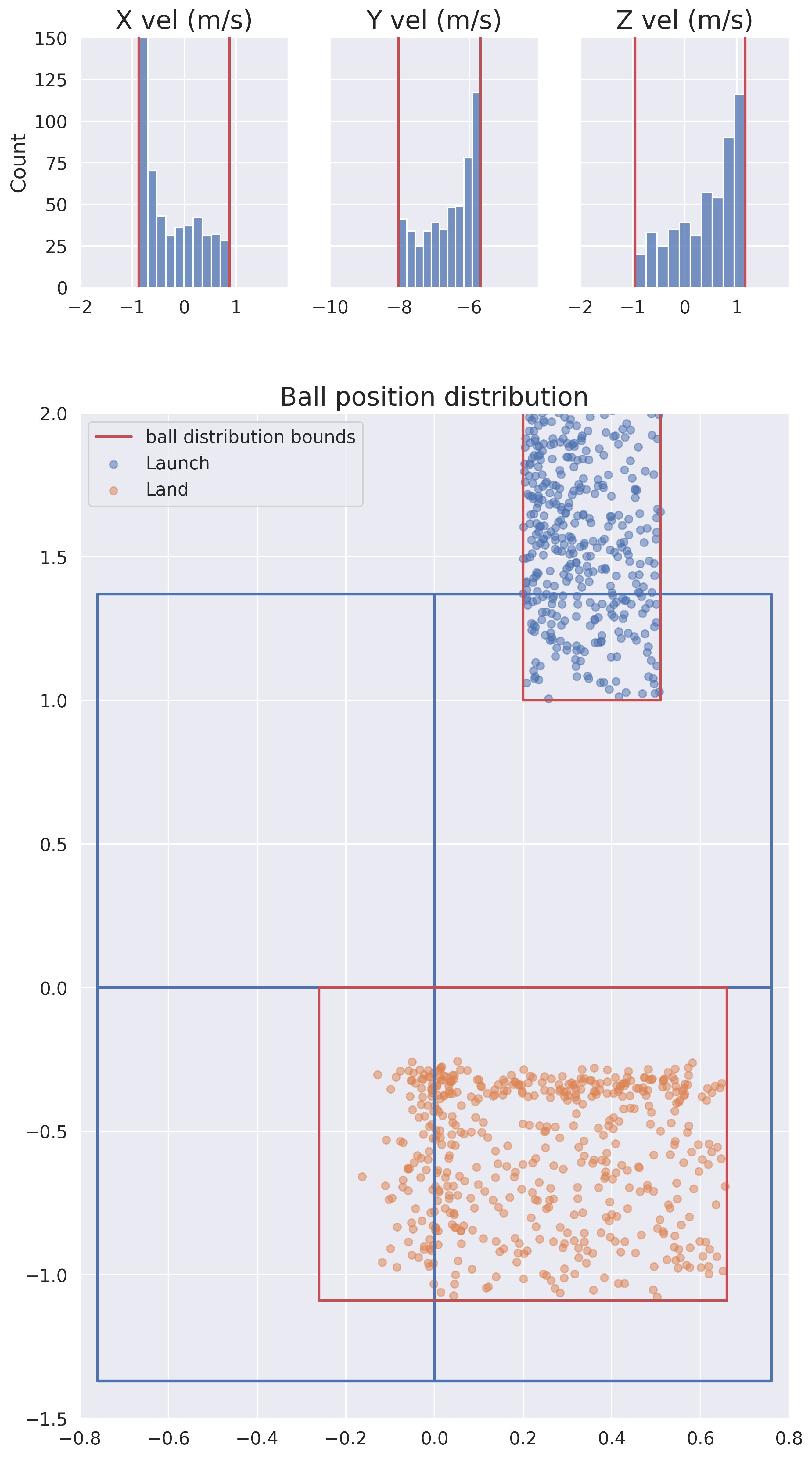}
\includegraphics[width=0.28\textwidth]{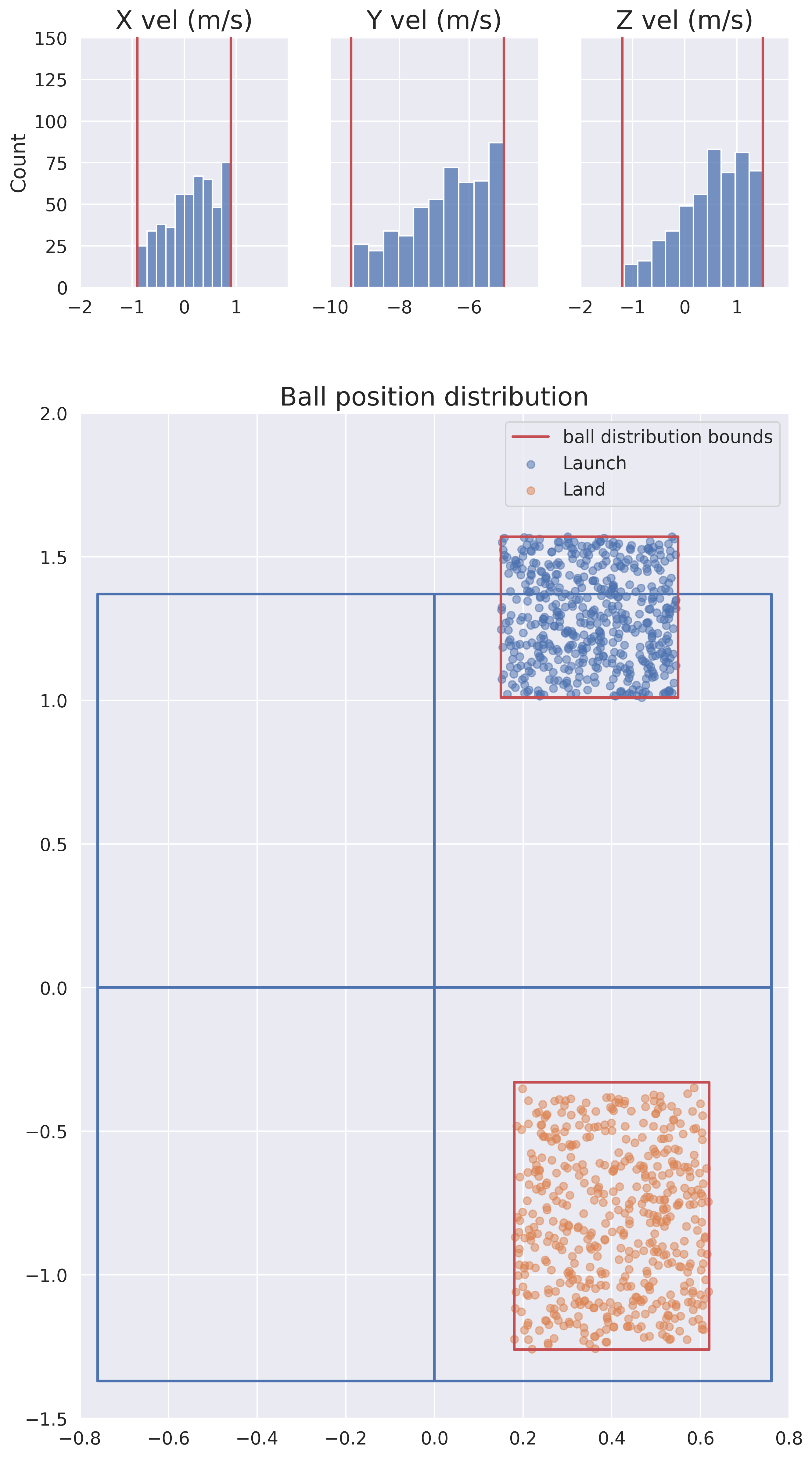}
\includegraphics[width=0.28\textwidth]{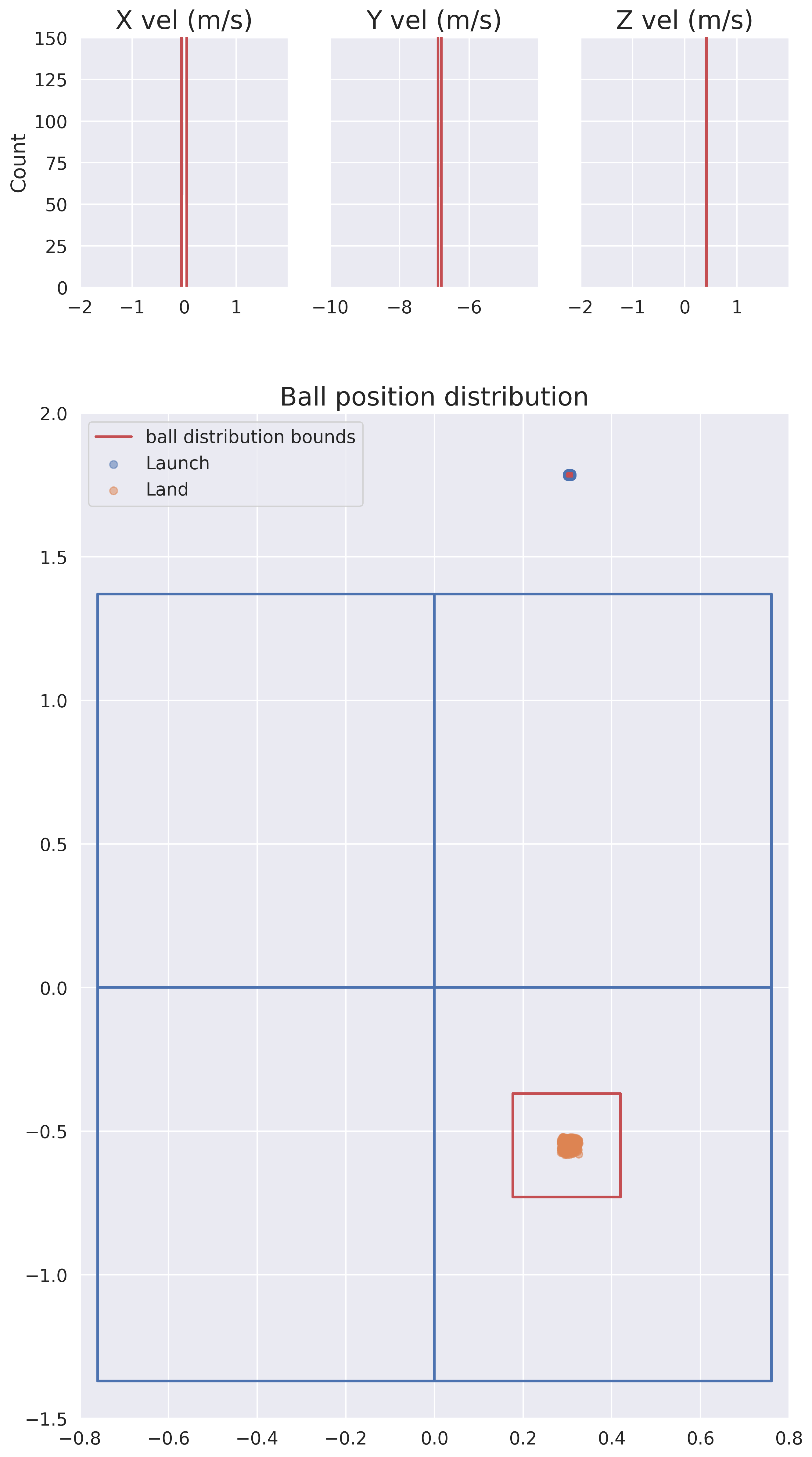}
\includegraphics[width=0.28\textwidth]{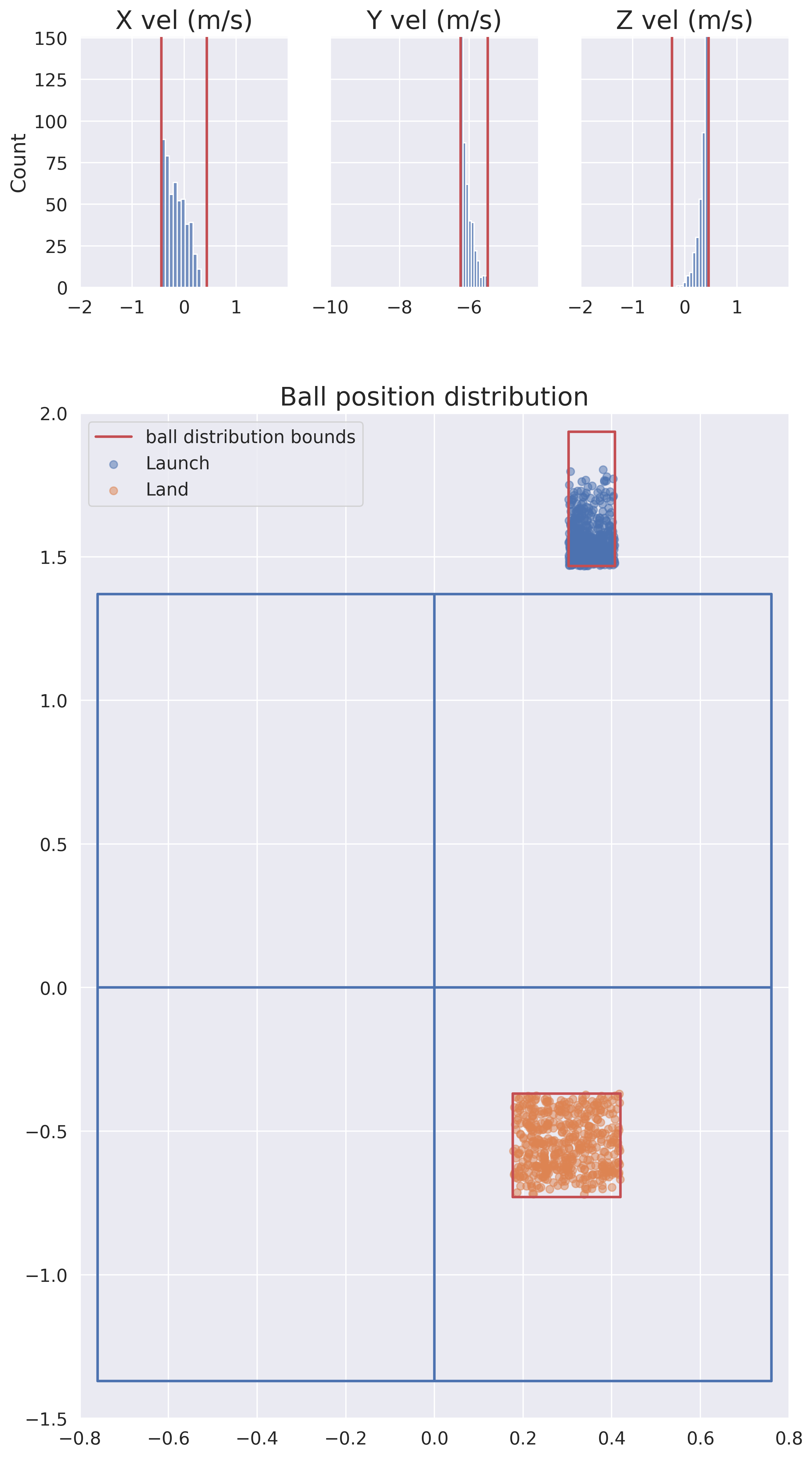}
\caption{Visualization of all the ball distributions used in the simulated parameter ball distribution study. A distribution is visualized by sampling 500 initial ball conditions and plotting a histogram of the ball velocity in each dimension (3 small charts), and plotting the initial ball xy and the landing ball xy below it. Red lines mark the boundaries of the distribution. Distributions are shown as follows: (top left) thrower (baseline), (top center) medium, (top right) wide, (bottom left) thrower 2, (bottom center) tiny, (bottom right) velocity offset.}
\label{fig:ball_dists_viz}
\end{figure*}

\autoref{table:latency-study-values} presents the latency values for each of the assessed settings. \autoref{table:ball-study-values} contains the details of all of the different ball distributions and \autoref{fig:ball_dists_viz} visualizes them. A distribution is visualized by sampling 500 initial ball conditions and plotting a histogram of the ball velocity in each dimension, and plotting the initial ball xy and the landing ball xy below it. \autoref{table:physical-study-values} gives details of the tuned and measured physical parameters.

\begin{table*}
\centering
\scriptsize
\begin{tabular}{|| l | c c c c c || }
\hline
\multicolumn{1}{|| l |}{}  & \multicolumn{5}{c ||}{Latencies (ms): $\mu$ ($\sigma$)} \\
Component & 100\% (baseline) & 0\% & 20\% & 50\%  & 150\%  \\
\hline \hline
Ball observation & 40 (8.2) & 0 & 8 (3.7) & 20 (5.8) & 60 (10.0) \\
ABB observation & 29 (8.2) & 0 & 5.8 (3.7) & 14.5 (5.8) & 43.4 (10.0) \\
Festo observation & 33 (9.0) & 0 & 6.6 (4.0) & 16.5 (6.4) & 49.5 (11.0) \\
ABB action & 71 (5.7) & 0 & 14.2 (2.5) & 35.5 (4.0) & 106.5 (7.0) \\
Festo action & 64.5 (11.5) & 0 & 12.9 (5.1) & 32.3 (8.1) & 96.8 (14.1) \\
\hline
\end{tabular}
\caption{Values used in the simulated latency study.}
\label{table:latency-study-values}
\end{table*}

\begin{table*}
\centering
\scriptsize
\begin{tabular}{|| l | c c c c c ||}
\hline
 Component & Thrower (baseline) & Medium & Wide & Tiny & Thrower 2 \\
 \hline \hline
 \textit{Initial ball velocity} &  &  &  &  &  \\
 Min x velocity & -0.44 & -0.55 & -0.87 & -0.05 & -0.9 \\
 Max x velocity & 0.44 & 0.55 & 0.87 & 0.05 & 0.9 \\
 Min y velocity & -7.25 & -7.45 & -8.04 & -6.90 & -9.4 \\
 Max y velocity & -6.47 & -6.27 & -5.68 & -6.80 & -5.0 \\
 Min z velocity & -0.24 & -0.42 & -0.95 & 0.41 & -1.2 \\
 Max z velocity & 0.46 & 0.63 & 1.16 & 0.42 & 1.5 \\
 \hline
 \textit{Initial ball position} &  &  &  &  &  \\
 Min x start & 0.30 & 0.28 & 0.20 & 0.30 & 0.15 \\
 Max x start & 0.41 & 0.43 & 0.51 & 0.31 & 0.55 \\
 Min y start & 1.47 & 1.35 & 1.00 & 1.78 & 1.01 \\
 Max y start & 1.94 & 2.05 & 2.40 & 1.79 & 1.57 \\
 Min z start & 0.55 & 0.54 & 0.50 & 0.57 & 0.25 \\
 Max z start & 0.61 & 0.63 & 0.67 & 0.58 & 0.64 \\
 \hline
 \textit{Ball landing bounds} &  &  &  &  &  \\
 Min x land & 0.18 & 0.12 & -0.26 & 0.18 & 0.18 \\
 Max x land & 0.42 & 0.48 & 0.66 & 0.42 & 0.62 \\
 Min y land & -0.73 & -0.82 & -1.09 & -0.73 & -1.26 \\
 Max y land & -0.37 & -0.28 & 0 & -0.37 & -0.33 \\
 \hline
\end{tabular}
\caption{Values used in the simulated ball distribution study.}
\label{table:ball-study-values}
\end{table*}

\begin{table*}
\centering
\scriptsize
\begin{tabular}{|| l | c c c||}
\hline
Parameter & Tuned (baseline)  & Measured & Re-measured post changes*\\
\hline \hline
\textit{Table} &  &  & \\
Restitution coefficient & 0.9 +/- 0.15 &  0.92 +/- 0.15 & 0.9 +/- 0.15\\
Lateral friction & 0.1 & 0.33 & 0.1 \\
Rolling friction & 0.1 & 0.1 & 0.001\\
Spinning friction & 0.1 & 0.1 & 0.001\\
\hline
\textit{Paddle} &  &  &\\
Mass & 80g & 112g & 136g\\
Restitution coefficient & 0.7 +/- 0.15 &  0.78 +/- 0.15 & 0.44 +/- 0.15\\
Lateral friction & 0.2 & 0.47 & 1.092\\
Rolling friction & 0.2 & 0.1 & 0.001\\
Spinning friction & 0.2 & 0.1 & 0.001\\
\hline
\textit{Ball} &  &  &\\
Mass & 2.7g & 2.7g & 2.7g\\
Restitution coefficient & 0.9 & 0.9 & 0.9\\
Lateral friction & 0.1 & 0.1 & 0.1\\
Rolling friction & 0.1 & 0.1 & 0.001\\
Spinning friction & 0.1 & 0.1 & 0.001\\
Linear damping & 0.0 & 0.0 & 0.0\\
Angular damping & 0.0 & 0.0 & 0.0\\
 \hline
\end{tabular}
\captionsetup{justification=centering}
\caption{Values used in the simulated physical parameters study.\\+/- values indicate randomization range. If no +/- value is given the\\value is not randomized during training. * see Appendix \autoref{app:system_id_revisited}.}
\label{table:physical-study-values}
\end{table*}

\subsection{Task Space Studies: Additional Results}
\label{app:task_space}

Additional results from training in task space (Figure \ref{fig:task_space_app}) show that it enables more seeds to solve the task, likely by making the problem more intuitive to learn for the training algorithm. This trend is more pronounced in the harder problem setting (damped environment). Here only 10\% of joint space policy seeds solve the task compared with around 80\% of task space policy seeds. We also show the results of zero-shot transfer of the converged seeds. In the original environment, the transfer performance of task space policies is slightly lower than joint space. Looking at the behavior, we see that most of the balls are returned short, and hit the net. This phenomena can be explained by the sim-to-real gap and that policies trained in task space prefer softer returns (with less velocity on the paddle). On the other hand, in the harder (damped) environment, task space policies learn to return faster and more dynamically. In this setting, transfer to the real hardware is much better, with task space policies returning 97\% of the balls and scoring 1.95 out of 2.0.

\subsection{Debugging}

The many interacting components in this system create a complex set of dependencies. If the system suddenly starts performing worse, is it a vision problem, a hardware failure, or just a bad training run? A major design decision was to be able to test as many of these components independently as possible and to test them as regularly as possible. The design of the system allows it to remotely and automatically run a suite of tests with the latest software revision every night, ensuring that any problems are detected before anyone needs to work with the robot in the morning.  

Each test in the suite exercises a particular aspect of the system and can help isolate problems. For example, a test that simply repeats a known sequence of commands ensures the hardware and control stack are functional while a test that feeds the policy a sequence of ball positions focuses on the inference infrastructure, independent of the complex vision stack. Additionally, by running these tests every day, an acceptable range of performance can be ascertained and trends can be tracked. The independent evaluation also enables testing and verification of changes to the various components. For example, when changing from Python to C++ control discussed in \autoref{sec:control} the metrics from the nightly tests were used to judge if the new control mechanisms were working as expected.

\begin{figure*}[!htb]
    \centering
    \includegraphics[width=\textwidth]{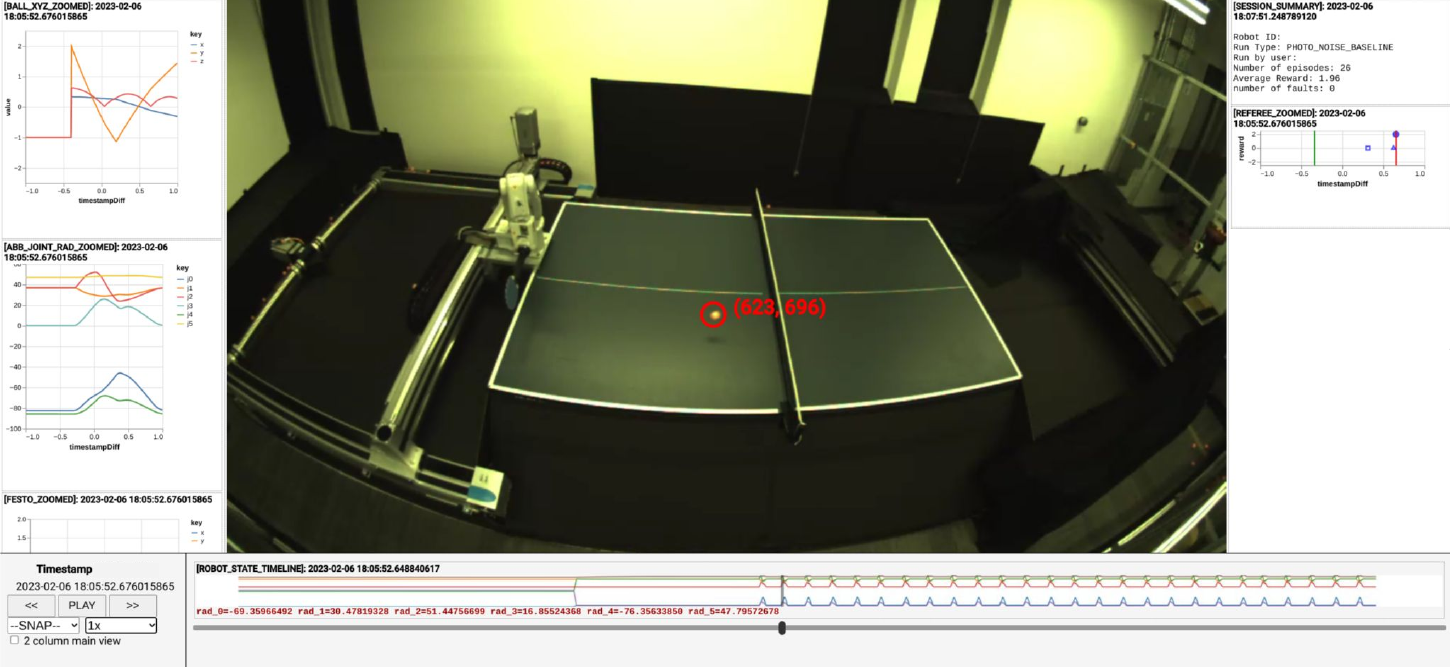}
    \caption{Debugging visualization used in the system. Sensitive information has been redacted. This interface synthesizes logs from several components in a unified interface that makes debugging the system and understanding its state very straightforward.}
    \label{fig:viz}
\end{figure*}

Due to the agile and interconnected nature of a system, it is also nearly impossible to debug in real time and many issues can only be reproduced when the whole system is running at full speed. Another key decision was to log \emph{everything}. In the nightly tests described above, not only are the results logged but many key metrics such as latency are captured which can further isolate failures. Additionally, the state of all aspects of the robot, environment, and even intermediate states (e.g. the safety simulator) are logged and can be played back later in a convenient interface (\autoref{fig:viz} that shows the user many aspects of the system at once and allows them to step through the environment states in a way that's impossible to do on the actual robot. While the initial costs of planning and executing efficient logging system are high, they more than pay for themselves in diagnostic ability. The next section dives more deeply into the various aspects of logging.

\subsection{Logging}
\label{app:logging}

Logging throughout the system is very useful for multiple reasons: Debugging, Timing \& Performance analysis, Visualization, Stats \& Metric tracking. The logs are primarily divided between:
\begin{enumerate}
    \item Structured Logs: This includes detailed high frequency logging used for timing \& performance analysis as well as visualizations. This data is post-processed to give lower-granularity summary metrics.
    \item Unstructured Logs: These are human readable logs, primarily capturing informational data as well as unexpected errors \& exception cases.
    \item Experiment Metadata: A small amount of metadata describing each experiment run helps organize all logs by runs.
\end{enumerate}

High frequency logging is useful to introspect into the performance of the system. Individual raw events are logged, including hardware feedback as well as the data at different steps in the pipeline through transformations, agent inference and back to hardware commands. Logging the same data through multiple steps of transformation along with tracing identifiers, helps us track the time taken between steps to analyze the performance of the system. Raw high frequency data allows us to capture fine-grained patterns in the timing distribution that are not as easily revealed by just summary metrics. Care must be taken when logging at high frequency/throughput to not have performance issues from the logging system itself. Logging system overhead is low on the Python side, saving the heavy lifting for an asynchronous C++ thread that actually saves and uploads the logs. Other performance metrics we log include CPU and Thread utilization.

\end{document}